\documentclass[sigconf]{acmart}
\AtBeginDocument{%
  \providecommand\BibTeX{{%
    \normalfont B\kern-0.5em{\scshape i\kern-0.25em b}\kern-0.8em\TeX}}}

\usepackage{xspace}
\usepackage{subcaption}
\usepackage{booktabs}
\usepackage{enumitem}

\usepackage{xcolor}
\usepackage{soul}

\usepackage[final]{changes}

\newif{\ifhidecomments}
\ifhidecomments
    \newcommand{\chenhao}[1]{}
    \newcommand{\viv}[1]{}
    \newcommand{\sam}[1]{}
    \newcommand{\vera}[1]{}
\else
    \newcommand{\chenhao}[1]{\textcolor{blue}{[#1 ---\textsc{ct}]}}
    \newcommand{\viv}[1]{\textcolor{red}{[#1 ---\textsc{viv}]}}
    \newcommand{\sam}[1]{\textcolor{violet}{[#1 ---\textsc{sam}]}}
    \newcommand{\vera}[1]{\textcolor{orange}{[#1 ---\textsc{vera}]}}
    
\fi

\newcommand{\ind}{in-distribution\xspace}

\newcommand{\ood}{out-of-distribution\xspace}
\newcommand{\Ood}{Out-of-distribution\xspace}

\newcommand{\para}[1]{\noindent{\bf #1}}
\newcommand{\figref}[1]{Fig.~\ref{#1}}
\newcommand{\secref}[1]{Section \ref{#1}}
\newcommand{\tabref}[1]{Table~\ref{#1}}

\copyrightyear{2022} 
\acmYear{2022} 
\setcopyright{rightsretained} 
\acmConference[CHI '22]{CHI Conference on Human Factors in Computing Systems}{April 29-May 5, 2022}{New Orleans, LA, USA}
\acmBooktitle{CHI Conference on Human Factors in Computing Systems (CHI '22), April 29-May 5, 2022, New Orleans, LA, USA}
\acmDOI{10.1145/3491102.3501999}
\acmISBN{978-1-4503-9157-3/22/04}

\begin{document}

\title{Human-AI Collaboration via Conditional Delegation: \\A Case Study of Content Moderation
}

\author{Vivian Lai}
\email{vivian.lai@colorado.edu}
\affiliation{%
  \institution{University of Colorado Boulder}
  \city{Boulder}
  \state{CO}
  \country{USA}
}

\author{Samuel Carton}
\email{samuel.carton@colorado.edu}
\affiliation{%
  \institution{University of Colorado Boulder}
  \city{Boulder}
  \state{CO}
  \country{USA}
}

\author{Rajat Bhatnagar}
\email{rajat.bhatnagar@colorado.edu}
\affiliation{%
  \institution{Amazon}
  \city{Seattle}
  \state{WA}
  \country{USA}
}

\author{Q. Vera Liao}
 \authornote{Part of this work was completed while the fourth author was working at IBM Research.}
 \email{veraliao@microsoft.com}
\affiliation{%
  \institution{Microsoft Research}
  \city{Montreal}
  \country{Canada}
}

\author{Yunfeng Zhang}
 \authornote{Part of this work was completed while the fifth author was working at IBM Research.}
 \email{yunfengz@twitter.com}
 \affiliation{%
 \institution{Twitter Inc.}
  \city{New York}
  \state{NY}
  \country{USA}
}

\author{Chenhao Tan}
\email{chenhao@uchicago.edu}
\affiliation{%
  \institution{University of Chicago}
  \city{Chicago}
  \state{IL}
  \country{USA}
  }

\renewcommand{\shortauthors}{Lai et al.}

\begin{abstract}

Despite impressive performance in many benchmark datasets, AI models can still make mistakes, especially among out-of-distribution examples. 
It remains an open question how such imperfect models can be used effectively in collaboration with humans.
Prior work has focused on AI assistance that helps people make individual high-stakes decisions, which is not scalable for a large amount of relatively low-stakes decisions, e.g., moderating social media comments. 
Instead, we propose conditional delegation as an alternative paradigm for human-AI collaboration where humans create rules to indicate trustworthy regions of a model. 
Using content moderation as a testbed\deleted{and leveraging two datasets to simulate \ind and \ood scenarios}, we develop novel interfaces to assist humans in creating conditional delegation rules and conduct a randomized experiment \added{with two datasets to simulate \ind and \ood scenarios}. 
Our study demonstrates the promise of conditional delegation in improving model performance and provides insights into design for this novel paradigm, including the effect of AI explanations.

\end{abstract}

\ccsdesc[500]{Human-centered computing~Collaborative and social computing}
\ccsdesc[500]{Computing methodologies~Artificial intelligence}
\ccsdesc[500]{Applied computing~Law, social and behavioral sciences}

\maketitle

\section{Introduction}
\label{sec:introduction}

As AI performance grows rapidly and even surpasses humans in benchmark datasets \citep{kleinberg2018human,he2015delving,mckinney2020international,silver2018general,brown2019superhuman},
AI models hold great promise for improving human decision making in a wide variety of domains.
However, full automation may not be desirable for ethical, legal, and safety reasons, especially in high-stakes domains~\cite{cai2019human,lubars+tan:19,lai2019human,green2019principles}.
In particular, one well-known problem with the current 
AI models is {\em distribution shift}.
Namely, AI performance can significantly drop for \ood examples that are different from the training data (\ind examples)~\citep{mccoy-etal-2019-right,clark2019don,jia2017adversarial,beede2020human}.

Human-AI collaboration is thus critical for effective integration of AI models into human decision making processes~\cite{cai2019hello,wang2019human,arous2020opencrowd,ashktorab2020human,nguyen2018believe,bansal2019beyond,bansal2021does,o2020human}.
Many studies have investigated the role of AI in assisting humans in making individual decisions~\cite{lai2019human,lai+liu+tan:20,green2019principles,green2019disparate,zhang2020effect,poursabzi2021manipulating,carton2020feature,jung2020limits,weerts2019human,beede2020human,wang2021explanations,lundberg2018explainable}, e.g., predicting whether a person will recidivate in the near future.
Such decisions are non-trivial even for human experts (e.g., judges) and AI models can potentially offer insights through their predictions and explanations.
This approach is well suited for high-stakes domains, where humans are expected to make the final decision on every case (e.g., judges in bailing decisions).
However, human-AI collaboration on every single decision is not scalable and is thus less appropriate for tasks involving a large amount of relatively low-stakes decisions.
One such example is content moderation, where moderator decisions on individual comments for further actions (e.g., hiding the content
or prompting further review, 
depending on the community policy) are 
of limited consequence; instead the key challenge lies in dealing with the massive scale of comments.
Such tasks can benefit from a greater level of automation~\citep{gillespie:2018a,gorwa2020algorithmic,chandrasekharan2019crossmod}.

\begin{figure*}[t]
    \includegraphics[width=0.9\textwidth]{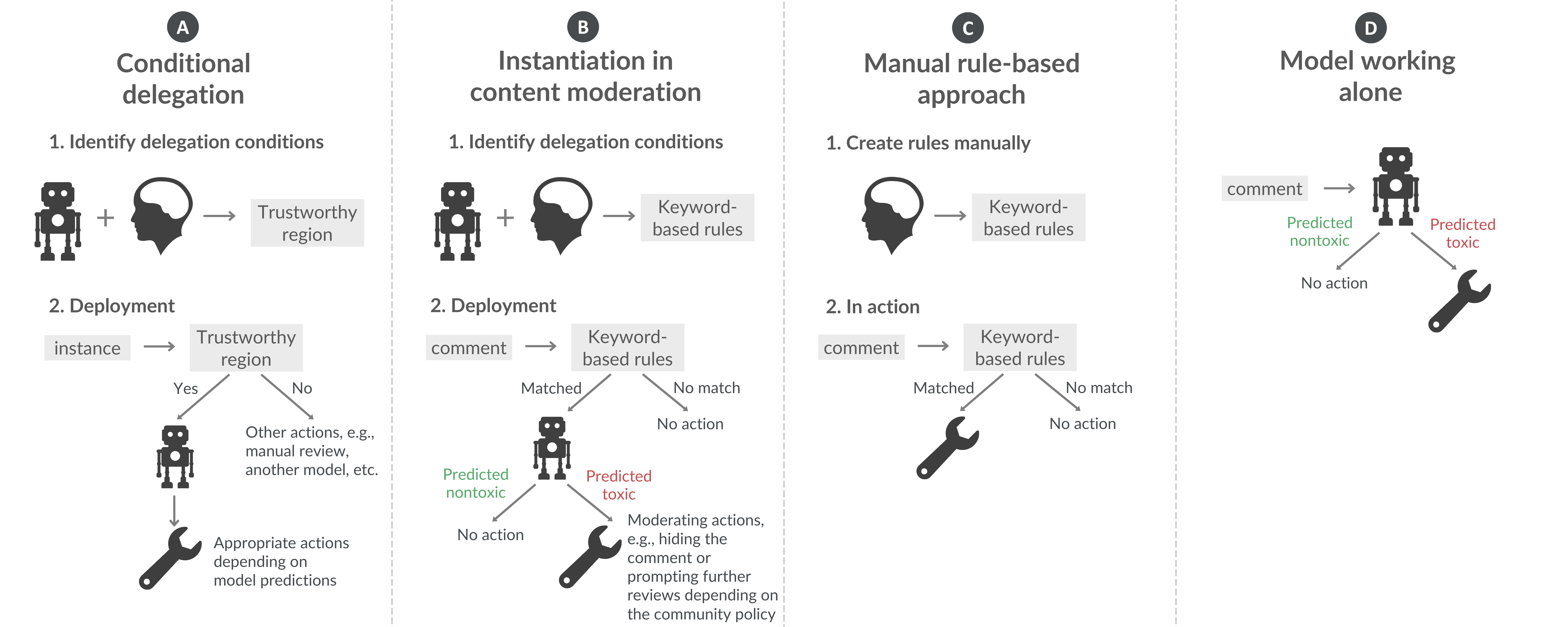}
    \caption{Illustration of conditional delegation. Part A shows a general form of conditional delegation. Humans and AI work together to identify trustworthy regions of AI. Then once deployed, the AI model only affects the instances that belong to the trustworthy regions. Part B instantiates conditional delegation in the context of content moderation for this work. The right columns shows the contrast of the current manual rule-based approach for content moderation (Part C) and the model working alone (Part D). 
    }
    \Description{There are four potential approaches for content moderation. They are condition delegation, instantiation in content moderation, manual rule-based approach, and just the model.}
    \label{fig:intro}
\end{figure*}

In this work, we propose an alternative paradigm of human-AI collaboration --- conditional delegation. 
\figref{fig:intro}(A) illustrates a general form of conditional delegation.
Human and AI work together to identify trustworthy regions of AI before deployment, i.e., model decisions are reliable or trustworthy for examples within these regions. Once deployed, the AI model only affects decisions for instances in the trustworthy regions. For the rest, another set of actions can be taken such as manual review or employing a different model since the given AI's decisions on them cannot be trusted. This approach employs a greater level of automation than human-AI collaboration on every single decision and provides human with active control on when to use an AI model and in what ways.

We use content moderation as a testbed.
\figref{fig:intro}(B) shows one possible instantiation in this context.
Trustworthy regions can be operationalized with a collection of keyword-based rules created by human-AI collaboration before deployment. 
For example, after inspecting AI predictions on comments with the word ``retard'', the human may decide that AI works well on them and set ``retard'' as a conditional delegation rule. 
Once deployed, comments that fall within these trustworthy regions, i.e.,  {\em containing any keywords} specified by human, if {\em predicted toxic}, can be reliably %
reported for final actions, such as being hidden or sent for further review, depending on the community policy. 

Notably, the task for humans to create \textit{conditional delegation rules} share some similarity with what many social media moderators are already doing by writing manual automation rules to deal with the massive amount of comments (\figref{fig:intro}(C)). For example, moderators on Reddit use a tool called AutoModerator, with which they manually customize a rule-based system to automatically identify comments for deleting or reporting for further review~\cite{jhaver2019human,chandrasekharan2019crossmod}. This approach, however, misses out the benefit of AI especially since rigid rules often do not work on informal languages such as social media posts (e.g., containing swear words without being toxic). Without significantly altering content moderators' workflow, conditional delegation offers a promising approach to utilize AI, even if the model is not optimized for the community-specific content and should not be blindly trusted to work alone for every comment (\figref{fig:intro}(D)).

In this instantiation, a key difference from individual human-AI decision making lies in the success criteria: while the quality of individual decisions (e.g., accuracy) is often the target in individual decision making, \textit{precision} and \textit{coverage} are critical for conditional delegation because moderation actions will only happen on comments that are predicted toxic.\footnote{Depending on the workflow, avoiding false negatives could be important in other instantiations.}
Precision ensures that AI behavior is indeed trustworthy in the delegation mode and avoids unnecessary actions, whether it is mistaken deletion or extra work for further review. Coverage warrants that the AI model can identify as many toxic comments as possible to alleviate the scalability issues. In the context of content moderation, recall (identifying all toxic comments) is often less of a priority given the limited time for content moderators, who are often volunteers, to deal with a massive amount of incoming comments. This is reflected in the current workflow using the manual rule-based approach (\figref{fig:intro}(C)), where comments falling outside the rules are ignored without taking an action.  We assume the same workflow in our study and only focus on the precision and coverage related metrics for comments within the scope of keywords rules. 

In this study, our {\em primary} interest is to investigate whether humans can effectively identify trustworthy regions for conditional delegation to improve the model precision with a good coverage, compared to the current manual rule-based approach (\figref{fig:intro}(C)) and the model working alone (\figref{fig:intro}(D)). Furthermore, we explore the effectiveness in two different AI scenarios: using an AI trained on the community specific data (\textit{in-distribution}), and one trained on different data (\textit{out-of-distribution}). The out-of-distribution model would perform much worse, but conditional delegation offers a potential means to improve through human-AI collaboration.

Our second set of contribution is to inform design of interfaces that support people to create high-quality conditional delegation rules. When given an AI model, content moderators often do not have labeled comments to quantify model performance. It would be helpful for them to observe model behaviors on their own data of interest to identify good delegation rules (i.e., trustworthy regions). To facilitate the creation of keyword-based rules, we develop an interface that allows participants who act as moderators to perform keywords search and observe model behavior on the search results. We provide and study the effects of several delegation support features, including predicted labels, local explanations that show the rationales behind predictions, and global explanations that provide an overview of the model.  

To summarize, we ask the following research questions:

\begin{itemize}[topsep=0pt]
    \item[RQ1.] Can users create keyword-based rules for conditional delegation that improves model precision,
    so that these rules correspond to trustworthy regions?
    \item[RQ2.] How do the performance of conditional delegation and user experiences (such as engagement and subjective perceptions) vary between \ind and \ood AI?
    \item[RQ3.] What are the effects of delegation support features on performance and user experiences, including showing prediction labels, local explanations, and global explanations?  
\end{itemize}

Through a randomized experiment with 240 mechanical turkers, we show that even crowdworkers are able to create high-quality rules that lead to higher precision with conditional delegation than the model working alone. Especially when applied to an in-distribution AI, which already outperforms the manual rule-based approach for content moderation, conditional delegation further enhances the performance, leading to ``complementary performance'' (i.e., human+AI > AI and human+AI > human) \citep{bansal2021does}. For out-of-distribution AI used in this study, conditional delegation improves the model performance but does not suffice in compensating for the performance disadvantage of AI to outperform the manual rule-based approach. We also found that model explanations can improve efficiency in identifying delegation conditions and, with weak evidence, improve user experiences. 

Overall, our work provides a new perspective to the emerging area of human-AI collaboration. 
Our \replaced{core contribution}{results} is to demonstrate that conditional delegation is a promising alternative paradigm that allows users to control when to trust or distrust AI. We also contribute a set of interface features to assist people in creating conditional delegation rules and an empirical understanding of their effects. The diverging performance of \ind and \ood highlights the importance of considering the effect of distribution shift when conducting empirical studies of human-AI collaboration to inform the generalizablity of results, echoing recent findings in other studies \citep{chiang2021you,liu2021understanding}.

\section{Related Work}
\label{sec:related_work}

\subsection{Human-AI Collaboration}

Terms like ``human-AI collaboration''~\cite{cai2019hello,wang2019human,arous2020opencrowd,ashktorab2020human}, ``human-AI partnership''~\cite{nguyen2018believe}, ``human-AI teaming''~\cite{bansal2019beyond,bansal2021does,o2020human} have emerged in various literature studying the use of AI systems. They reflect a shift of perspective away from complete automation by AI. Fostering effective human-AI collaboration is not only critical for safety reasons,
especially in high-stakes domains~\cite{cai2019human}, but also necessary to harnessing the complementarity of human and AI intelligence to achieve optimal outcome~\cite{bansal2019updates,wilder2020learning}, reduce computational complexity~\cite{holzinger2016interactive}, and enable novel technologies that are beyond the current capabilities of AI~\cite{wang2019human,cranshaw2017calendar}.

Many forms of human-AI collaboration have been explored. The term ``human-in-the-loop'' is used broadly, but often refers to interactive training paradigm where the AI receives input from the human to improve its performance. For example, the field of interactive Machine Learning 
~\cite{holzinger2016interactive,fails2003interactive,amershi2014power,dudley2018review}, at the intersection of ML and HCI, develops systems that allow end users to guide model behavior. This kind of paradigm allows humans to directly impact the working of AI, and requires using algorithms that can incorporate human input to update the model, which can be technically challenging or infeasible in practice. 

Another rich area to study human-AI collaboration is AI-assisted~\cite{zhang2020effect,wang2021explanations,buccinca2021trust} or ``machine/algorithm-in-the-loop'' decision-making~\cite{green2019principles,lai2019human}. In this paradigm, AI performs an assistive role by providing a prediction or recommendation, while the human decision maker makes the final call and may choose to accept or reject the AI recommendation. Several studies explored the questions of whether and how to achieve {\em complementary performance}, i.e., the collaborative decision outcome outperforming human or AI alone~\cite{zhang2020effect,bansal2021does,lai2019human}. The empirical results, however, are mixed at best, because there was either insufficient complementarity in human and AI's domain knowledge or a lack of ability for people to judge the reliability of AI recommendations. 
This approach tends to focus on high-stakes decisions and are not scalable in the number of decisions because humans are required to make each  decision.

Another line of work explores intelligent systems and considers different tasks that AI can perform and the optimal level of automation versus human agency~\cite{wang2021much,mackeprang2019discovering,lai2019human}. For example, building on a classic model of levels of automation~\cite{parasuraman2000model},
 \citet{mackeprang2019discovering} proposed a design framework that decomposes the design space of an intelligent system into sub-tasks then allocates human, AI or both to perform each sub-task. 
The goal of our work is to have AI partially automate a large volume of decisions rather than assisting individual decisions. Extending existing models of human agency and automation~\cite{parasuraman2000model,mackeprang2019discovering}, we introduce \textit{proactive} human agency, with which human can act and exercise control prior to model deployment, instead of reacting to model outputs. By conducting a controlled experiment, we explore whether this new human-AI collaboration paradigm can achieve complementary performance by outperforming AI and manual approaches.
\added{While some prior work also discussed delegation based on predicted outputs (e.g., predicted probability) \citep{keswani2021towards,chandrasekharan2019crossmod}, our work focuses on identifying trustworthy regions in the input space.
Furthermore, to the best of our knowledge, our work is the first study with controlled experiments to examine the effect of conditional delegation.}

\subsection{AI explanations for human-AI interaction}
Mental model, defined as an understanding of how a system works, is a key concept in human-computer interaction~\cite{norman2013design}. Having an appropriate mental model allows people to accurately anticipate a system's behaviors and interact more effectively. People's mental model can be refined by explanations of how the system works. Therefore, explanation and transparency features have long been an interest of HCI research on various technologies~\cite{abdul2018trends,herlocker2000explaining,lim2009and,rader2018explanations}.

Recently, 
AI explanations have gained much attention~\cite{lai2019human,liao2020questioning,ghai2020explainable,dodge2019explaining,buccinca2021trust,bansal2021does,zhang2020effect}. The popularity of complex, inscrutable AI models such as deep neural networks make the difficulty of understanding a primary challenge for modern AI technologies. This challenge has given rise to a technical field of explainable AI (XAI), producing an abundance of techniques that aim to make AI more understandable by people. While the landscape of XAI technique is beyond the scope of this paper~\cite{guidotti2018survey,adadi2018peeking,gilpin2018explaining}, an important distinction relevant to our study is the contrast between \textit{local explanations}, which focus on explaining the rationale for a particular prediction, versus \textit{global explanations}, which aim to give a high-level understanding of how the AI works.  We explore the effect of both types of explanation in our study and will discuss the details of the XAI techniques used for our toxicity prediction model in the next section.

HCI studies on XAI have found explanations to improve user understanding of AI systems~\cite{cheng2019explaining,ghai2020explainable,buccinca2020proxy}, and somewhat mixed results on enhancing user trust~\cite{cheng2019explaining}, satisfaction~\cite{ghai2020explainable} and willingness to adopt AI systems~\cite{tsai2021exploring}. Moreover, explanations provide additional information that can be utilized to assist the task that people perform. For example, Lai and Tan proposed a spectrum between human agency and full automation for machine learning to assist human decision-making~\cite{lai2019human}, and considered showing explanation as an additional form of machine assistance beyond solely providing prediction labels, and thus increase the level of automated assistance. In interactive machine learning, explanation has been studied as a primary means for people to directly inspect the model limitations, instead of just observing model behaviors, for people to provide feedback~\cite{stumpf2009interacting,ghai2020explainable} to improve the model.

For our conditional delegation task, we hypothesize that explanations of AI model predictions, i.e., keywords that the model bases its prediction on, can give hints to people about keyword rules they should consider, and potentially help them judge the effectiveness of a given rule.

\subsection{\added{Distribution Shift and Experimental Studies on \Ood Examples}}

Current AI models rely on identifying patterns in training datasets.
In a real-world scenario, it is unlikely that models are used to classify data that is exactly the same as the training dataset.
For instance, a moderation team would likely work with a model trained on an existing dataset, then applied to the data on their platform. 
The difference between the training dataset and the deployment data is called distribution shift, which often results in a performance drop \citep{mccoy-etal-2019-right,clark2019don,jia2017adversarial}.  
For instance, \citet{mccoy-etal-2019-right} find that state-of-the-art models in natural language inference adopt three fallible syntactic heuristics and perform around random chance when tested on examples where these heuristics fail.

Despite substantial interest in distribution shift in the AI community, the effect has been rarely examined in empirical studies of human-AI collaboration, with a few recent exceptions \citep{liu2021understanding,chiang2021you}.
\citet{liu2021understanding} demonstrated that there exists a clear difference between \ind and \ood examples when human and AI collaborate to make individual decisions in recidivism prediction and profession prediction.
They suggested that complementary performance is more plausible for \ood examples because of AI's performance drop.
\citet{chiang2021you} examined human reliance on the model in human-AI decision making and found that surprisingly humans rely on AI more \ood, where the AI performance is worse.

The existence of distribution shift is a strong motivation for some form of conditional delegation so that humans can identify the trustworthy regions.
In our setup, however, as we conditionally delegate decisions to AI, strong AI performance \ind is likely more critical for the human-AI collaborative performance.
We thus hypothesize that it is more challenging to identify the trustworthy regions for \ood examples because the model behavior is likely more spurious.

\subsection{Content Moderation}

Content moderation has attracted substantial interest from the research community due to its growing importance in online communities \citep{kiesler2012regulating}.
There is a large body of research studying the effect of moderation on community behavior, including whether one should regulate at all \citep{Chancellor:2016:TIC:2818048.2819963,chandrasekharan2017you,srinivasan+dnm+lee+tan:19,jhaver2019does,Chang-Recidivism:19,seering_shaping_2017}.
In contrast, our work is concerned with the practice of content moderation, i.e., how moderators can efficiently deal with a large number of comments.
The scale of content is the most important argument for some form of automation in content moderation \citep{gillespie:2018a,gorwa2020algorithmic}.
Moreover, an active line of research has investigated the ``emotional labor'' of moderation work by the volunteer moderators \citep{dosono2019moderation,matias2016civic,roberts2014behind}, further highlighting the importance of avoiding burnout for moderators through automation.

One strategy is to use rule-based methods.
For instance, Reddit moderators can configure an AutoModerator bot to set rules for reporting or deleting all comments that contain certain words.\footnote{\url{https://www.reddit.com/wiki/automoderator}.}
The key advantage of this method is that it is entirely under the control of moderators.
Through interviews with 16 moderators, \citet{jhaver2019human} found that AutoModerator improves the efficiency of moderation. However, there exists a need for audit tools to monitor the performance of the keyword rules.
They also highlight the fact that AutoModerator fundamentally changes the work of moderators and may introduce additional unnecessary work.
\citet{chandrasekharan2019crossmod} also found that hard-coded rules are prone to mistakes.

An alternative strategy is to use AI models beyond rule-based approaches.
Toxic comment detection or hatespeech detection has attracted a lot of interest from the AI community \citep{wulczyn_ex_2017,qian-etal-2019-benchmark,wiegand_detection_2019,nobata2016abusive}.
Notably, the Perspective API is reportedly used by the New York Times, Disqus, and other platforms.\footnote{\url{https://www.perspectiveapi.com/case-studies/}.}
However, researchers increasingly recognize the pitfalls of full automation: 
1) models are trained with historical data and can present issues such as gender bias and racial bias in AI models \citep{sap_risk_2019,park-etal-2018-reducing}, potentially exacerbating structural inequalities \citep{blackwell2017classification};
2) there exist diverse rules and preferences of austerity and value in different communities \citep{chandrasekharan_internets_2018,fiesler+al:2018,scheuerman2021framework,smith2020keeping}.
Anecdotally, we deployed a version of our model on a subreddit to report comments that are predicted as toxic, and the moderators asked us to shut it down due to high false positive rates (i.e., low precision).
Inspired by the diversity of rules, \citet{chandrasekharan2019crossmod} proposed a new system that combines classifiers based on different communities and advocated that this tool be configured as part of moderation workflow.

Our effort represents a new direction in exploring the mixed initiative in content moderation.
Conditional delegation combines traditional rule-based approaches and AI models by providing moderators with the ability to decide when to trust or distrust the AI model.
Such rules can be created for any model of choice, so it is orthogonal to the research on improving the capability of AI.
It can also be used to tailor different requirements of precision and tune the tradeoffs between false positives and false negatives.

\section{AI Model}
\label{sec:method}

A critical component of our study is the model used to assist people in content moderation. In this section, we present details of how we obtain the model used in our study and provide an overview of its properties.

\subsection{Model Development}
\label{sec:data_model}

Current AI models are driven by the data used to train the model.
We choose two datasets to simulate the \ind and \ood scenarios. 
We then develop an interpretable model that is trained on the \ind data and achieves reasonable performance on the \ood data.

\para{Data.} In this work, we use a dataset of Wikipedia comments \cite{wulczyn_ex_2017} (henceforth {\em WikiAttack}),
made public by Wikipedia and Google Jigsaw. 
Notably, Jigsaw powers the Perspective API\footnote{https://www.perspectiveapi.com/}, a popular free service for toxic comment detection.
Therefore, using a model derived from this dataset allows ecological validity to our study as the dataset is used by real-world social media platform and community moderators.
We use the original train/test split of \citet{wulczyn_ex_2017}, resulting in 70k comments in the training set and 23K comments in the test set.
We use the test set to evaluate 
the ability of participants to create keyword-based rules for conditional delegation for {\em WikiAttack}.
To simulate the \ood scenario, we use another dataset of hate speech on Reddit \cite{qian-etal-2019-benchmark}, consisting of 22K comments, on which we apply the same model mentioned above.
As a result, the datasets that participants explore to create rules are of comparable size between Wikipedia (\ind) and Reddit (\ood).
Throughout the rest of the paper, we will use {\em \ind} and {\em WikiAttack}, {\em \ood} and {\em Reddit} interchangeably.

\begin{figure}[t]
    \includegraphics[width=0.48\textwidth]{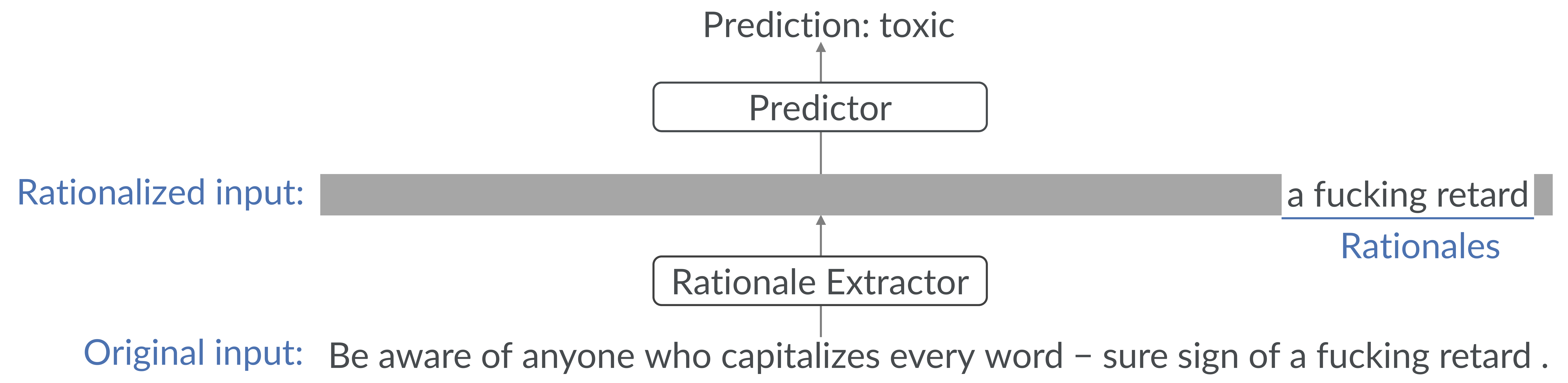}
    \caption{Illustration of the model with an example. The rationale extractor first identifies ``rationales'' in the input, and then the predictor makes the prediction based on the rationales. This model can achieve competitive accuracy while having built-in interpretability because the prediction is made exclusively based on rationales.}
    \Description{The model extracts rationales from the input and makes prediction based on the rationales.}
    \label{fig:model}
\end{figure}

\para{Model.} We use a rationale-style neural architecture \citep{lei2016rationalizing} as the classifier underpinning our tool, producing both explanations and predictions. \figref{fig:model} illustrates our model architecture. It uses one text encoder to identify rationales (i.e., a subset of tokens) from the input, and another text encoder to make predictions based on the rationales.\footnote{Technically, the predictor uses the masked input.}
Trained in tandem with a sparsity objective on the rationales, this model attempts to obscure as much of the input as possible while still leaving enough to make an accurate classification. 
In short, this model achieves competitive accuracy while having the ability to provide explanations directly by showing the rationales on which the prediction is based on.

For the generator and predictor, we use independent, pretrained BERT \citep{devlin2018bert} instances distributed by HuggingFace \citep{wolf2019huggingface}. 
We use Pytorch Lightning\footnote{\url{https://www.pytorchlightning.ai/}.} for fine-tuning. We use Gumbel Softmax \cite{jang2016categorical} to enforce a binary constraint on the predicted rationale, such that a token is either fully included or fully excluded from the input. As an implementation detail, we find it highly useful to pre-fine-tune the predictor layer on the full (un-masked) input before further training it in tandem with the generator.

\begin{figure}[t]
    \centering
    \begin{subfigure}[t]{0.35\textwidth}
        \centering
        \includegraphics[width=\textwidth]{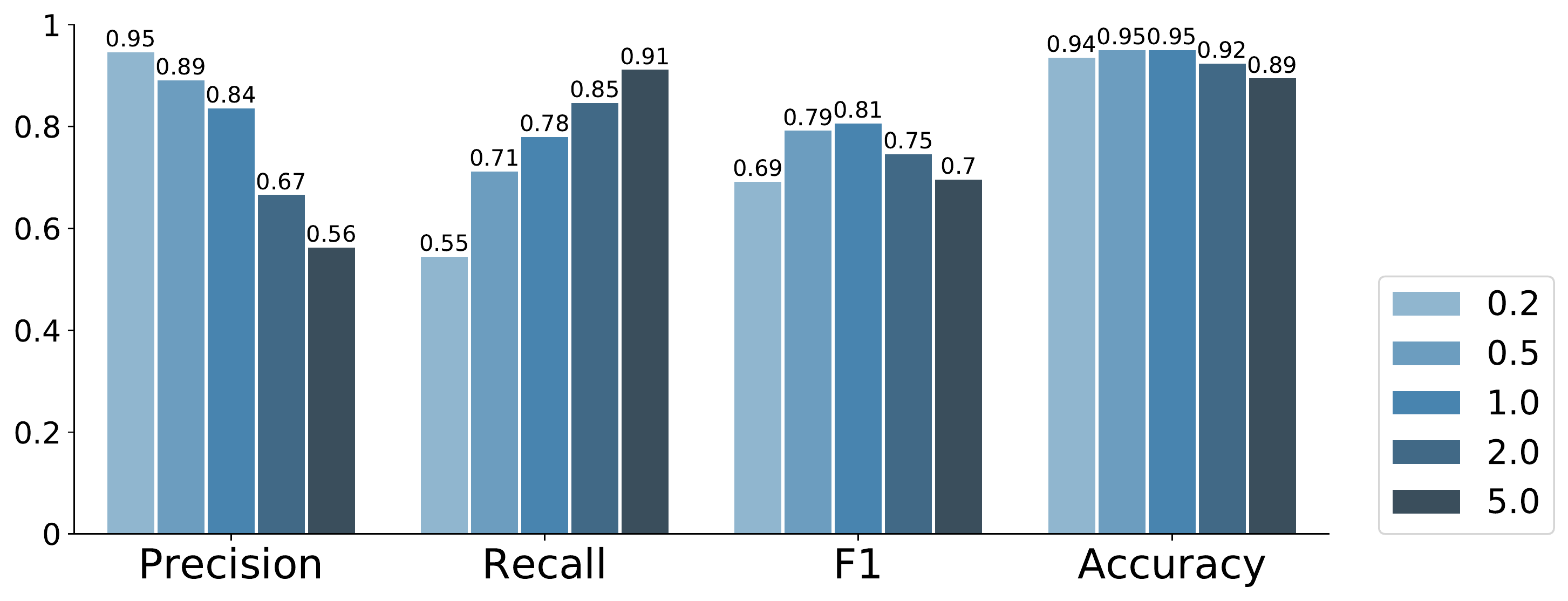}
        \caption{WikiAttack (\ind)}
        \label{fig:performance_wiki}
    \end{subfigure}
    \begin{subfigure}[t]{0.35\textwidth}
        \centering
        \includegraphics[width=\textwidth]{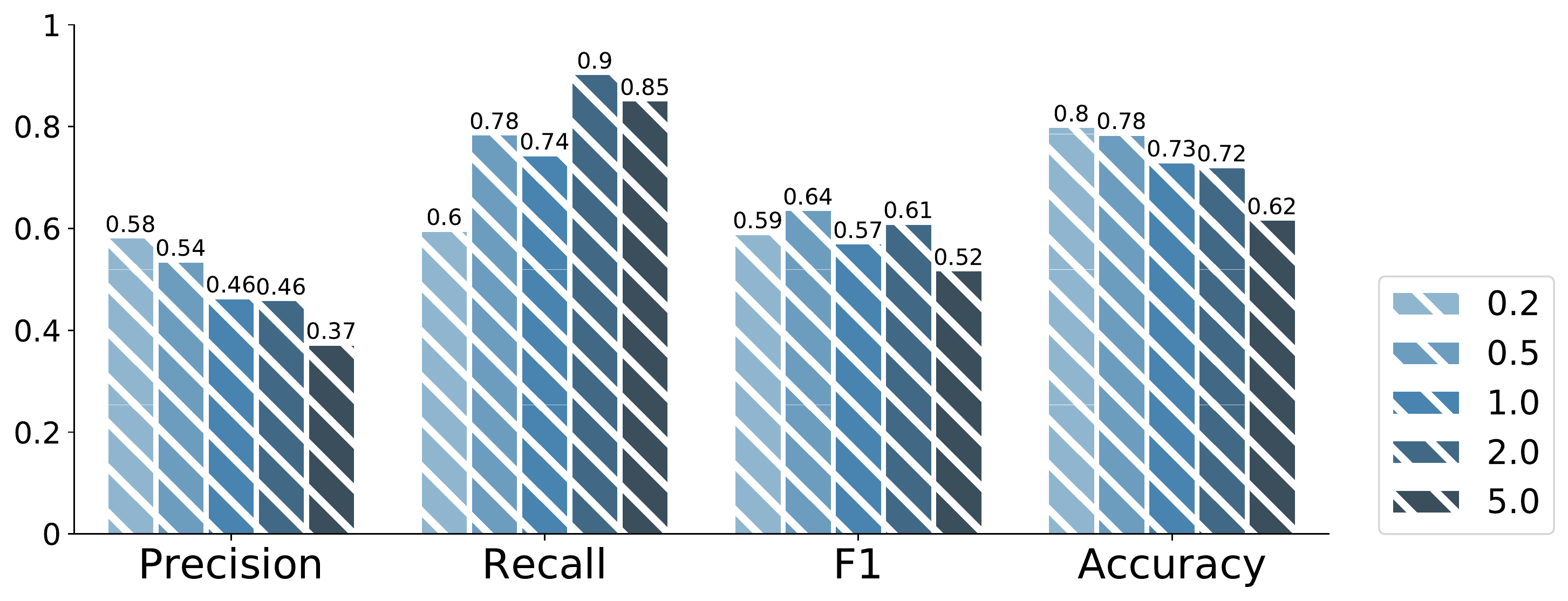}
        \caption{Reddit (\ood)}
        \label{fig:performance_reddit}
    \end{subfigure}
    \caption{An overview of model performance with different hyper-parameters. The hyperparameter shows the relative weight of recall vs. precision. 
    We choose the model with 0.5 because it achieves competitive performance both \ind and \ood. 
    Throughout this paper, we use solid bars for WikiAttack (\ind) and bars with white slashes for Reddit (\ood).
    }
    \Description{We experiment with different parameters to trade-off precision and recall. We chose the model with 0.5 as the hyperparameter value as it achieves competive performance in both datasets.}
    \label{fig:model_performance}
\end{figure}

Because our task emphasizes precision over accuracy,
we experiment with different parameters to trade-off precision and recall.
\figref{fig:model_performance} shows model performance both \ind and \ood with different parameters.
We observe a clear performance drop \ood (e.g., F1 drops from about $\sim$0.8 to $\sim$0.6)\added{, which validated our choice of Reddit as an \ood scenario}.
In our experiments, we choose the model with recall weight 0.5 (the second bar).
Note that participants did not have access to this performance data because our goal is to simulate the scenario where moderators work with a model developed on an existing dataset.
It is up to the moderators to figure out how well the model performs and when to trust or distrust the model.

\begin{table*}[t]
    \small
    \centering
    \begin{tabular}{p{0.2\textwidth}p{0.75\textwidth}}
    \toprule
        \multicolumn{2}{c}{WikiAttack} \\
        True positive & \definecolor{highlight}{RGB}{152,193,217}\sethlcolor{highlight}== Hello == \hl{Your}\hl{ }\hl{such}\hl{ }\hl{a}\hl{ }\hl{fucking}\hl{ }\hl{freak}\hl{ }\hl{mate}. Get a life. And get a facelift \hl{you}\hl{ }\hl{ugly}\hl{ }\hl{cunt}. \\
        
        True positive & \definecolor{highlight}{RGB}{152,193,217}\sethlcolor{highlight}== \hl{FUCKYOU} == wow,just wow \hl{james} delenoy. u \hl{are} a \hl{class} A \hl{faggot}. i could \hl{pwn}\hl{ }\hl{u}\hl{ }\hl{in}\hl{ }\hl{a}\hl{ }\hl{coma}. i dont give a \hl{fuck} if u block this account cuz we \hl{are} a \hl{freight} train from \hl{hell} and we will make more and \hl{come}\hl{ }\hl{at} u \hl{with} the \hl{ferocity}\hl{ }\hl{of}\hl{ }\hl{a}\hl{ }\hl{mama}\hl{ }\hl{bear}. \hl{SUCK}\hl{ }\hl{OUR}\hl{ }\hl{GRUNDLES}\hl{}\hl{,}\hl{ }\hl{the}\hl{ }\hl{entire}\hl{ }\hl{waltmanist}\hl{ }\hl{community} \\
        
        False positive & \definecolor{highlight}{RGB}{152,193,217}\sethlcolor{highlight}::Haha, \hl{I}\hl{ }\hl{fucking}\hl{ }\hl{pissed}\hl{ }\hl{myself} reading this ') Thanks, i needed that after 3 hours \hl{trecking} around in warrington with the \hl{worst} cold ever. I mean it, thanks )) \\
        \midrule
        \multicolumn{2}{c}{Reddit} \\
        True positive & \definecolor{highlight}{RGB}{152,193,217}\sethlcolor{highlight}\hl{Crawl} back to \hl{you} r/politics safe space. I'm \hl{sick} of \hl{fucking}\hl{ }\hl{faggot} liberals coming into communities where they aren't welcome and \hl{trying} to \hl{ruin} them from the inside out. You're on KiA, so you already know how that works\hl{.}\hl{ }\hl{Fuck}\hl{ }\hl{you} \\ 
        
        True positive & \definecolor{highlight}{RGB}{152,193,217}\sethlcolor{highlight}Ah yes, coming from \hl{the}\hl{ }\hl{idiot}\hl{ }\hl{that} calls people a \hl{retard} when he disagrees. \hl{Get} the \hl{fuck}\hl{ }\hl{out}\hl{ }\hl{of}\hl{ }\hl{your} internet bubble and look around \hl{you}\hl{ }\hl{irl} for a moment. \\
        
        False positive & \definecolor{highlight}{RGB}{152,193,217}\sethlcolor{highlight}Someone in chat baited me into saying "paki". I got banned. I'\hl{m}\hl{ }\hl{fucking} Pakistani and that word has never been "offensive" to anyone. I've used it for 14 years myself. I had 1k hours in that game but it's staying uninstalled \\
    \bottomrule
    \end{tabular}
    \caption{Comments that are predicted toxic from WikiAttack and Reddit with their identified rationales by our model.}
    \label{tb:model_predictions_comments}
\end{table*}

This model can achieve BERT-like accuracy while being able to precisely and parsimoniously identify the rationale responsible for its prediction. \tabref{tb:model_predictions_comments} presents example rationales for comments that are predicted toxic, both correctly and incorrectly. 
We find qualitatively that the model produces sensible rationales in this application. \added{While it identifies some surprising tokens as toxic such as ``you'', it does}
succeed in learning that the primary evidence of non-toxicity is a lack of toxic tokens: it retains only 2\% of tokens on average for predicted-nontoxic comments, versus 15\% for predicted-toxic comments. 
\added{Note that this explanation method has attracted some criticism for producing rationales that don't necessarily align with human reasoning \citep{zheng_irrationality_2021}, but it has the advantage of producing rationales that are, by construction, sufficient (in the logical sense) for the model's prediction.
Generating high-quality explanations is an active area of research, and our paradigm of conditional delegation can be used for any model of choice.
} 

The rationale produced by this model is a form of \textbf{local explanations}, identifying important words in each prediction. Our experiment also includes %
\textbf{global explanations}, which convey an overview of the model behavior across all inputs. We generate these global explanations by identifying the tokens that occur most frequently in the rationales of the model on the \ind and \ood data respectively.
We display these top-15 most frequent rationale tokens (\tabref{tb:frequent_words}) as the ``global explanations''.
We can immediately observe differences between Reddit and WikiAttack: ``cunt'' and ``retard'' are not common in rationales in WikiAttack but are among the top five on Reddit.

\begin{table*}[t]
    \centering
    \small
    \begin{tabular}{ll}
        \toprule
        WikiAttack & you, fuck, your, suck, die, shit, nigger, faggot, cock, my, bitch, stupid, go, ass, i\\ 
        Reddit & you, fuck, cunt, retard, shit, your, stupid, faggot, bitch, hate, she, i, guy, her, idiot\\
        \bottomrule
    \end{tabular}
    \caption{Words that are most frequently used in rationales.}
    \label{tb:frequent_words}
\end{table*}

\begin{figure}
    \centering
    \includegraphics[width=0.3\textwidth]{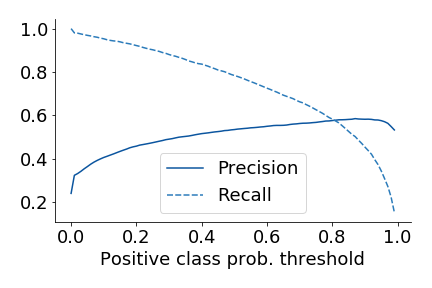}
    \caption{\added{Precision-recall plot of model on Reddit dataset (out-of-distribution). Even at high positive class probability thresholds, precision remains low.} 
    }
    \Description{The precision-recall plot of model on Reddit dataset shows that even with high positive class probability threshold of 0.93, the model only reaches 0.58 precision.}
    \label{fig:reddit_precision_recall_curve}
\end{figure}

\added{To provide context for our later findings, it is difficult to produce precise classification out-of-distribution. \figref{fig:reddit_precision_recall_curve} shows that even with a high positive class probability threshold (0.93), the model only climbs to 0.58 precision. Thus, even a very conservative application of this model is still producing 2 false positives for every 3 true positives--impractical for use by real moderators, and something we would like to be able to improve on via conditional delegation.}

\subsection{Performance of Individual Words}
\label{sec:analysis}

The goal of this work is to explore human-created keywords rule for conditional delegation to AI, such as  ``if a comment contains word X and is predicted toxic, the model will be trusted to report the comment for moderation action''.
In comparison,
with a manual rule-based approach
(e.g., the current AutoModerator system used by content moderators of Reddit), such a rule takes a form like ``if a comment contains word X, that comment will be reported''. 
Our hypothesis is that with the proper choice of rules, humans can produce a system which is more precise than either the manual rule-based approach or the model working alone.

We perform preliminary analysis to characterize the scope of the potential improvement and to contextualize our experimental results.
A crucial question in motivating our approach is whether there exist trustworthy regions of the model, i.e., are there certain words that occur systematically in comments where the model achieves high precision.

\begin{figure}
    \centering
    \begin{subfigure}[t]{0.35\textwidth}
        \centering
        \includegraphics[width=\textwidth]{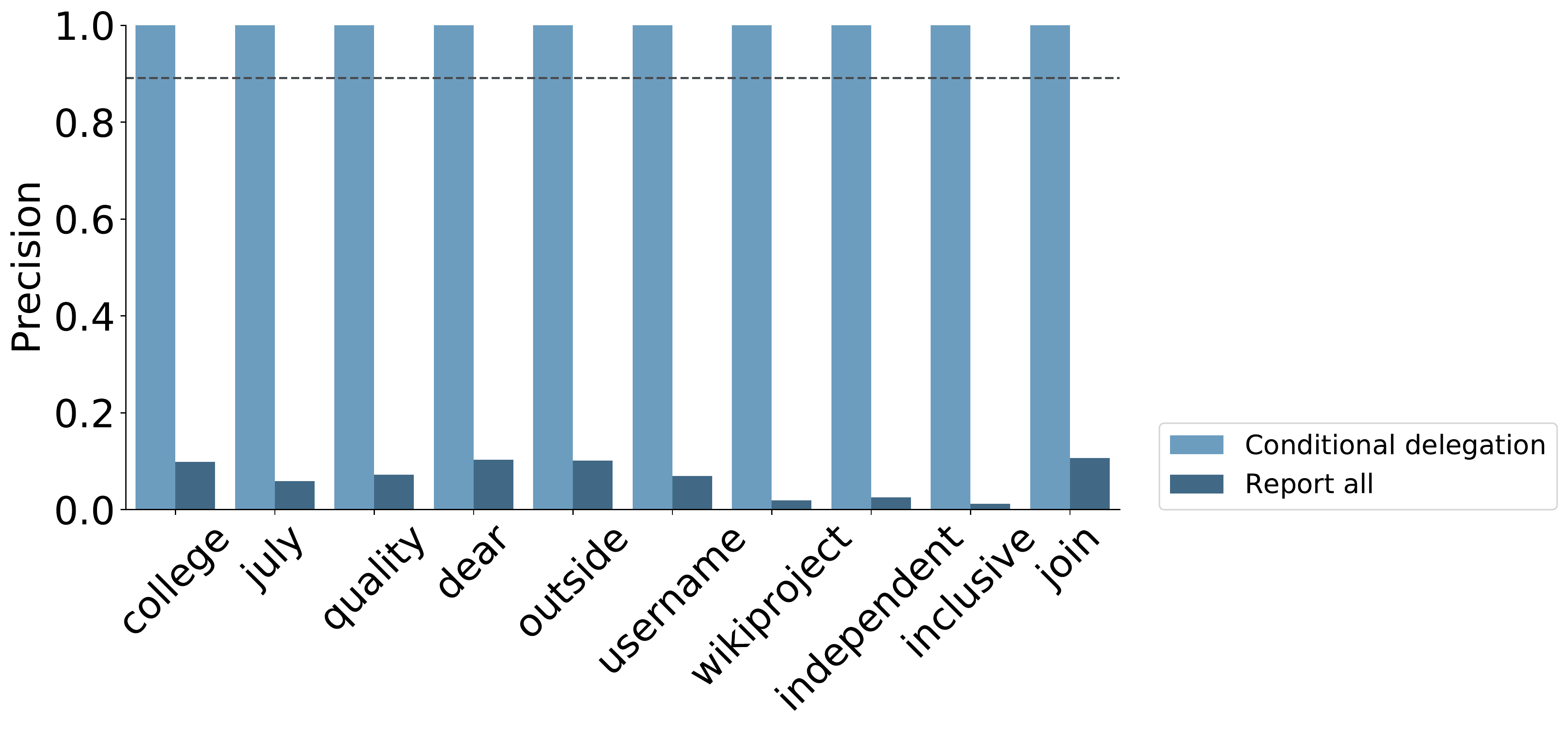}
        \caption{WikiAttack.}
        \label{fig:top_word_wiki}
    \end{subfigure}
    \begin{subfigure}[t]{0.35\textwidth}
        \centering
        \includegraphics[width=\textwidth]{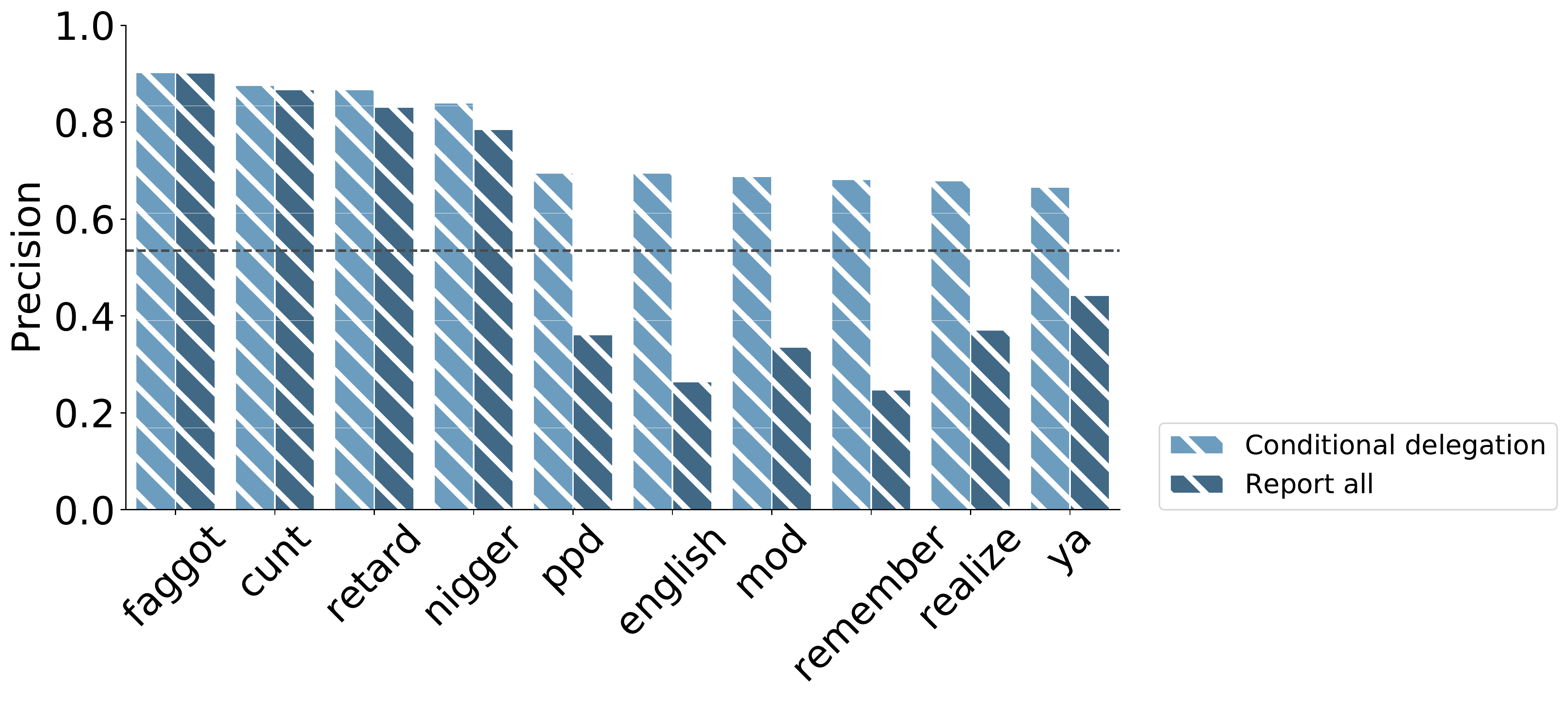}
        \caption{Reddit.}
        \label{fig:top_word_reddit}
    \end{subfigure}
    \caption{Words with top precision on WikiAttack (\ind) and Reddit (\ood). ``Conditional delegation'' shows precision among comments with the word based on model predictions, while ``Report all'' shows this measure if we consider a comment toxic as long as it contains the word (manual rule-based approach). Dashed lines show model precision on all comments (i.e., the precision of the model working alone). In both cases, 
    all the top 10 words lead to greater precision than the model working alone. 
    }
    \Description{This figure shows the top 10 words with highest precision as conditional delegation rules.}
    \label{fig:top_word_precision}
\end{figure}

First, we compute the precision of conditional delegation %
for all words that show up in at least 100 comments. 
\figref{fig:top_word_precision} shows the 10 words with the highest precision as conditional delegation rules (i.e., based on model predictions for all comments containing them) on WikiAttack and Reddit respectively. Conditional delegation based on these words leads to greater precision than the model working alone (dashed lines),
suggesting that users can improve the precision of the model by identifying these words for conditional delegation.
In addition, we compare that with the precision of using the word as a ``report all'' rule as with manual rule-based approach, by considering all comments containing the word as toxic. 
We can see generally, for these words with top precision, trusting the model leads to higher precision than ``report all'', both \ind and \ood. 
However, the difference is much smaller for Reddit (\ood). In particular, ``faggot'', ``cunt'', ``retard'', and ``nigger'' achieve very high precision on this dataset even if one simply reports all comments that contain any of those words.
These results indicate that conditional delegation can outperform both the manual rule-based approach and the model working alone if users are able to make good choices of keywords rules. 

\begin{figure}
    \centering
    \begin{subfigure}[t]{0.35\textwidth}
        \centering
        \includegraphics[width=\textwidth]{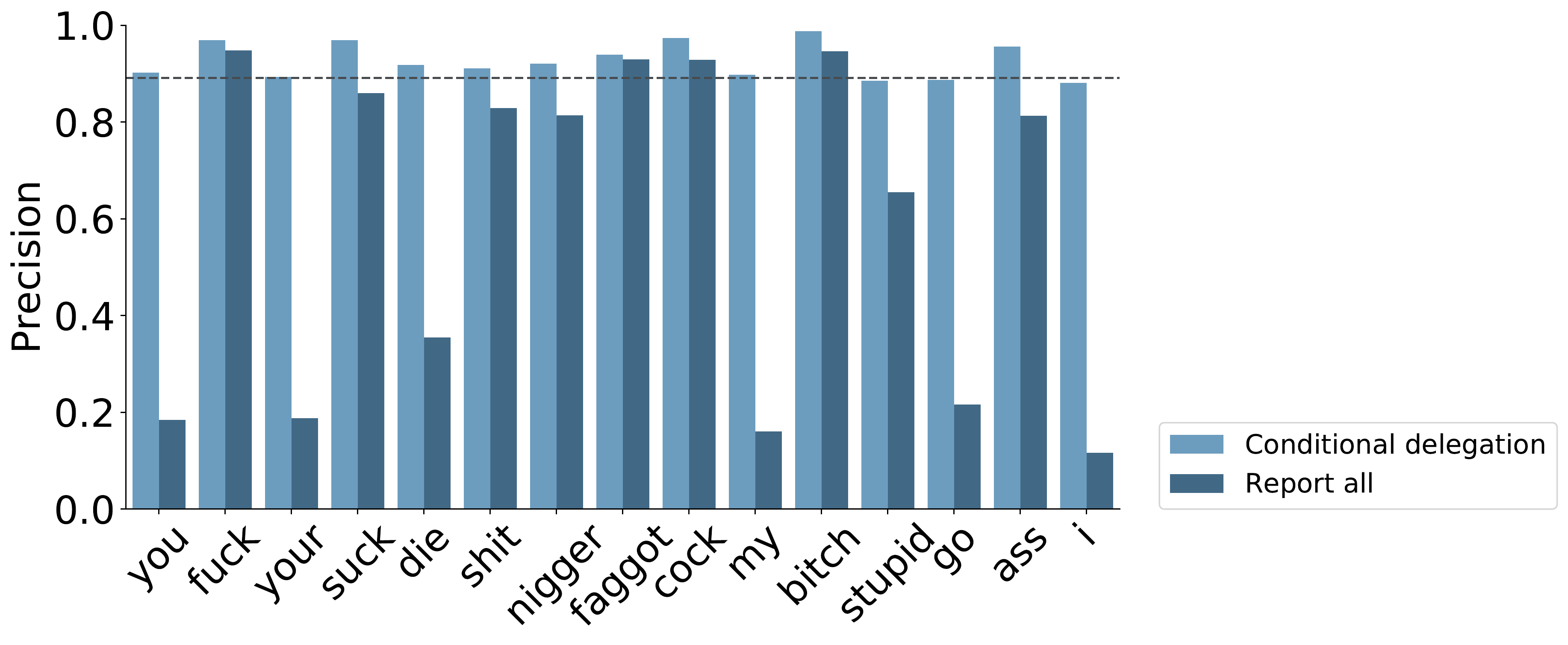}
        \caption{WikiAttack.}
        \label{fig:frequent_word_wiki}
    \end{subfigure}
    \begin{subfigure}[t]{0.35\textwidth}
        \centering
        \includegraphics[width=\textwidth]{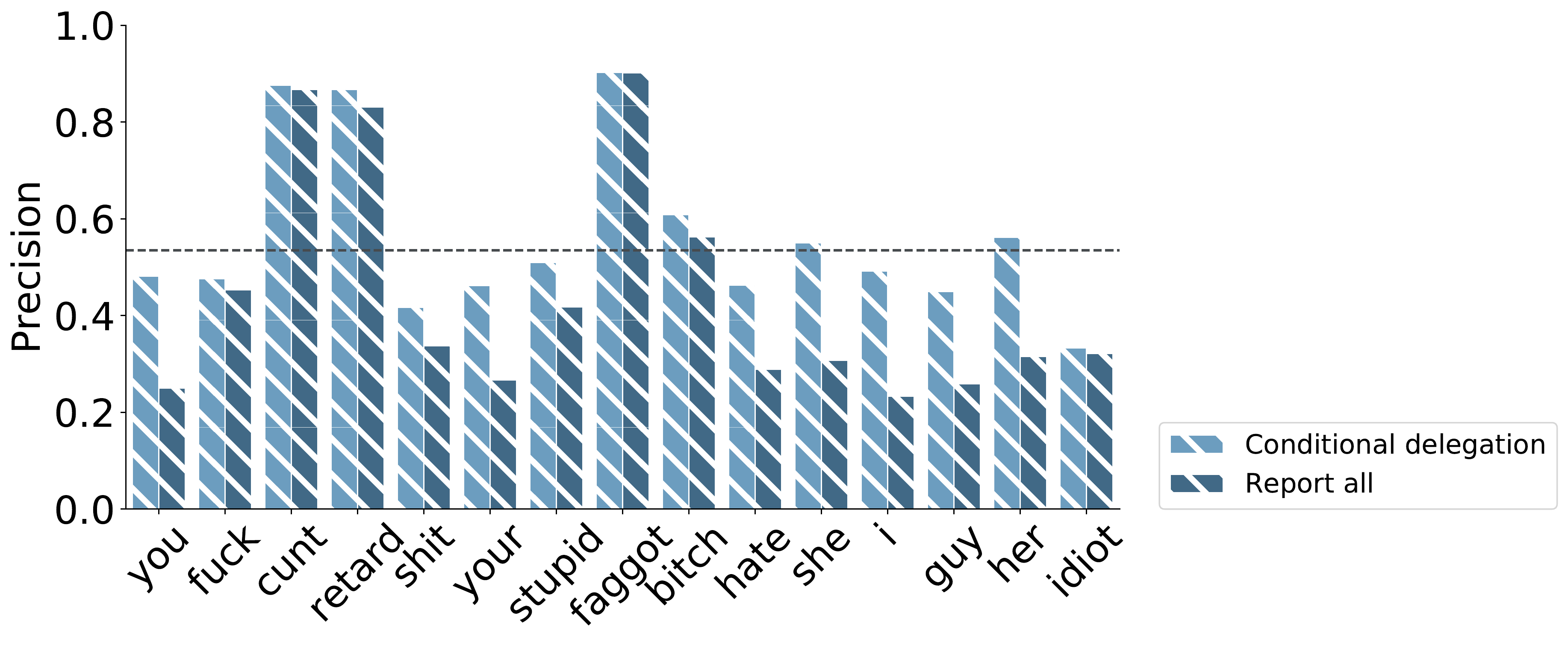}
        \caption{Reddit.}
        \label{fig:frequent_word_reddit}
    \end{subfigure}
    \caption{Precision for words that show up most frequently in rationales on WikiAttack (\ind) and Reddit (\ood) (ordered by frequency). ``Conditional delegation'' shows precision among comments with the word based on model predictions, while ``Report all'' shows this measure if we consider a comment toxic as long as it contains the word (the manual rule-based approach). Dashed lines show model precision on all comments (i.e., the precision of the model working alone). 
    }
    \Description{This figure shows that the majority of global explanations on WikiAttack achieve greater precision than the model working alone.}
    \label{fig:frequent_word_precision}
\end{figure}

\begin{figure}[t]
    \centering
    \begin{subfigure}[t]{0.35\textwidth}
        \centering
        \includegraphics[width=\textwidth]{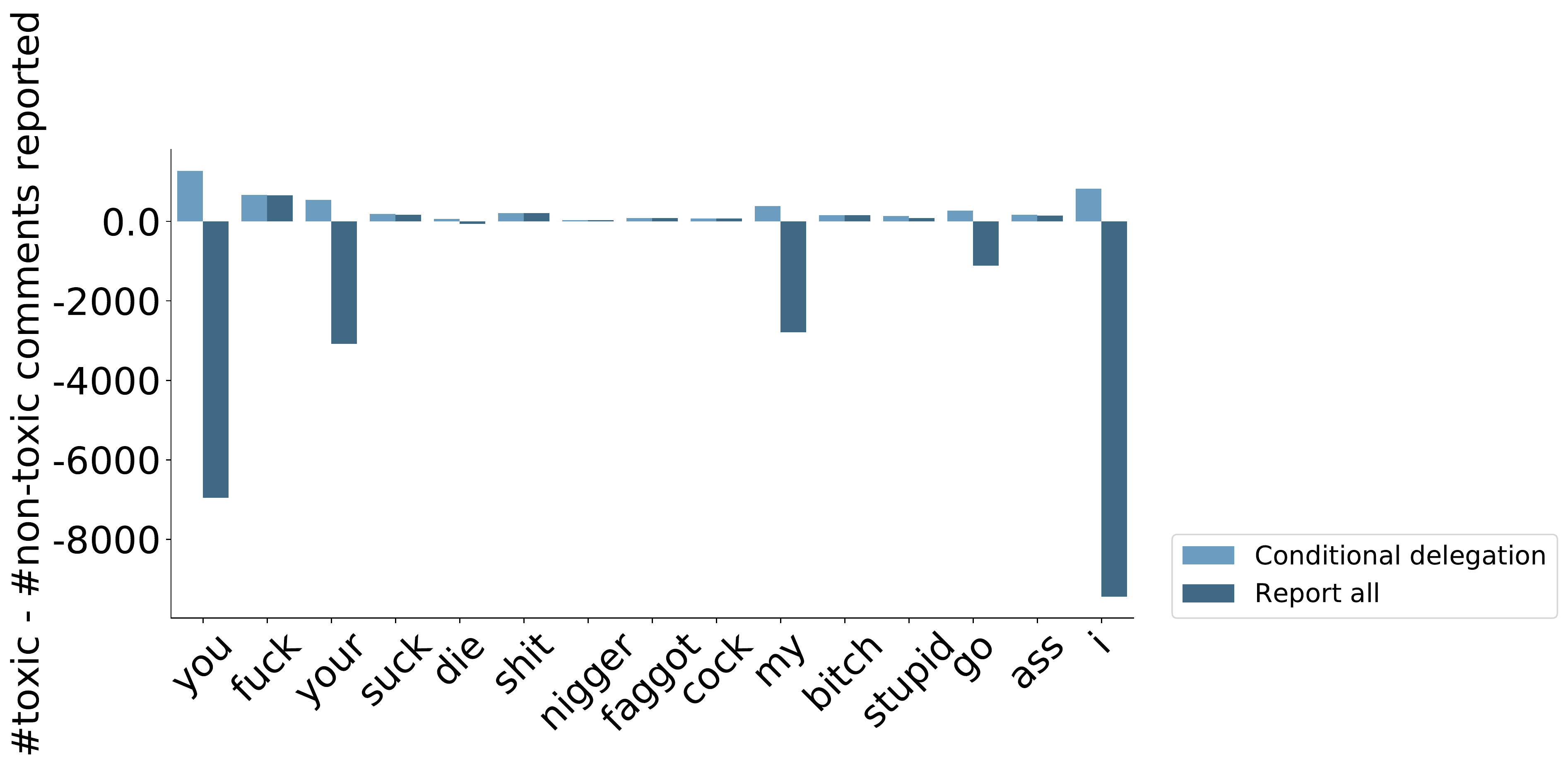}
        \caption{WikiAttack}
        \label{fig:reward_global_wiki}
    \end{subfigure}
    \begin{subfigure}[t]{0.35\textwidth}
        \centering
        \includegraphics[width=\textwidth]{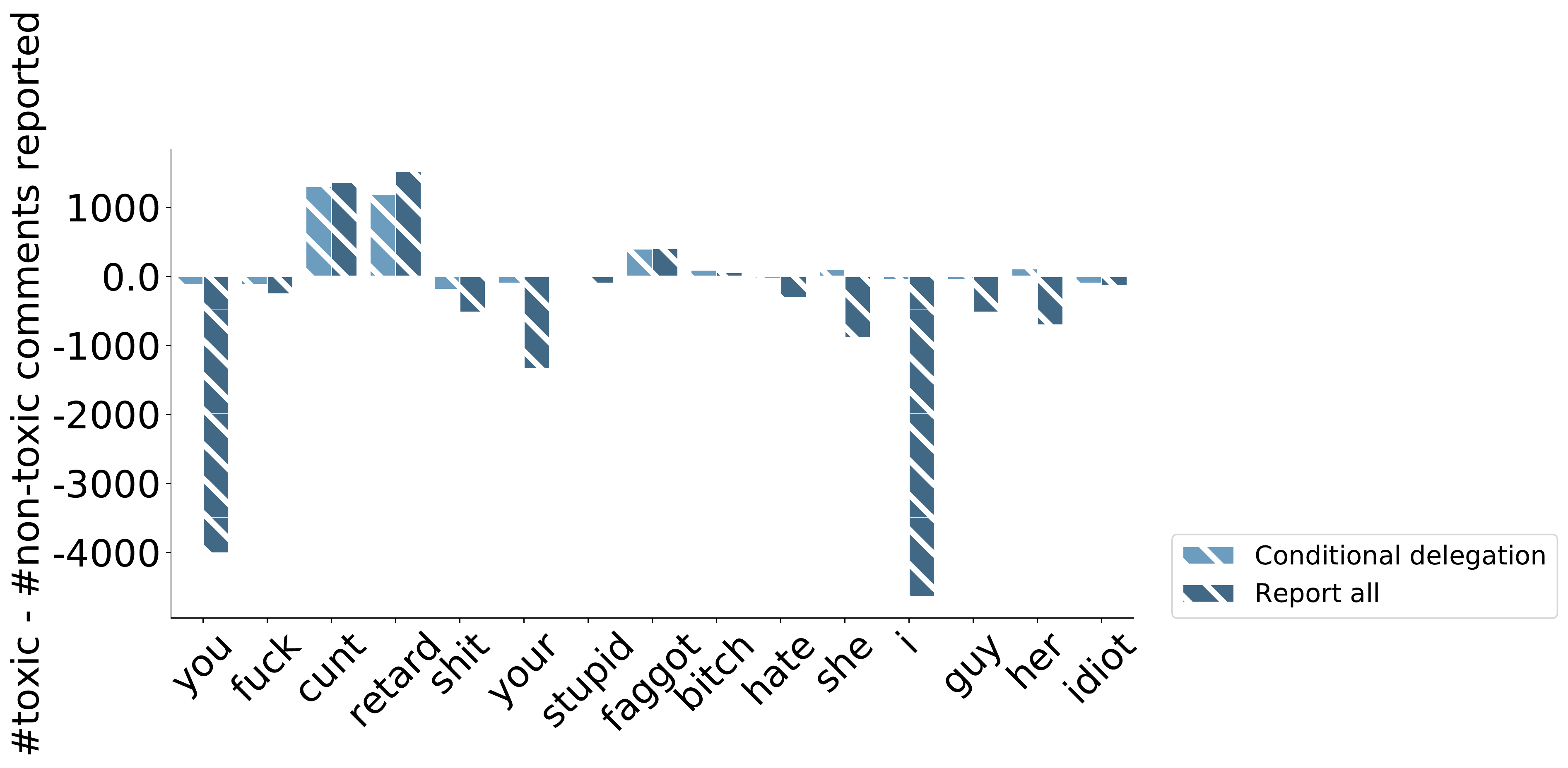}
        \caption{Reddit}
        \label{fig:reward_global_reddit}
    \end{subfigure}
    \caption{Reward (number of reported toxic comments - number of reported non-toxic comments), a measure combining precision and coverage, for words that show up most frequently in rationales on WikiAttack (\ind) and Reddit (\ood) (ordered by frequency). ``Conditional delegation'' shows reward for comments with the word based on model predictions, while ``Report all'' shows this measure if we consider a comment toxic as long as it contains a word (the manual rule-based approach).}
    \Description{This figure shows the reward on both datasets. On WikiAttack, most global explanations lead to positive rewards while on Reddit, rewards are dominated by ``retard'' and ``cunt'' due to high coverage.}
    \label{fig:global_explanations_reward}
\end{figure}

Next, we examine the precision of the words that are most frequent in rationales (\tabref{tb:frequent_words}, to be shown as global explanations).
\figref{fig:frequent_word_precision} shows that on WikiAttack, the majority of global explanations achieve greater precision than the model working alone.
However, on Reddit, this is true only for six words (``cunt'', ``retard'', ``faggot'', ``bitch'', ``she'', ``her''), indicating the challenge of creating good rules for \ood data.

\figref{fig:global_explanations_reward} further shows (the number of reported toxic comments - the number of reported non-toxic comments) if that word is chosen as a rule. This measure reflects both the coverage and precision of a keyword rule.
We refer to this measure as {\em reward} because it is used as an incentive in our human subject experiments, to be introduced later.
On WikiAttack, most global explanations lead to positive rewards.
We also observe the clear advantage of using conditional delegation over ``report all''.
This advantage becomes smaller on Reddit and disappears for ``retard'' and ``cunt'' because these two words have great precision and coverage by simply reporting all comments with them.
In fact, rewards on Reddit are dominated by ``retard'' and ``cunt'' due to their high coverage. \added{A user could achieve a quite high reward (and outperform the model) simply by reporting all comments with either of these two words.}

In summary, there are specific words that delineate trustworthy regions of the AI model, \added{and even certain words (``retard'', ``cunt'') where simply reporting all comments containing these words would be more effective in terms of reward than delegating such comments to the model, particularly in the \ood setting. However, we are able to recognize such words with the benefit of a fully-labeled dataset (i.e., oracle access) --- discovering them in a real-world setting could be very challenging. We explore this challenge in a rigorous human subject experiment in the next section.}
\section{Experimental Design}
\label{sec:experimental_design}

Equipped with the model, one goal of this study is to design interfaces with different support features to enable people to come up with effective keyword-based rules for conditional delegation. We then examine the effect of these support features through a human-subject experiment. In this section, we start by introducing different types of support features that we consider and then explain the study procedure.

\subsection{Experimental Conditions and Interface Design}
\label{sec:interface_design}

In order to enable people to create keyword-based rules for conditional delegation and observe model behaviors with them, the basic function of our tool is to search for a keyword and browse comments that contain it. 
This allows users to determine whether a keyword would serve as a good rule.
For different experimental conditions, our design space mainly involves what information we provide when returning the search results.

\para{Experimental conditions.} As discussed in \secref{sec:method}, in addition to predicting whether a comment is toxic or not, our model can provide local explanations (i.e., which words are used as rationales for the prediction) as well as global explanations (i.e., most frequent words that show up in the rationales).
Therefore, we consider the following four conditions:

\begin{itemize}[leftmargin=*]
    \item \textbf{Predicted labels.} Predicted labels are shown along with the searched comments.
    \item \textbf{Predicted labels + local explanations.} In addition to predicted label for each comment, we highlight rationales, i.e., words in the comment the model uses to determine toxicity for comments that are predicted toxic.   We refer to this condition as ``\textit{local explanations}''.

    \item \textbf{Predicted labels + local explanations + global explanations.} Participants have access to all of the features in the previous condition and are also provided a list of words that the model typically uses in determining comment toxicity (\tabref{tb:frequent_words}). 
    We refer to this condition as ``global explanations''.
    \item \textbf{Manual condition.} The final condition is designed to simulate the current state of AutoModerator, where moderators come up with ``report all'' rules.
    We create a consistent interface where participants have the ability to search comments and browse returned results to assess whether they are indeed toxic, instead of whether the model prediction is precise. Participants do not have access to any model-related information.
\end{itemize}

In the rest of this paper, we refer to these conditions as {\em experimental conditions} and WikiAttack vs. Reddit as {\em distribution types}.

\begin{figure*}[t]
    \centering
    \includegraphics[width=0.63\textwidth]{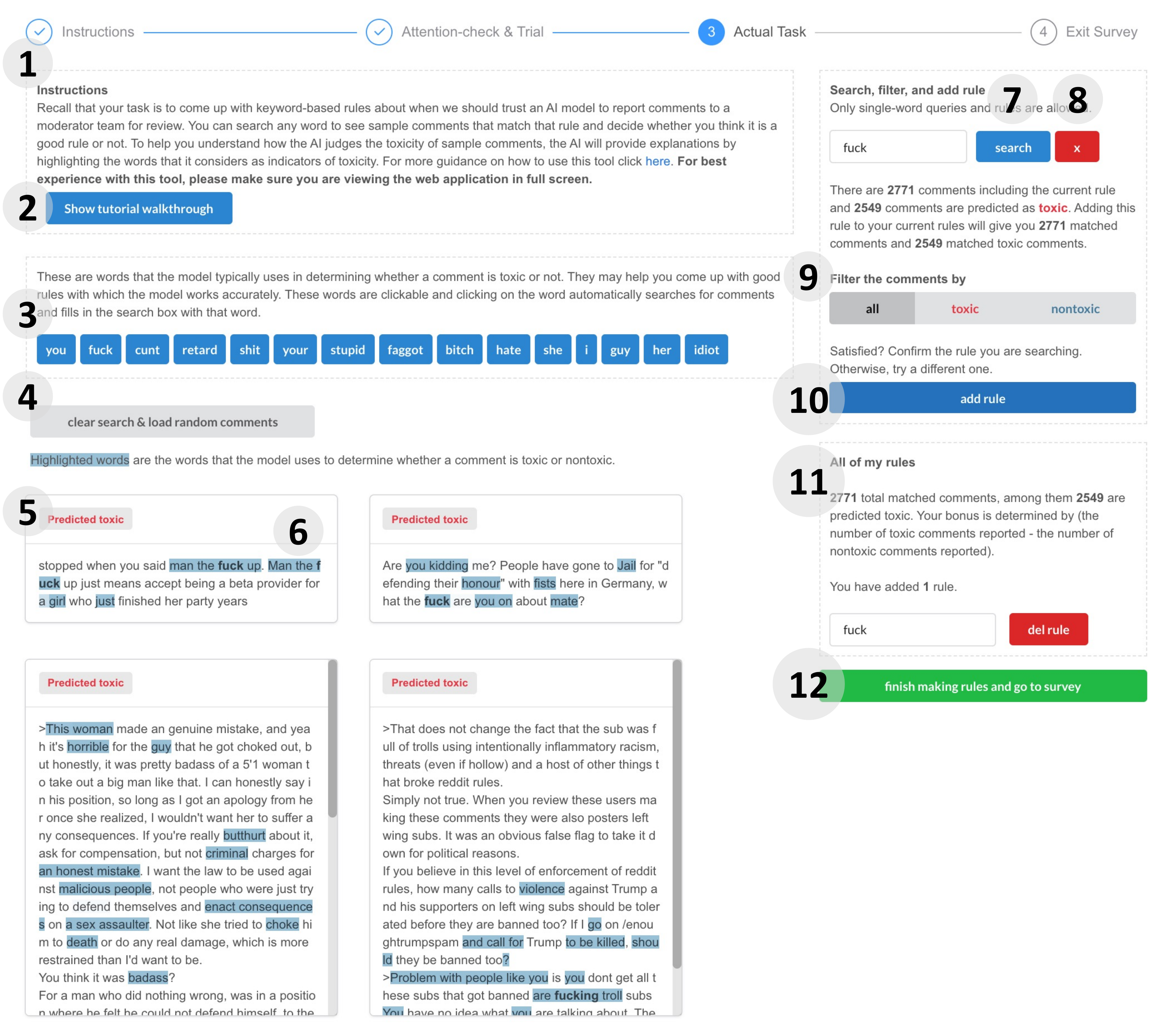}
    \caption{Interface for ``Predicted label + local explanations + global explanations''. We use this interface to go through the design of all delegation support features.}
    \label{fig:interface_predicted_label_local_global}
    \Description{This interface includes all delegation support features.}
\end{figure*}

\begin{figure*}
    \centering
    \begin{subfigure}[t]{0.31\textwidth}                                                        
        \centering
        \includegraphics[width=\textwidth]{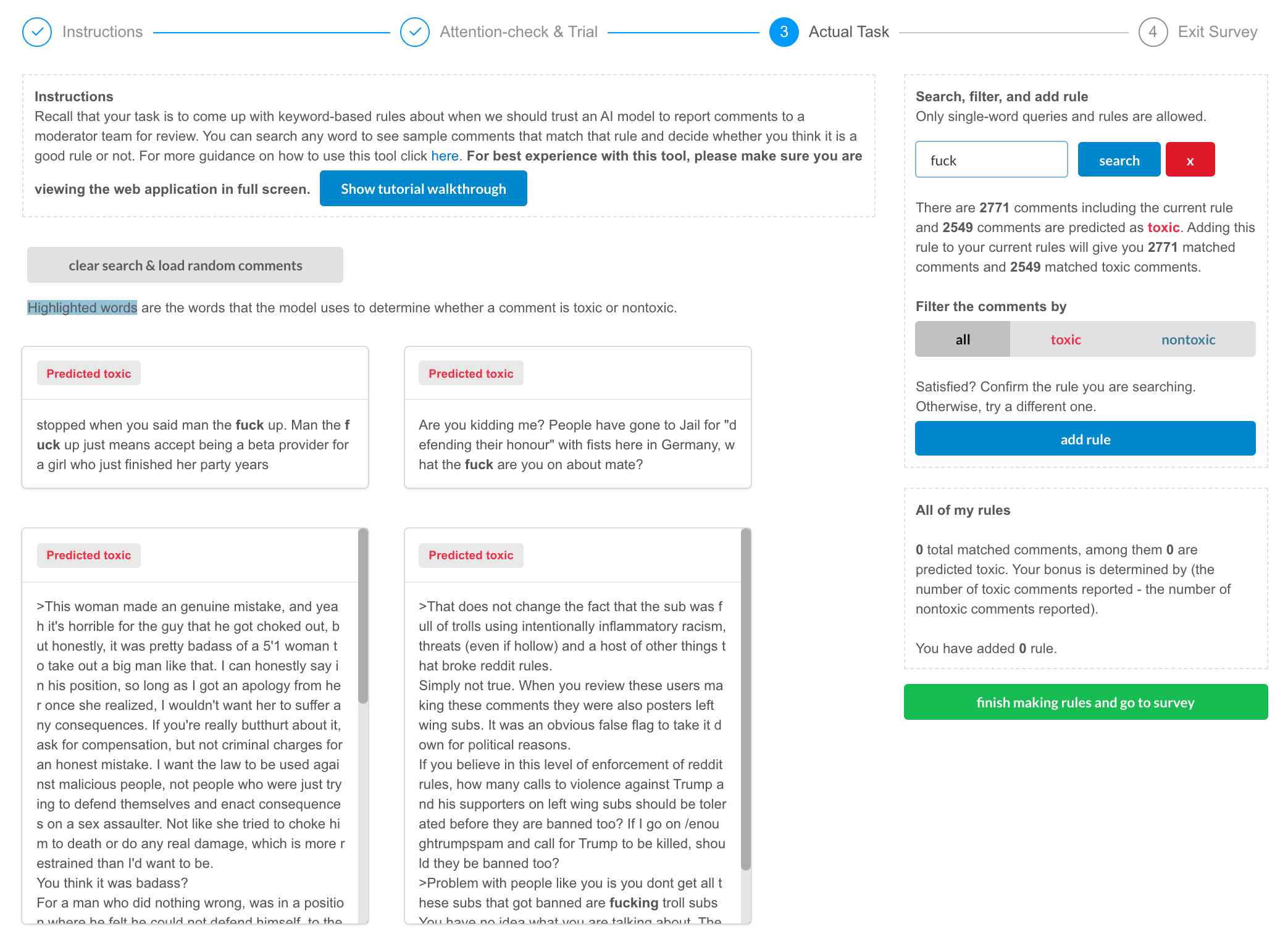}
        \caption{Predicted labels condition}
        \label{fig:interface_predicted_label}
    \end{subfigure}
    \begin{subfigure}[t]{0.35\textwidth}
        \centering
        \includegraphics[width=0.9\textwidth]{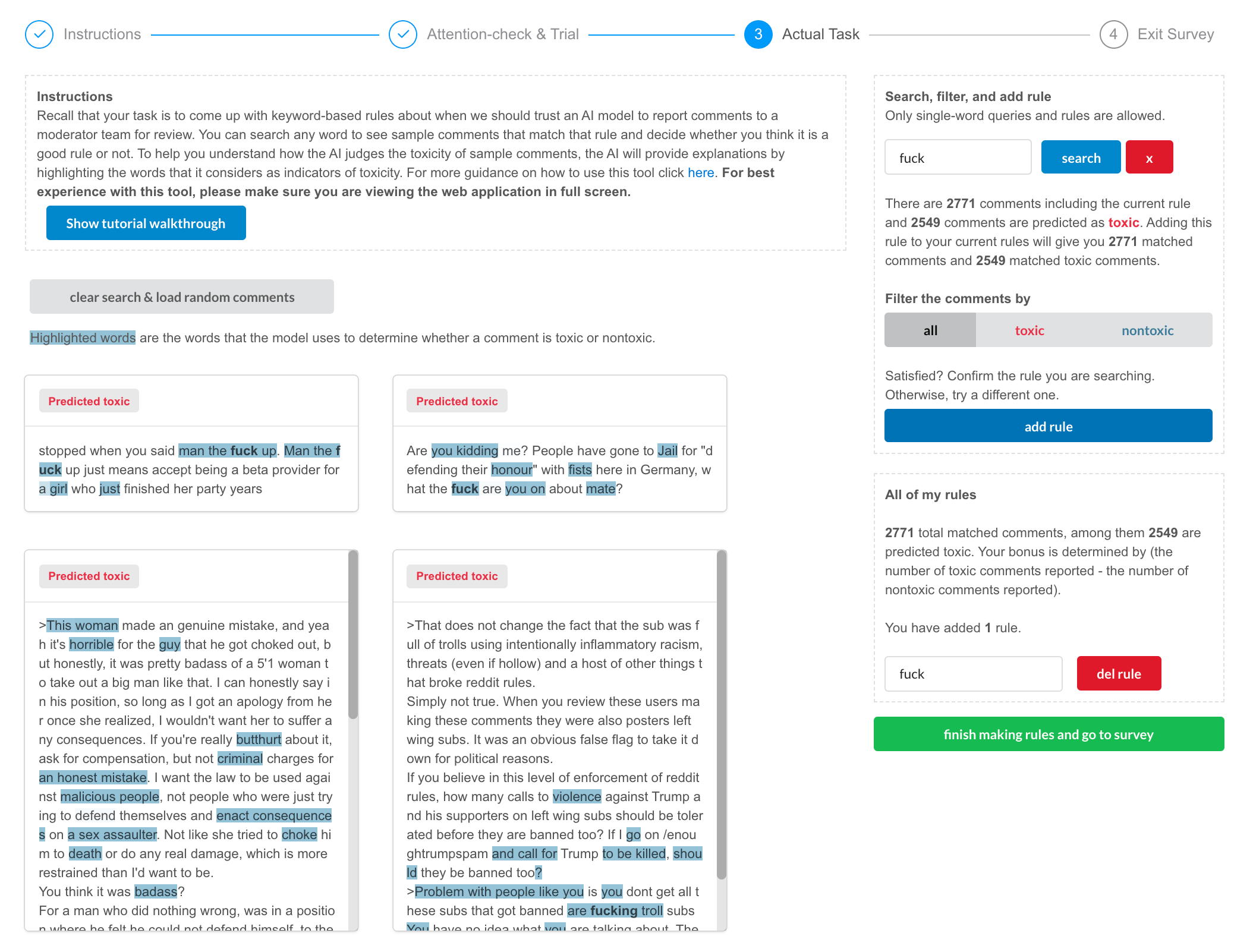}
        \caption{Predicted labels + local explanations condition}
        \label{fig:interface_predicted_label_local}
    \end{subfigure} 
    \begin{subfigure}[t]{0.31\textwidth}
        \centering
        \includegraphics[width=\textwidth]{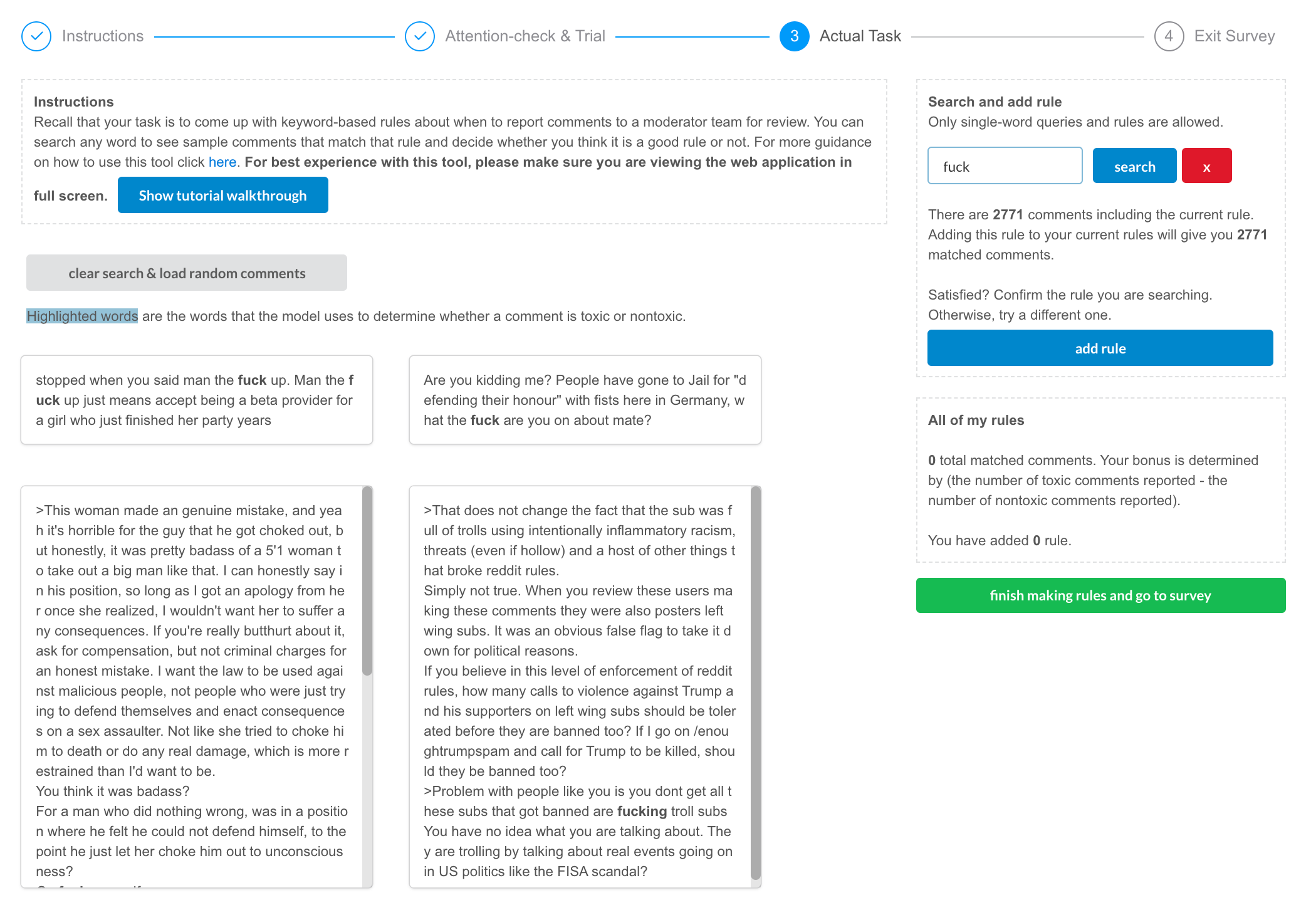}
        \caption{Manual condition}
        \label{fig:interface_manual}
    \end{subfigure}
    \caption{Interfaces for the other three experimental conditions.}
    \Description{These interfaces include some of the delegation support features.}
    \label{fig:interface}
\end{figure*}

\para{Interface design.} 
We start by introducing the interface for ``Predicted label + local explanations + global explanations'', which includes all possible components of the other conditions (see \figref{fig:interface_predicted_label_local_global}).
The widgets in the interface are arranged in two columns,
where instructions and comments are displayed on the left, 
while the search bar and the current set of rules are on the right. 
The instructions box (\figref{fig:interface_predicted_label_local_global}(1)) reminds participants of the task and provides more information about the interface to ensure that they can fully leverage the tool's features.
Global explanations are shown below the instructions box (\figref{fig:interface_predicted_label_local_global}(3)).
When the participant clicks on a rule that is represented by a button, it automatically searches comments with the respective keyword-based rule.
In addition to searching particular words, we also allow users to load random comments\added{, which can be used to explore the data} (\figref{fig:interface_predicted_label_local_global}(4)). 
Upon a query or loading random comments,
the comments are displayed as \textit{cards} below the \textit{load random comments} button.
Depending on the condition, a comment could have a predicted label (\figref{fig:interface_predicted_label_local_global}(5)) 
and rationales could be highlighted (\figref{fig:interface_predicted_label_local_global}(6)).
On the right side, the first two widgets are the \textit{search bar} (\figref{fig:interface_predicted_label_local_global}(7)) and \textit{clear} button (\figref{fig:interface_predicted_label_local_global}(8)).
The participant enters keyword-based rules and then comments with the respective rule are shown on the left, as described in the previous paragraph.
Participants can filter comments by their predicted label (\figref{fig:interface_predicted_label_local_global}(9)).
By default, both predicted toxic and nontoxic comments are shown.
When the participant is satisfied with the rule, they may click on the \textit{add rule} button (\figref{fig:interface_predicted_label_local_global}(10)) to add the rule to their list.
All of the participant's rules are displayed in the component below the \textit{add rule} button.
We also display their total matched comments and predicted toxic matched comments (\figref{fig:interface_predicted_label_local_global}(11)).
Finally, participants may click on the \textit{finish making rules and go to survey} button to submit their rules and proceed to the exit survey (\figref{fig:interface_predicted_label_local_global}(12)).

\figref{fig:interface_predicted_label_local_global} shows interfaces for the other three conditions.
``Predicted labels + local explanations'' condition (\figref{fig:interface_predicted_label_local}) removes the global explanations (\figref{fig:interface_predicted_label_local_global}(3)) and is otherwise the same as ``Predicted labels + local explanations + global explanations''.
``Predicted labels'' condition (\figref{fig:interface_predicted_label}) further removes the highlights of rationales when returning search results.
``Manual'' condition (\figref{fig:interface_manual}) only has ``search'' and ``load random comments''.
We conducted pilot studies with lab mates at the same university and mechanical turkers to gather feedback and ensured that interactions were intuitive and the interface was easy to use.
With many widgets in the interface, it could potentially become 
overwhelming for the user.
To overcome that problem, we included a short tutorial walk-through (\figref{fig:interface_predicted_label_local_global}(2)) that explains each widget in sufficient details.
To minimize confusion, we also included a separate link in the instructions box (\figref{fig:interface_predicted_label_local_global}(1)) that enumerates how each widget should 
be used.

\subsection{Study Design and Procedure}
\label{sec:study}
We employ a between-subject design with the four conditions described above.
We aimed to obtain 0.95 power to detect an effect size of 0.1 in precision at the standard 0.01 alpha error probability using F-tests. 
As such, we recruited 30 participants for each condition, summing up to 120 participants each for \ind and \ood and 240 participants in total.
The study is approved by the University IRB.

\para{Task flow.} Participants went through four phases during the study: (1) consent and introduction; (2) interface tutorial; 3) the actual task; (3) exit survey.
During the introduction phase, participants were shown an accordion of instructions.
Due to the nature of toxic comments, they are warned that the task will contain profanity and hate speech and are cautioned to only proceed if they are okay with reading toxic comments.

After the instruction, participants are asked  three attention-check questions to allow us to filter out participants who did not pay attention to the instructions. They are given toxicity trial questions 
with four simple comments to label if they are toxic or nontoxic: (1) \textit{I had a really bad day today}. (2) \textit{I just made a fucking great steak!} (3) \textit{Go fuck yourself}. (4) \textit{I hope you die painfully tomorrow}.
These questions allow us to filter participants who finish this task hastily without paying effort and prepare them for toxicity judgment.
We remove participants whose accuracy is less than or equal to 50\% on these questions.
As a result, we filter 6 out of 240 in our analysis.

To familiarize participants with the interface, we include a tutorial walk-through when they first land on the page to instruct them on how to use each feature. We also include a link that featured more detailed instructions and a short demonstration video.
Participants are required to submit at least ten rules, and can then 
exit the task whenever they are satisfied with the set of rules created.

In the exit survey, we collected basic demographic information, their knowledge and familiarity of AI and content moderation, and subjective measures, to be introduced in the later section.

\para{Reward.} To motivate quality work, in addition to a base payment, we design a bonus incentive as follows: participants will be awarded \$0.10 for every 100 toxic comment their rules correctly reported, and penalize them \$0.10 for every 100 nontoxic comment their rules mistakenly reported (lower bounded by \$0 and upper bounded by \$2). This bonus thus rewards both precision--how likely comments under the rule (for the manual condition) or conditional delegation with the rule are correctly classified as toxic, and coverage--the quantity of comments covered by the rule. To make the calculation easy to understand for participants, this reward makes a simplified assumption that the cost of wrongly reported non-toxic comments (false positive) equals the benefit of correctly reported toxic comments (true positive). 

This reward mechanism is explained to participants, and we include one question 
in attention check to ensure they understood it. We also explicitly suggest that, to optimize for the reward,  their goal should be to come up with keywords that meet the following criteria: (1) that occur in a lot of comments; (2) with which the model makes accurate predictions on the comments, and (3) that are a diverse set so they may cover different kinds of toxic comments.

\para{Participant information.} 
We recruited participants from Amazon Mechanical Turk. \added{We note that while this recruiting choice may limit the generalizability of our results, social media content moderation is often performed by part-time volunteers whose expertise varies. Furthermore, we believe turkers are a sufficiently good sample for us to compare whether conditional delegation improves the content moderation outcomes over the baseline condition, and expert users are likely to further enhance the improvement pattern, if any. We encourage future work to further test the paradigm in realistic social media contexts.}

To ensure high quality responses, all participants 
satisfy the following criteria: (1) performed at least 1000 HITs; (2) approval of 99\% performed HITs in previous requesters; (3) reside in the United States; 4) has the adult content qualification since our task shows toxic comments.
The experiment follows a between-subject design therefore we do not allow any repeated participants.

There were 115 male, 116 female, 2 non binary, and 1 preferred not to answer.
52 participants are aged 18-29, 114 aged 30-39, 34 aged 40-49, 26 50-59, 7 aged over 59, and 1 I prefer not to answer.
Participants rated their knowledge on artificial intelligence (25 had no knowledge, 156 had little knowledge, 49 had some knowledge, 4 had a lot of knowledge), and social media content moderation (36 had no knowledge, 113 had little knowledge, 66 had some knowledge, 19 had a lot of knowledge) on five-point Likert scales.
Participants were paid an average wage of \$11.80 per hour. 

Overall, most turkers are satisfied with our task design and interface. 
Here are two quotes from their feedback: ``\textit{This was super intriguing. I had never participated in an activity like this before. It was hard coming up with bad words since they are not part of my vocabulary. It was interesting to see which words usually coincided with toxic subjects. Overall, very interesting project}.'' and ``\textit{It was interesting. I see now how difficult moderation can be for some sites}.''

\subsection{Evaluation Measures}

We consider three types of evaluation measures to cover efficacy, efficiency, and subjective perception.

\para{Efficacy.} As discussed in \secref{sec:introduction}, our main goal is to examine whether humans can improve the precision of the model with a good coverage via conditional delegation. 
We consider two precision-based measures: \textit{average precision} and \textit{union precision}. 
For the first three experimental conditions with delegation support features, 
average precision is formally defined as 
$$\frac{1}{|R|}\sum_{r \in R}\frac{|\{x \text{ is toxic}~\&~x \text{ contains } r~\&~x \text{ is predicted toxic}\}|}{|\{x \text{ contains } r~\&~x \text{ is predicted toxic}\}|},$$
where $R$ is the set of rules that participants choose and $x$ refers to a comment, whereas union precision is formally defined as 
$$\frac{|\{x \text{ is toxic}~\&~x \text{ contains any } r \in R~\&~x \text{ is predicted toxic} \}|}{|\{x \text{ contains any } r \in R~\&~x \text{ is predicted toxic}\}|}.$$
As the manual condition does not have a model, these two definitions become ${\scriptstyle \frac{1}{|R|}\sum_{r \in R}\frac{|\{x \text{ is toxic}~\&~x \text{ contains } r\}|}{|\{x \text{ contains } r\}|}}$ and ${\scriptstyle \frac{|\{x \text{ is toxic}~\&~x \text{ contains any } r \in R\}|}{|\{x \text{ contains any } r \in R\}|}}$.

The difference in the denominators highlights the role of conditional delegation, which only affect the comments that the model predicts as toxic.
It follows that the performance with conditional delegation is also determined by the model's base performance, i.e., how well the model can identify toxic comments.
Intuitively, average precision reflects the average quality of every single rule a person provides, while union precision measures the performance when using all rules from the person as a set, and can be skewed by the performance of higher-coverage rule in the set. Thus, one's ability to come up with both high-precision and high-coverage rules can lead to better union precision.

Finally, we consider the \textit{reward} participants received, as introduced in \secref{sec:study}, which measures the quantity difference between reported toxic comments (true positive) and reported non-toxic comments (false positive). This measure reflects both precision and coverage. 
This metric is highly volatile because a small number of keywords can achieve much higher rewards than others, especially \ood (e.g., ``retard'' and ``cunt'' on Reddit as shown in \secref{sec:method}).
We believe that precision is the more reliable measure of efficacy given that our participants tended to only choose about 10 rules.

\para{Engagement and efficiency.} We consider number of logged actions a participant took during the experiment task and number of rules they added as measurements for engagement.  13 types of unique actions were logged, including searching a rule, filtering comments by predicted labels (toxic and nontoxic), load random comments, get page comments, etc. 
We consider the number of actions more indicative of engagement, since participants can search for a rule without adding it. 
For efficiency, we consider total elapsed time and rules per minute.
Elapsed time starts from the moment participants enter the interactive interface until they click on ``finish making rules and go to survey'', in minutes.
Rules per minute is the number of rules added divided by elapsed time.
Since rules are the final product of the task, rules per minute is more indicative of efficiency.
\para{Subjective measures.} Finally, we consider the following three categories of subjective perception, all gathered by the exit survey, using a five-point  Likert scale (Strongly Disagree to Strongly Agree) for all scale items.

\begin{itemize}[topsep=0pt,leftmargin=*]
    \item Subjective workload. We adopt three applicable items from NASA-TLX \citep{hart2006nasa}:
    \begin{itemize}
        \item \textbf{Mental demand.} I felt that the task was mentally demanding.
        \item \textbf{Feelings of success}. I felt successful accomplishing what I was asked to do.
        \item \textbf{Negative emotions.} I was stressed, insecure, discouraged, irritated, and annoyed during the task.
    \end{itemize}
    \item Confidence. There are multiple loci of confidence in this task: in the model, in one's own ability to create conditional delegation rules, and in the human-AI collaborative outcome. So we consider the following three measure (they were not asked in the manual condition since they do not apply):
    \begin{itemize}
        \item \textbf{Confidence in model.} I trust the model to be able to correctly identify most toxic comments.
        \item \textbf{Confidence in created rules} I am confident that my rules significantly improve the model’s accuracy in detecting toxic comments.
        \item \textbf{Confidence in deployment.} I am confident that my moderator team would feel comfortable relying on the AI model combined with the rules I provided.
    \end{itemize}

    \item Understanding. We are interested in whether global and local explanations could improve people's perceived understanding of the model. We consider both the global understanding of the AI model as a whole and the local understanding on the rationales behind predictions. These questions were skipped in the manual condition.
    \begin{itemize}
        \item \textbf{Understanding of model.} I felt that I had a good understanding of how the AI works.
        \item \textbf{Understanding of prediction.} I felt that I had a good understanding of why the AI identifies a comment to be toxic.  
    \end{itemize}
\end{itemize}

\section{Results}
\label{sec:results}

We report results based on the three sets of evaluation measured described above: efficacy, efficiency and engagment, and subjective measures.
We refer to the WikiAttack task as \ind and Reddit task as \ood and the terms will be used interchangeably.

\subsection{Efficacy}

\begin{figure}
    \centering
    \begin{subfigure}[t]{0.35\textwidth}
        \centering
        \includegraphics[width=\textwidth]{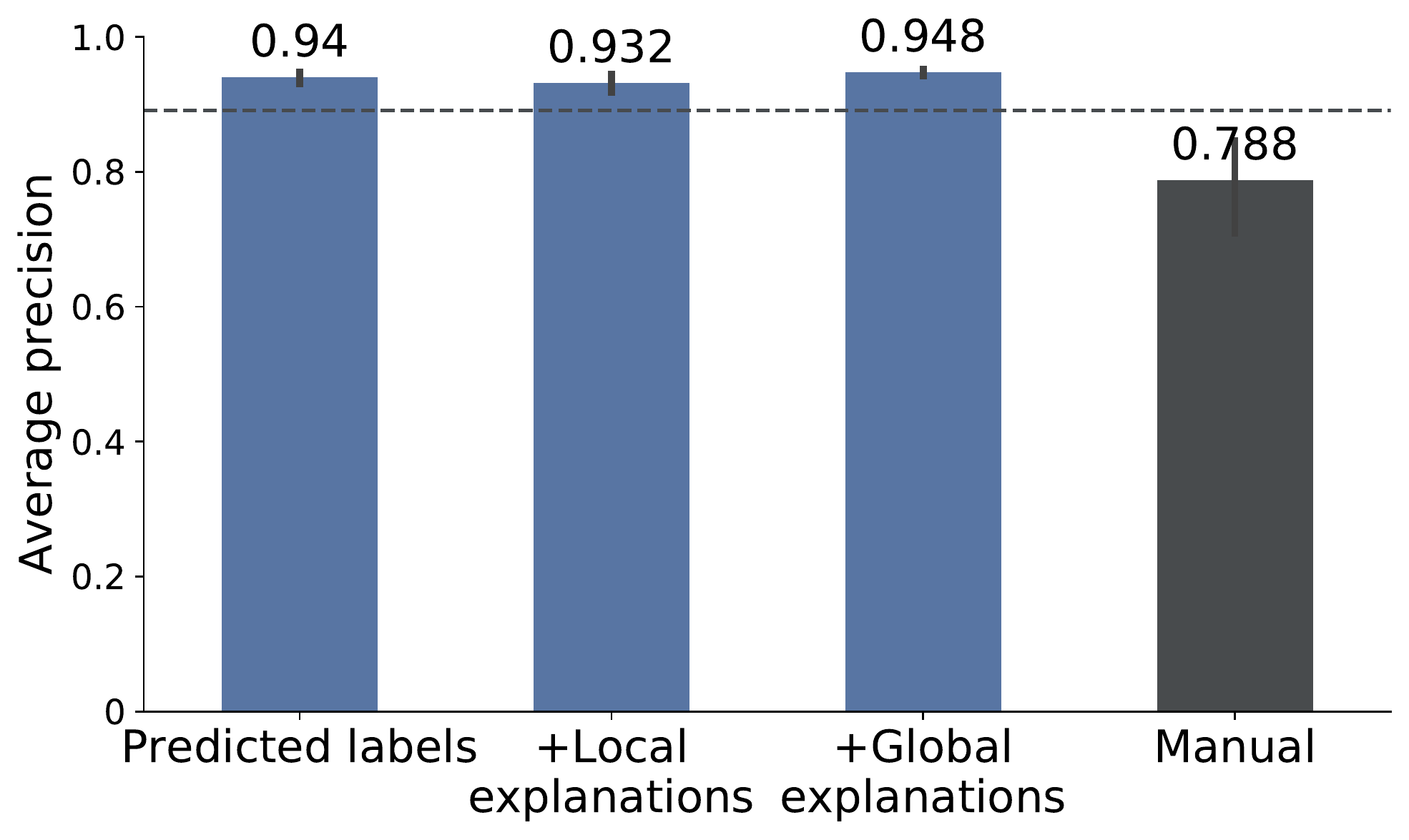}
        \caption{WikiAttack}
        \label{fig:wiki_mean_individual_condition_appropriate_precision}
    \end{subfigure}
    \begin{subfigure}[t]{0.35\textwidth}
        \centering
        \includegraphics[width=\textwidth]{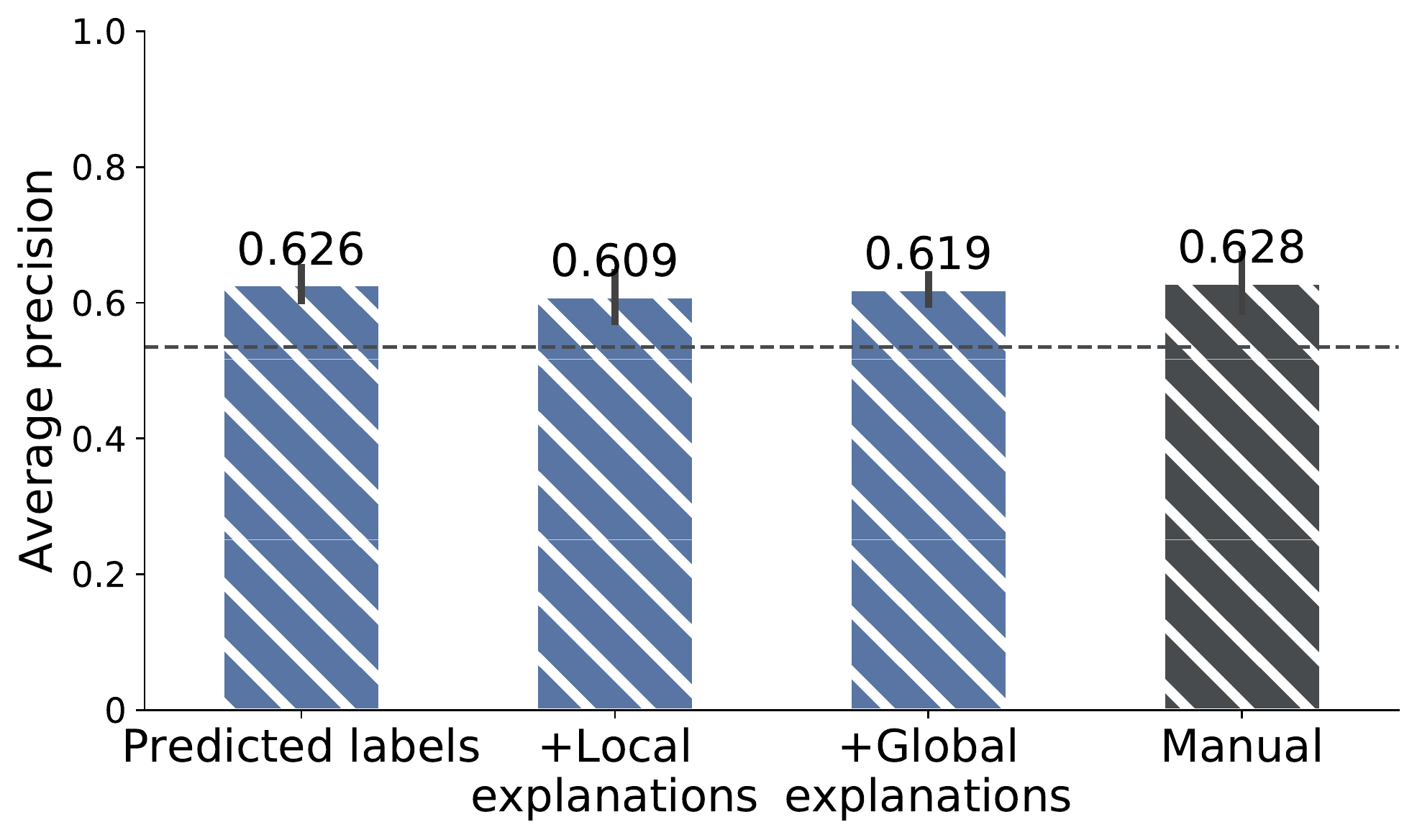}
        \caption{Reddit}
        \label{fig:reddit_mean_individual_condition_appropriate_precision}
    \end{subfigure}
    \caption{Average precision for WikiAttack (\ind) and Reddit (\ood). Error bar shows 95\% confidence interval throughout the paper, and the dashed lines show the precision with the model working alone. The first three conditions represent the precision with conditional delegation, while the manual condition reports precision via the manual rule-based approach by reporting all comments that contain any keyword.
    }
    \Description{This figure shows the average precision for WikiAttack and Reddit. There are four bars in each plot and the first three bars represent conditional delegation conditions while the last bar represents manual condition.}
    \label{fig:mean_precision}
\end{figure}

\begin{figure*}
    \centering
    \begin{subfigure}[t]{0.24\textwidth}
        \centering
        \includegraphics[width=\textwidth]{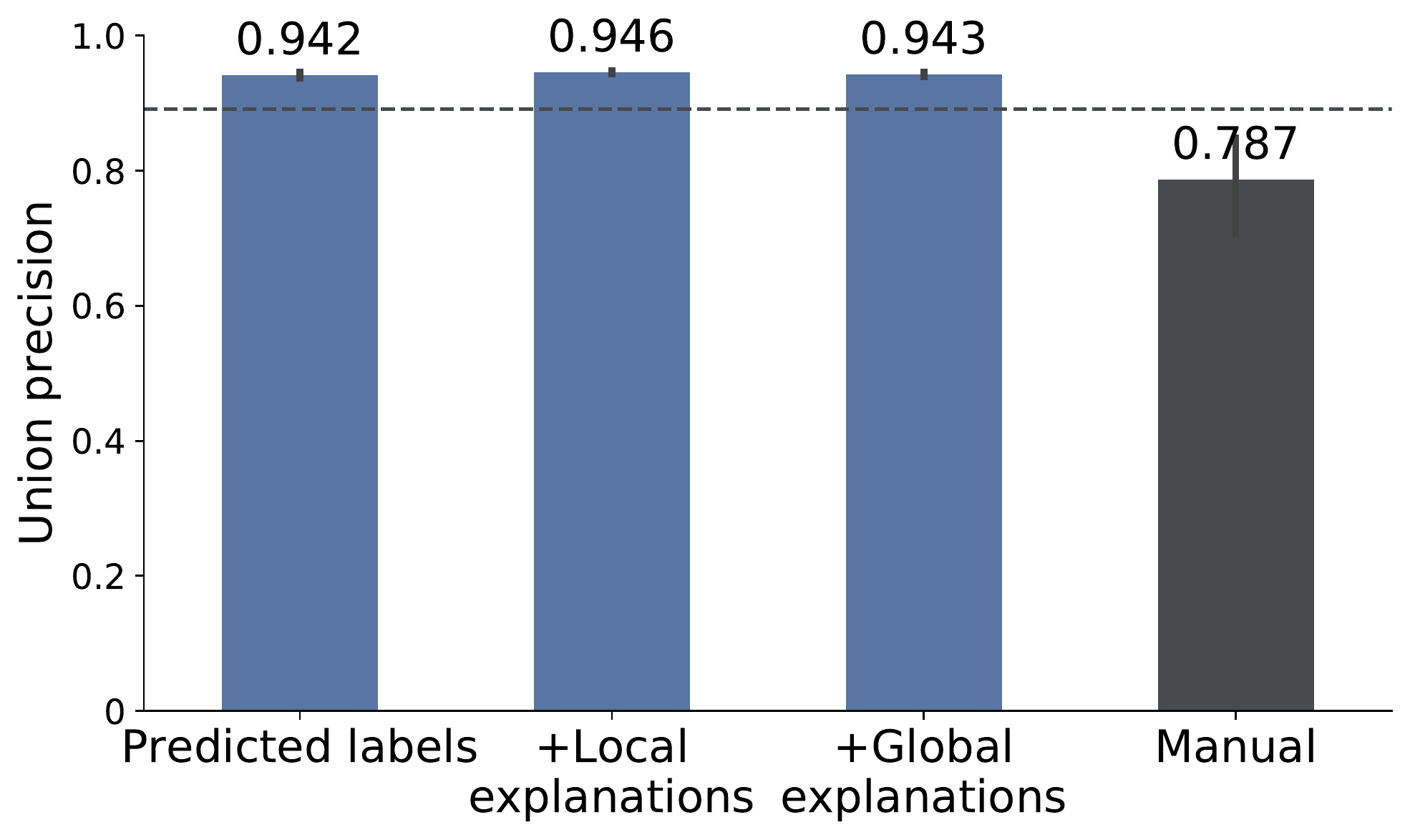}
        \caption{WikiAttack}
        \label{fig:wiki_condition_appropriate_precision}
    \end{subfigure}
    \begin{subfigure}[t]{0.24\textwidth}
        \centering
        \includegraphics[width=\textwidth]{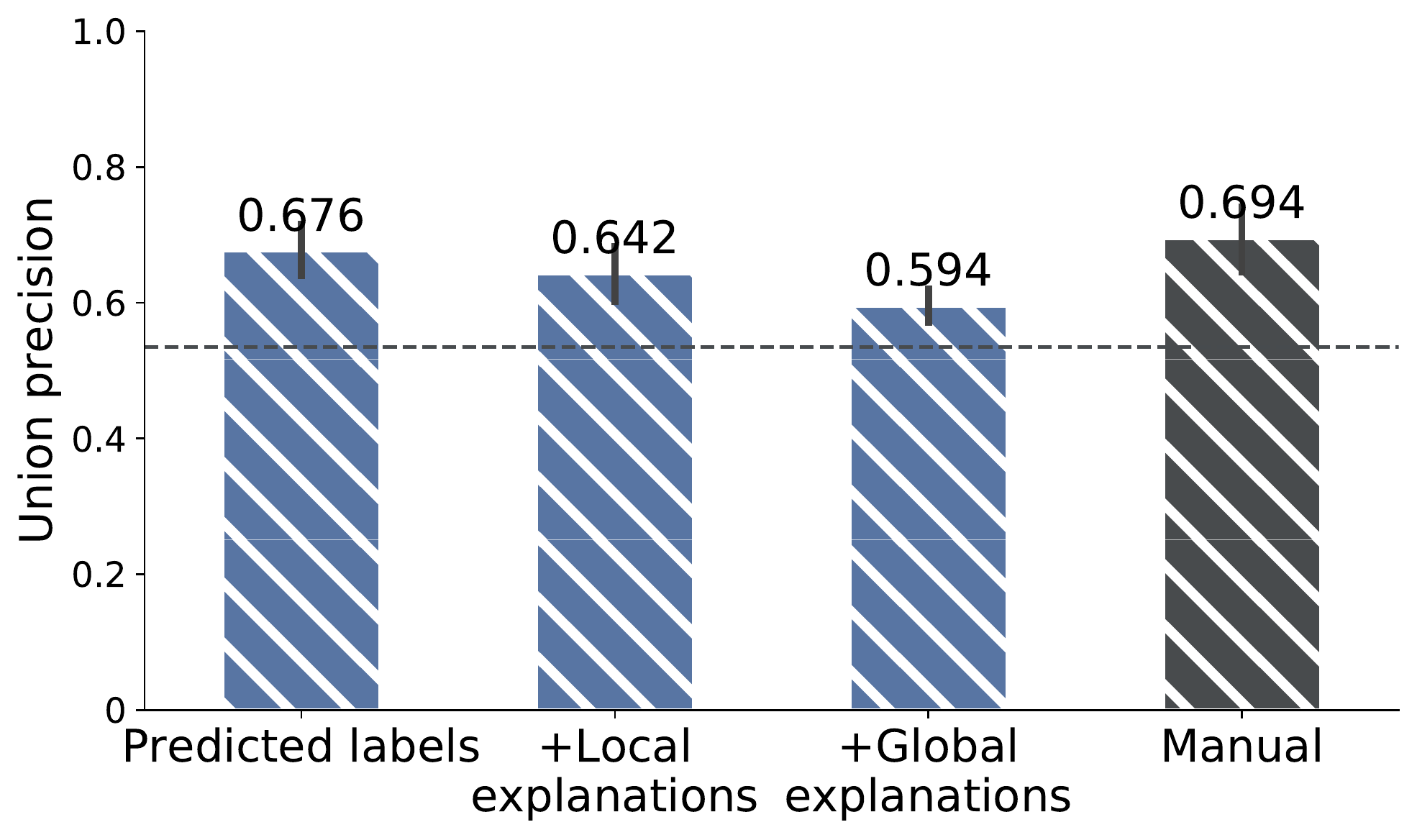}
        \caption{Reddit}
        \label{fig:reddit_condition_appropriate_precision}
    \end{subfigure}
    \begin{subfigure}[t]{0.24\textwidth}
        \centering
        \includegraphics[width=\textwidth]{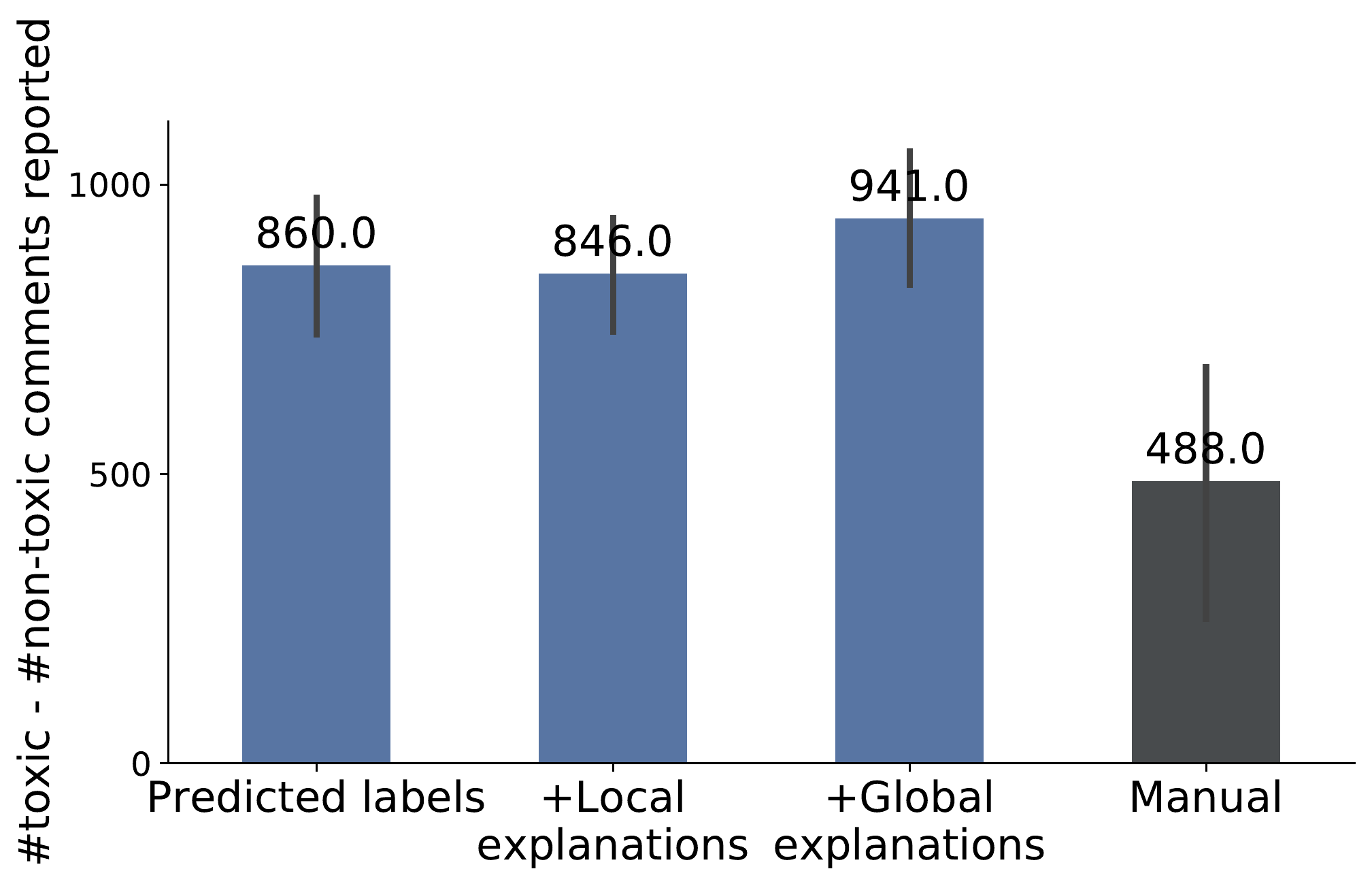}
        \caption{WikiAttack}
        \label{fig:wiki_condition_appropriate_reward}
    \end{subfigure}
    \begin{subfigure}[t]{0.24\textwidth}
        \centering
        \includegraphics[width=\textwidth]{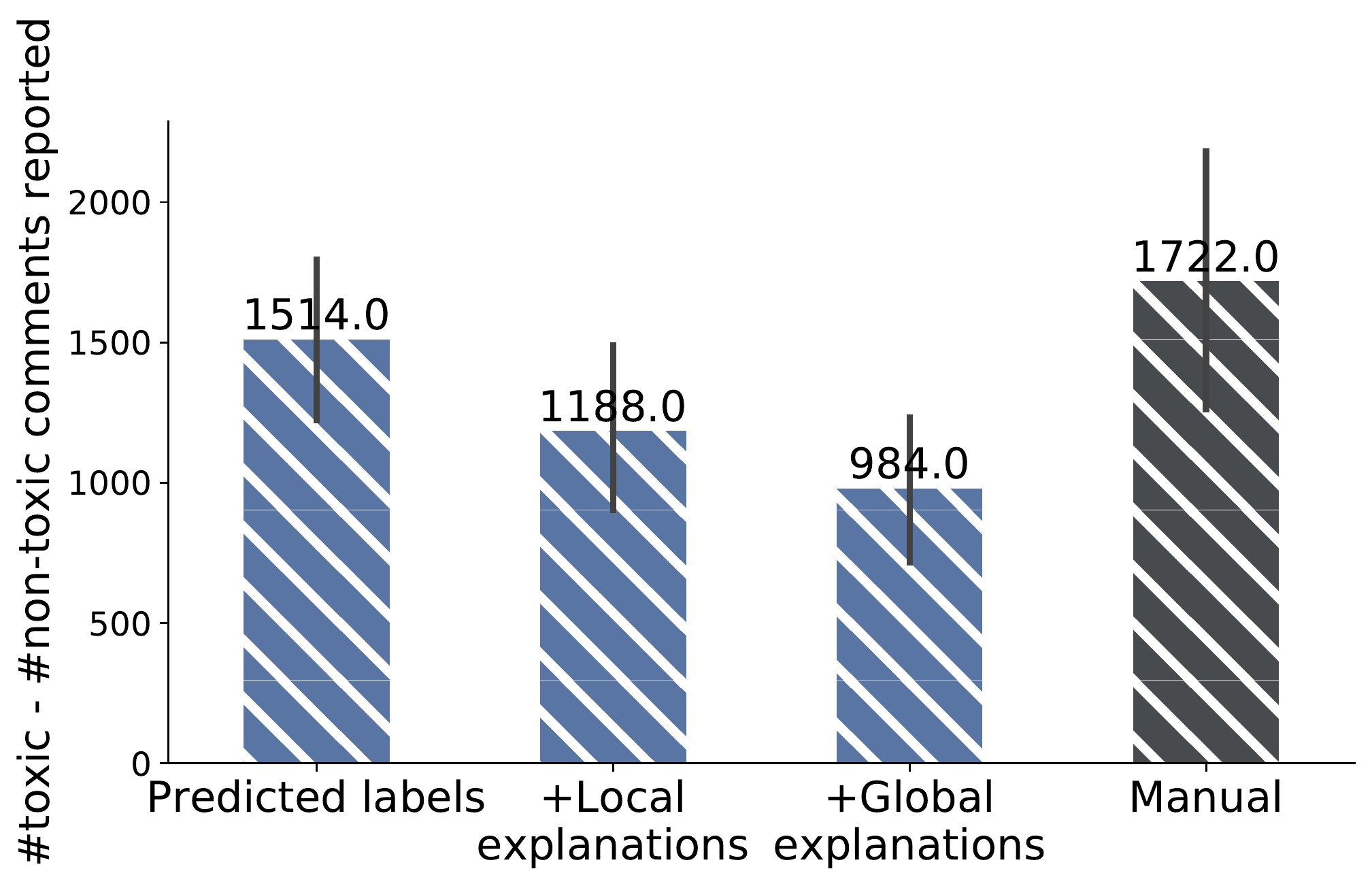}
        \caption{Reddit}
        \label{fig:reddit_condition_appropriate_reward}
    \end{subfigure}
    \caption{Union precision and reward for WikiAttack (\ind) and Reddit (\ood). The dashed lines in \figref{fig:wiki_condition_appropriate_precision} and~\ref{fig:reddit_condition_appropriate_precision} show the precision with the model working alone. 
    Reward is defined as (number of reported toxic comments - number of reported non-toxic comments).
    }
    \Description{These figure shows the union precision and reward for WikiAttack and Reddit.}
    \label{fig:union_precision}
\end{figure*}

\para{Even lay people are able to create rules with higher precision than the model working alone, both \ind and \ood (see \figref{fig:mean_precision} and \figref{fig:union_precision}).}
To determine whether participants are able to create rules that improve model precision,
we conduct $t$-test on the precision of conditional delegation with human-created rules vs. the model working alone.
We find that differences are all statistically significant ($p<$0.001), both on WikiAttack (\ind) and Reddit (\ood), based on average precision and union precision.
In particular, on WikiAttack, the model working alone already outperforms the manual condition, and conditional delegation further improves the precision.
These observations demonstrate that humans, in our case turkers who are not experts in content moderation, are able to create rules that improve model precision, suggesting that conditional delegation can be a promising direction to pursue.

Next, we examine the effect of distribution types and experimental conditions on precision.
We conduct two-way ANOVA of distribution types and experimental condition in average precision and union precision.
We find significant effects in distribution type, experimental condition, and their interaction ($p<$0.001).
The effect of distribution type is the most salient, suggesting a clear difference between in-distribution and out-of-distribution.

Given the significant interaction, we further conduct one-way ANOVA to understand the effect of experimental condition on performance separately for WikiAttack and Reddit, and if significant, conduct post-hoc analysis using Tukey's HSD.
For average precision (\figref{fig:mean_precision}), experimental condition has a significant effect in WikiAttack ($p<$0.001), but not in Reddit ($p=$0.864). 
Post-hoc Tukey's HSD shows that the manual condition is significantly worse than all other experimental conditions with delegation support features ($p<$0.001) on WikiAttack.
For union precision (\figref{fig:wiki_condition_appropriate_precision} and~\ref{fig:reddit_condition_appropriate_precision}) (using rules created by a participant as a set), experimental condition significantly affects performance in both WikiAttack and Reddit ($p<$0.001).
Post-hoc Tukey's HSD shows on WikiAttack, the manual condition is significantly worse than other experimental conditions with delegation support features ($p$ < 0.001). On Reddit, we found the global explanation condition is worse than the manual condition ($p=$0.004), and only showing prediction labels ($p=$0.028).

Finally, we examine the effect of distribution types and experimental conditions on reward. Two-way ANOVA finds a statistically significant effect of distribution type and interaction between distribution type and experimental condition ($p<$0.001). Therefore, we conduct one-way ANOVA to understand the effect of experimental condition on reward separately for WikiAttack and Reddit.
On WikiAttack (\figref{fig:wiki_condition_appropriate_reward}), we find a statistically significant effect of experimental condition ($p=$0.01), and post-hoc Tukey's HSD shows that the manual condition is significantly worse than other experimental conditions with delegation support features ($p<$0.001 for global explanations, $p=$0.007 for local explanations, and $p=$0.004 for predicted labels).
On Reddit, the experimental condition also has a statistically significant effect ($p=$0.018). Post-hoc Tukey's HSD shows that only the difference between the manual condition and global explanations is significant ($p=$0.018).

These results show that,
on WikiAttack, where the model performs well, people can easily identify rules with both high precision (average precision) and with high coverage (reflected by union precision and reward), as long as predicted labels are provided, achieving complementary performance. 
But on Reddit, where the model's base performance is significantly worse, 
it is more challenging to achieve better human-AI performance over the manual rule-based approach.
It follows that in both situations, we do not observe that explanations, either local or global, significantly improve the performance of conditional delegation.
However, adding the global explanation feature seems to unexpectedly hurt people's ability in choosing rules with both high coverage and high precision, and lead to slightly lower human-AI performance in union precision and reward.

\para{What rules do people make?}
To further make sense of their performance, we dive into the content of the rules provided by participants.
\tabref{tb:frequent_rules_by_participants} lists the top rules in each condition along with the percentage of people who chose that rule.
On WikiAttack, participants with delegation support features are more likely to choose ``fuck'' (above 60\%), a high-precision rule to identify toxic content in Wikipedia comments as shown in \secref{sec:method}, than the manual condition (only 36.7\%).
In comparison, for Reddit, ``fuck'' is a less precise rule (i.e., people also use the word in non-toxic comments). The word does not show up in top 10 for the manual condition, but shows up in the other conditions.

This observation suggests that the delegation support features can help users identify good rules when the model performs well, but may mislead people when the model performs poorly. The reason that global explanations slightly hurt the performance in union precision and reward could be that participants were led to choose some high-coverage rules with relatively low precision such as ``fuck'', which was listed in the global keywords (Figure~\ref{fig:interface_predicted_label_local_global}).

\begin{figure}[t]
    \centering
    \begin{subfigure}[t]{0.23\textwidth}
        \centering
        \includegraphics[width=\textwidth]{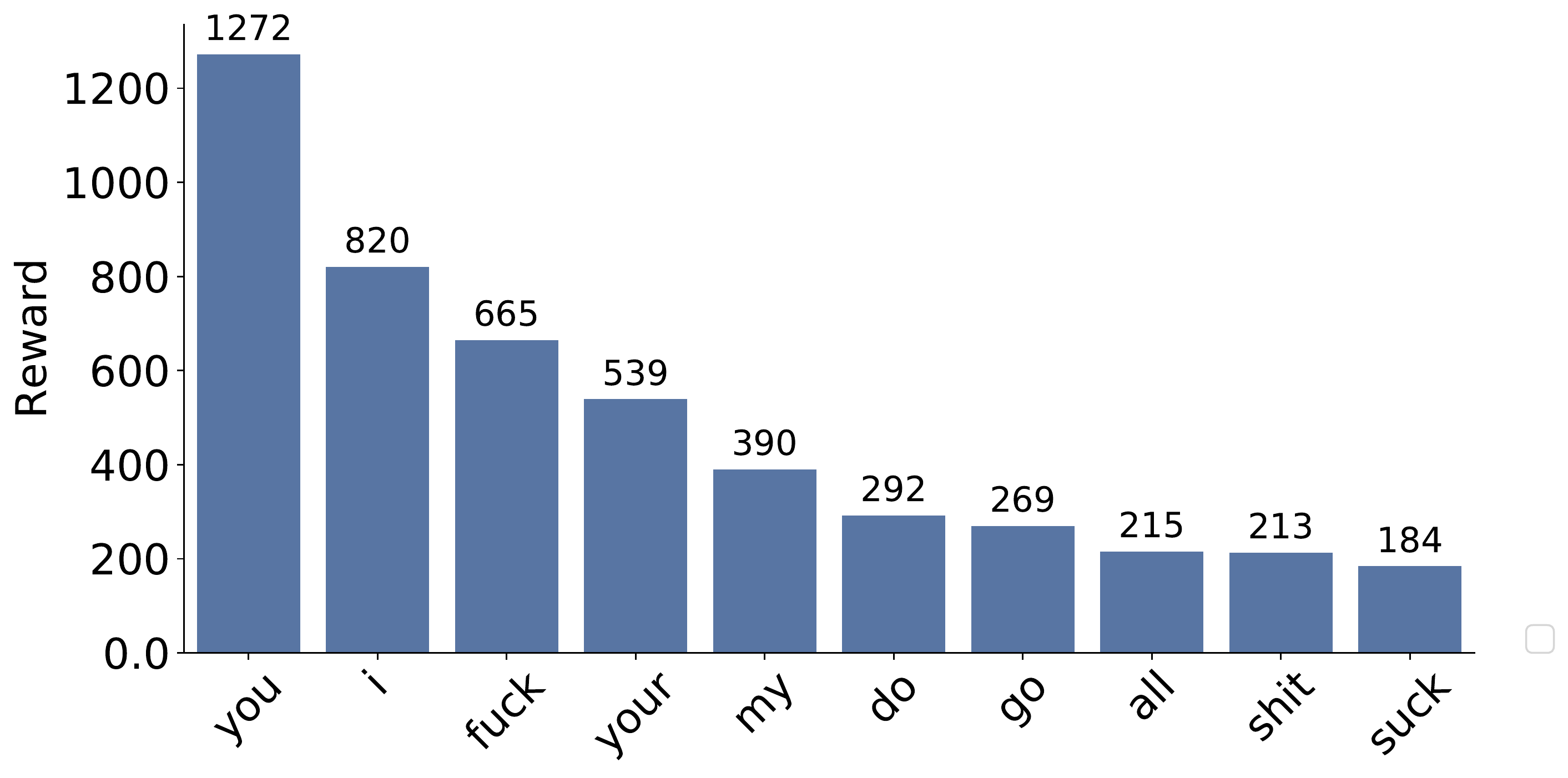}
        \caption{WikiAttack}
        \label{fig:high_model_reward_tokens_wiki}
    \end{subfigure}
    \begin{subfigure}[t]{0.23\textwidth}
        \centering
        \includegraphics[width=\textwidth]{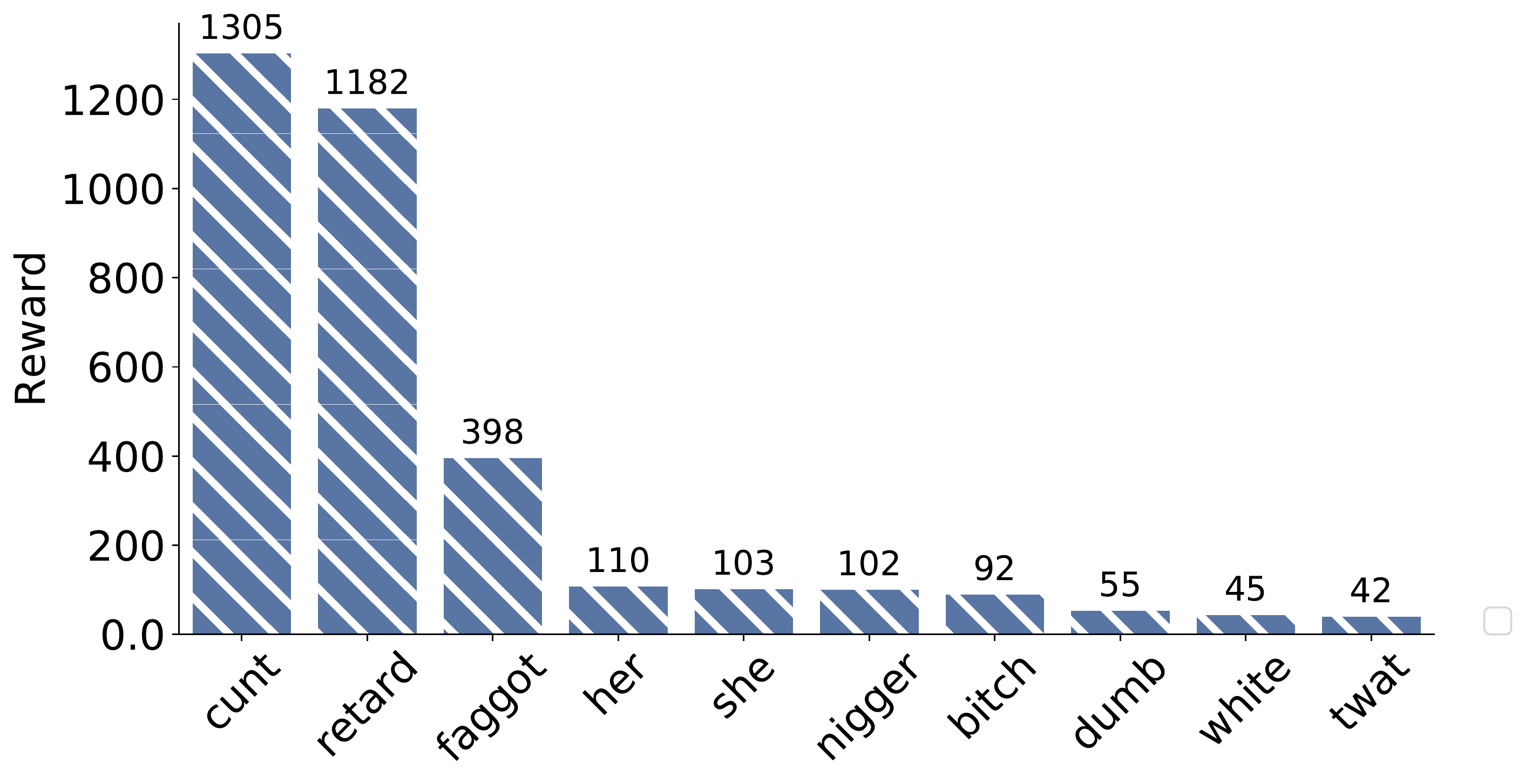}
        \caption{Reddit}
        \label{fig:high_model_reward_tokens_reddit}
    \end{subfigure}
    \caption{Top 10 human-created rules in reward when used for conditional delegation.}
    \Description{This plot shows the top 10 human-created rules in reward.}
    \label{fig:high_model_reward_subjects_tokens}
\end{figure}

\figref{fig:high_model_reward_subjects_tokens} further shows the top words in reward among the rules created by participants.
In addition to ``fuck'' on WikiAttack and ``cunt''/``retard'' on Reddit, the result highlights the advantage of conditional delegation.
Users can achieve high reward by trusting the model beyond swearing words, for instance, ``you'' on WikiAttack and ``her'' on Reddit.
The reason is that the AI model excels at deciding whether ``you'' is used for personal attack or simply for referring purposes.

\begin{table*}[t]
    \small
    \centering
    \begin{tabular}{p{0.2\textwidth}p{0.75\textwidth}}
    \toprule
    \multicolumn{2}{c}{WikiAttack} \\
    Predicted labels & bitch (69.0\%), asshole (62.1\%), \underline{fuck (62.1\%)}, cunt (62.1\%), nigger (51.7\%), dick (48.3\%), faggot (44.8\%), shit (44.8\%), fag (37.9\%), motherfucker (31.0\%) \\
    + Local explanations & bitch (71.4\%), cunt (71.4\%), asshole (67.9\%), \underline{fuck (60.7\%)}, faggot (53.6\%), pussy (42.9\%), dick (39.3\%), fag (35.7\%), cock (35.7\%), retard (35.7\%) \\
    + Global explanations & faggot (86.7\%), nigger (73.3\%), \underline{fuck (70.0\%)}, bitch (66.7\%), cunt (56.7\%), cock (56.7\%), ass (50.0\%), asshole (46.7\%), shit (46.7\%), pussy (36.7\%) \\
    Manual & cunt (70.0\%), nigger (63.3\%), faggot (60.0\%), fag (56.7\%), bitch (53.3\%), asshole (46.7\%), retard (43.3\%), whore (43.3\%), \underline{fuck (36.7\%)}, pussy (26.7\%) \\
    \midrule
    \multicolumn{2}{c}{Reddit} \\
    Predicted labels & cunt (86.2\%), bitch (72.4\%), faggot (62.1\%), \underline{fuck (58.6\%)}, retard (44.8\%), asshole (41.4\%), nigger (41.4\%), pussy (34.5\%), fag (31.0\%), whore (31.0\%) \\
    + Local explanations & cunt (72.4\%), bitch (65.5\%), \underline{fuck (55.2\%)}, pussy (55.2\%), nigger (48.3\%), asshole (41.4\%), faggot (41.4\%), retard (37.9\%), shit (37.9\%), dumbass (34.5\%) \\
    + Global explanations & bitch (69.0\%), cunt (65.5\%), faggot (62.1\%), \underline{fuck (58.6\%)}, retard (58.6\%), nigger (41.4\%), pussy (41.4\%), shit (37.9\%), dick (37.9\%), idiot (34.5\%) \\
    Manual & nigger (76.7\%), cunt (73.3\%), faggot (60.0\%), bitch (56.7\%), retard (43.3\%), whore (43.3\%), asshole (36.7\%), fag (30.0\%), spic (30.0\%), chink (30.0\%) \\
    \bottomrule
    \end{tabular}
    \caption{Most frequent rules chosen by participants.}
    \label{tb:frequent_rules_by_participants}
\end{table*}

\para{Summary.} In short, our results demonstrate that conditional delegation is more effective than the model working alone, and that even laypeople are able to create high-quality conditional delegation rules for content moderation. Compared to the manual rule-based approach currently used for content moderation, advantage of our human-AI collaborative approach via conditional delegation may depend on the base performance of the AI, and may not be sufficient if the AI significantly under-performs, e.g., when used on out-of-distribution comments. Further research is required to understand the necessary conditions for conditional delegation to outperform the manual rule-based approach. Our analysis did not find evidence that model explanations could help people create better rules for conditional delegation. We explore their benefits for other aspects of user experience later.

\subsection{Efficiency and Engagement}

We conduct two-way ANOVA to determine whether distribution type and experimental condition have a significant effect on user engagement (number of actions, number of rules) and efficiency (elapsed time, rules per minute).
In all evaluation measures of engagement and efficiency, we only find statistically significant effects of experiment conditions, suggesting that patterns with two distribution types are comparable. 
Therefore, in this section, we merge the data on WikiAttack and Reddit, and report results from one-way ANOVA on experimental conditions.

\begin{figure}
    \centering
    \begin{subfigure}[t]{0.23\textwidth}
        \centering
        \includegraphics[width=\textwidth]{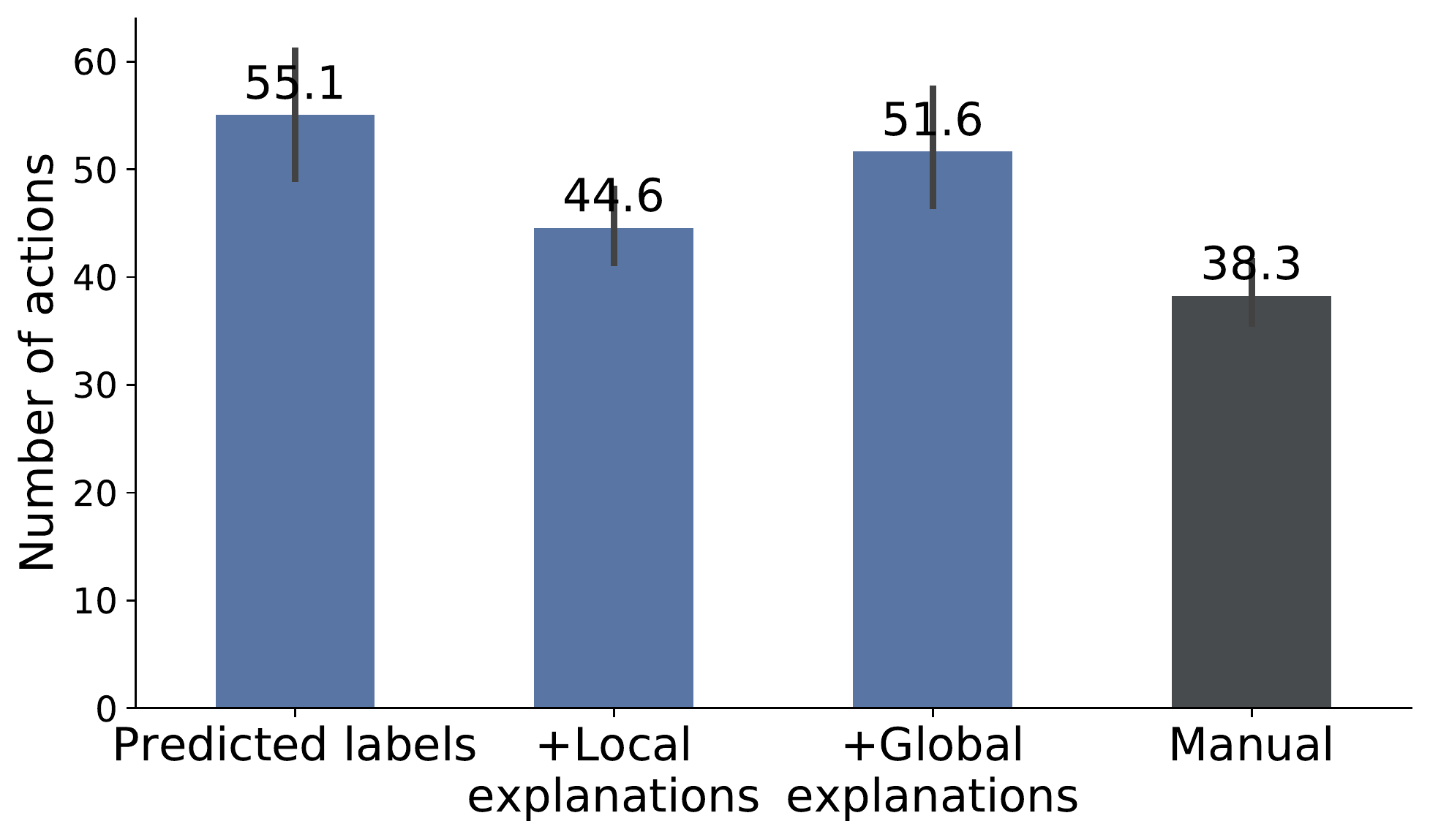}
        \caption{Number of actions.}
        \label{fig:number_of_actions}
    \end{subfigure}
    \begin{subfigure}[t]{0.23\textwidth}
        \centering
        \includegraphics[width=\textwidth]{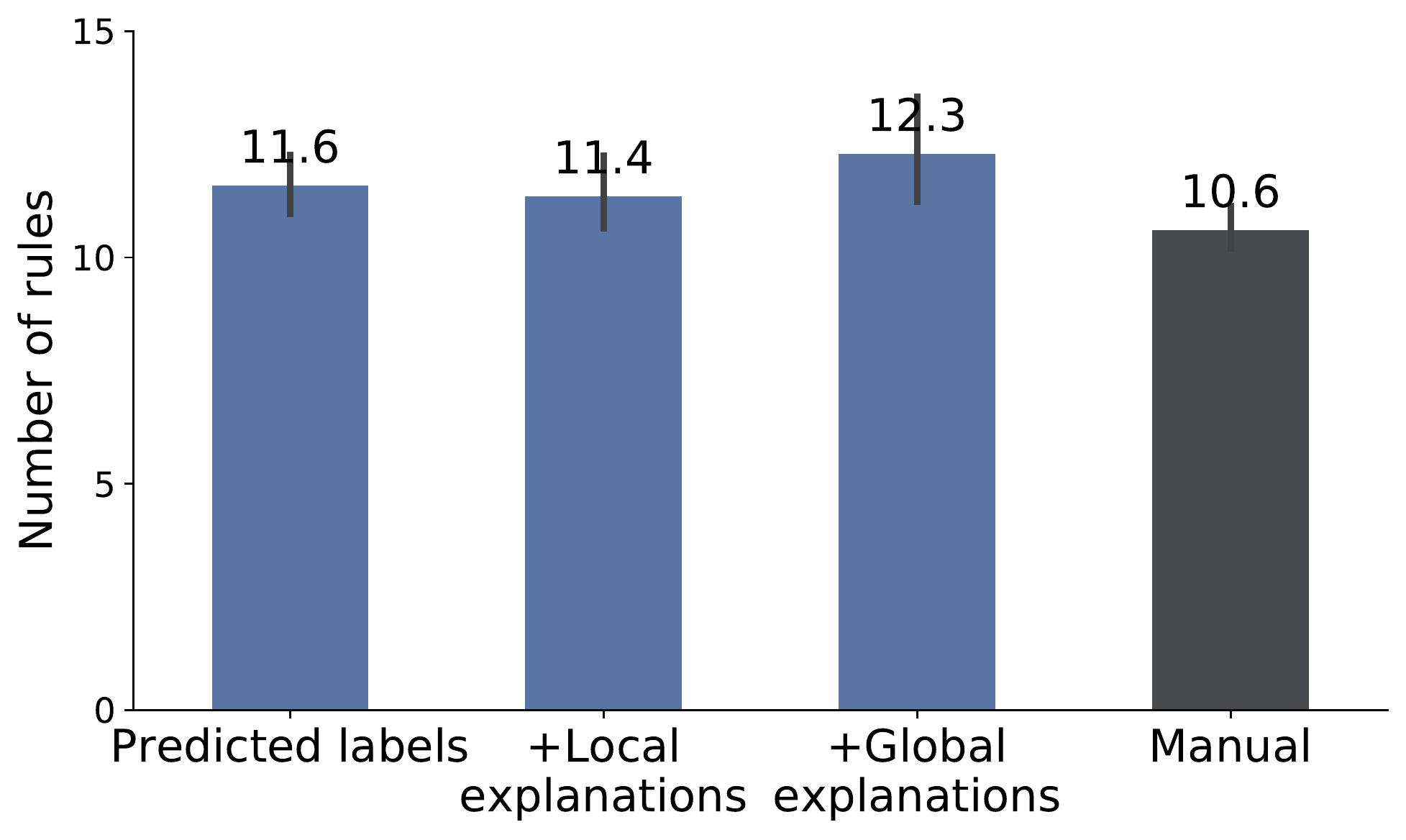}
        \caption{Number of rules.}
        \label{fig:number_of_rules}
    \end{subfigure}
    \caption{Engagement. Conditional delegation with all delegation support features leads to much better engagement and more submitted rules than the manual condition.}
    \Description{The plots show the number of actions taken and number of rules created by different experiment conditions.}
    \label{fig:engagement}
\end{figure}

\para{Participants working on conditional delegation are more engaged (see \figref{fig:engagement}).}
\figref{fig:number_of_actions} shows that participants with all delegation support features were much more engaged than the manual condition. In particular, predicted labels only condition incurred many more actions than other conditions.
One-way ANOVA also finds a statistically significant effect in experimental condition ($p<$ 0.001).
Post-hoc Tukey's HSD shows the negative difference between the manual condition and other experimental conditions are all statistically significant ($p<$0.001 for predicted labels and global explanations, $p=$0.009 for local explanations).

When it comes to number of rules, the outcome of task engagement, 
the difference is not as salient.
Because we require a minimum of 10 rules, every condition leads to a little above 10 rules: the manual condition is just above 10 at 10.6, while global explanations leads to 12.3 rules.
That said, one-way ANOVA still finds a significant effect in experimental condition ($p=$0.021).
Post-hoc Tukey's HSD shows that the difference between global explanations and the manual condition is statistically significant ($p=$0.028).

\begin{figure*}
    \centering
    \begin{subfigure}[t]{0.3\textwidth}
        \centering
        \includegraphics[width=\textwidth]{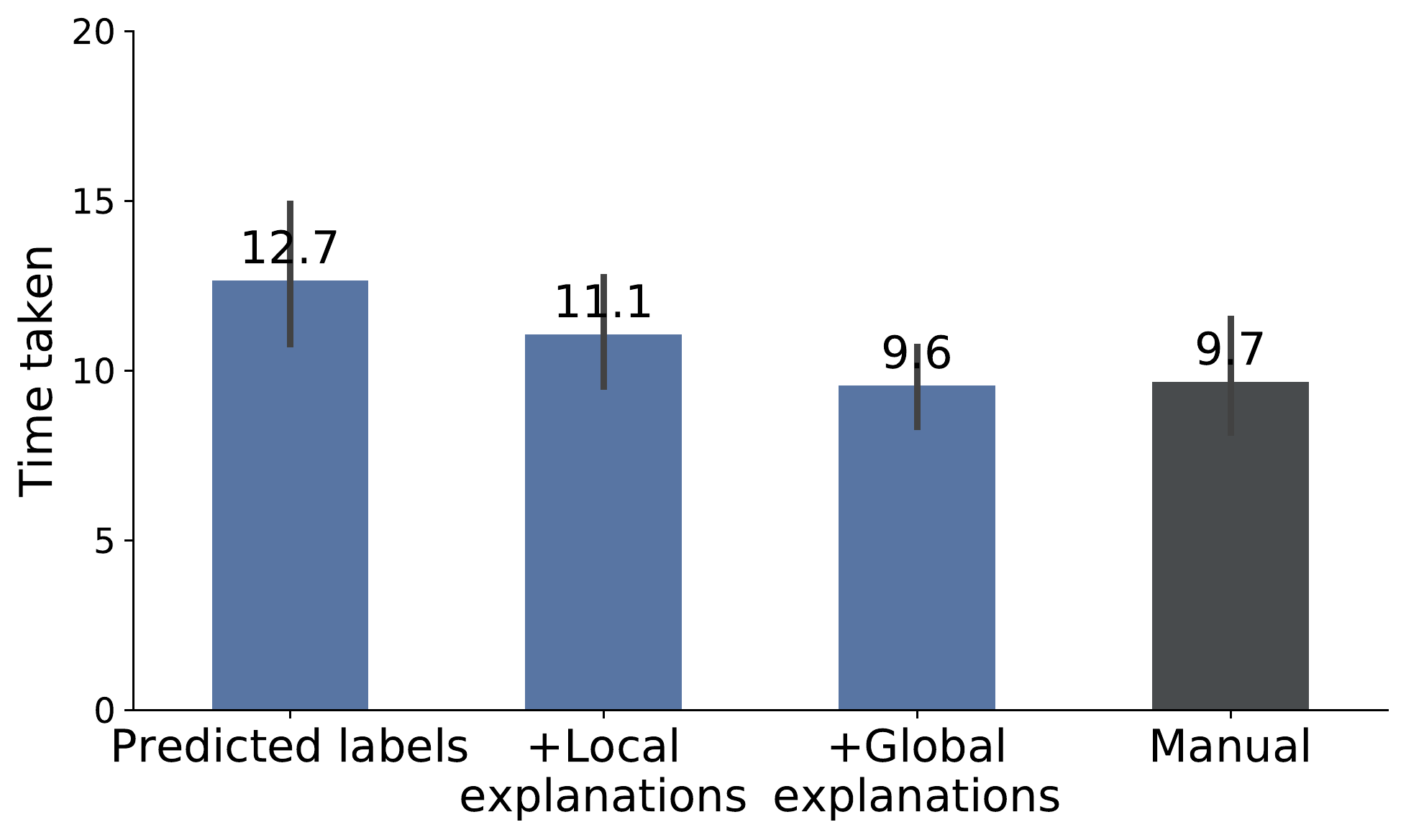}
        \caption{Elapsed time}
        \label{fig:elapsed_time}
    \end{subfigure}
    \begin{subfigure}[t]{0.3\textwidth}
        \centering
        \includegraphics[width=\textwidth]{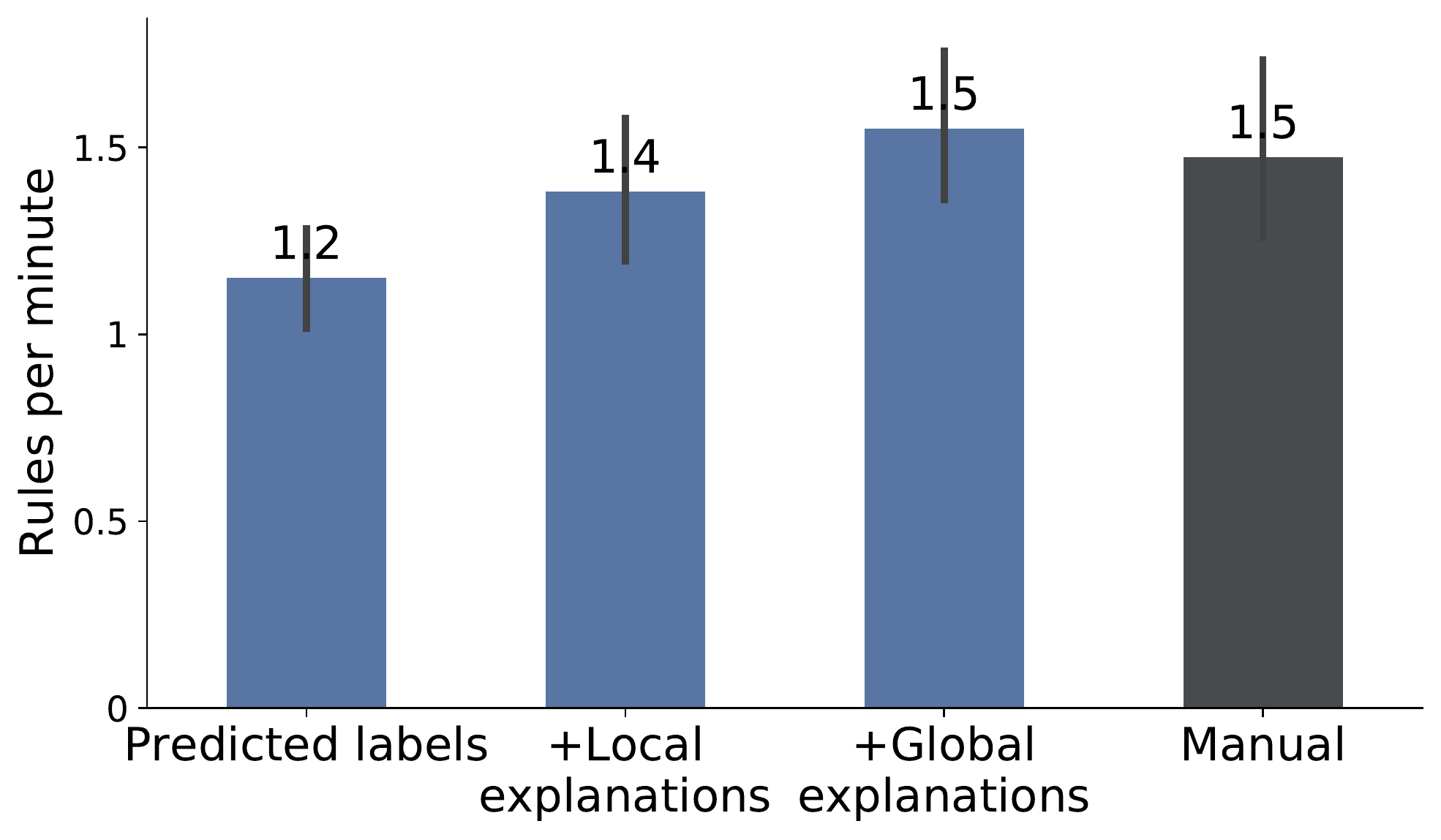}
        \caption{Rules per minute}
        \label{fig:rules_per_minute}
    \end{subfigure}
    \begin{subfigure}[t]{0.3\textwidth}
        \centering
        \includegraphics[width=\textwidth]{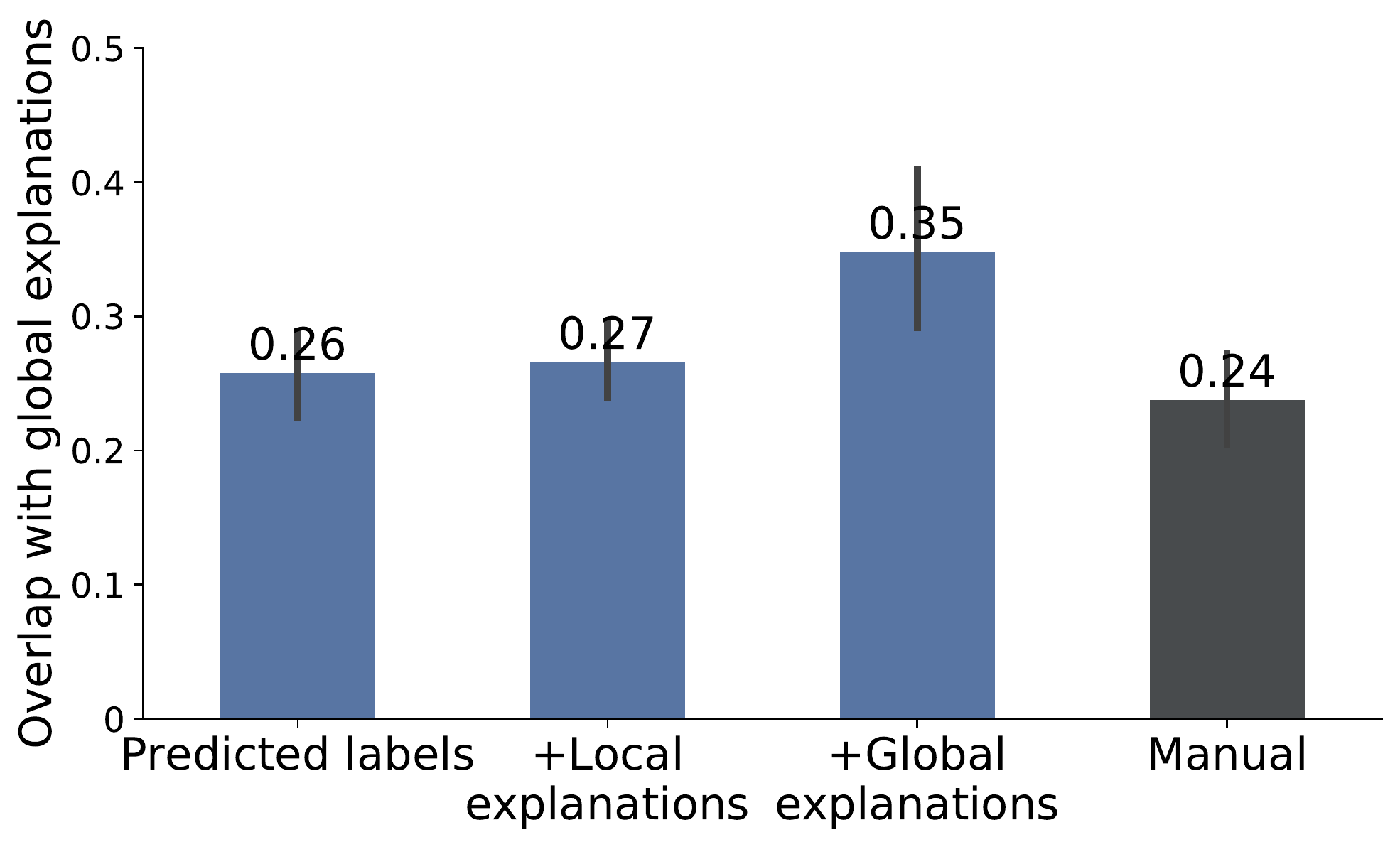}
        \caption{Overlap with global explanations.}
        \label{fig:top_15_token_agreement}
    \end{subfigure}
    \caption{Efficiency. Participants spent more time working on conditional delegation than the manual condition, but the efficiency is improved with explanations, especially global explanations.}
    \Description{The plots show time taken, rules per minute, and overlap with global explanations by different experiment conditions.}
    \label{fig:efficiency}
\end{figure*}

\para{Explanations improve task efficiency (see \figref{fig:elapsed_time} and~\ref{fig:rules_per_minute}).}
\figref{fig:elapsed_time} shows the time spent on the interactive interface in each condition: participants with predicted labels only spent the most time on this task.
This result is consistent with the number of actions because it likely requires more time to take more actions.
However, the difference is relatively weak: one-way ANOVA shows that the effect of experimental conditions is only borderline significant ($p=$0.074), and
post-hoc Tukey's HSD do not find any statistically different pairs.

Rules per minute is a better measure of efficiency since it reflects how long it takes for people to identify a rule that they are satisfied with.
The trend is reversed from the time spent: 
predicted labels only lead to the lowest number of rules per condition, however, with the help of explanations, humans are able to achieve a greater number of rules per minute. 
One-way ANOVA confirms a statistically significant effect of experimental condition ($p=$0.008).
Post-hoc Tukey's HSD suggests that the only statistically different pair is predicted labels only and global explanations ($p=$0.031), suggesting that global explanations helped participants to achieve the highest efficiency to come up with rules.

\para{Global explanations lead to much higher overlap between  most frequent words in rationales (\figref{fig:top_15_token_agreement}).}
To understand this improvement in efficiency, we examine the overlap between human-created rules and the most frequent words in rationales (\tabref{tb:frequent_words}), which are the words shown in global explanations and also more likely to have appeared in the highlighted words in local explanations.
\figref{fig:top_15_token_agreement} shows that global explanations lead to much higher overlap than the other conditions.
This observation confirms that global explanations provide direct hints for possible rules, thus improved the efficiency to come up with required number of rules.

\para{Summary.} Taken together, our results show that people are more engaged when performing conditional delegation than working on creating manual rules, with more actions and a tendency to spend more time on the task. This tendency comes with a cost of efficiency in creating rules when only showing predicted labels. Showing model explanations, especially global explanations, can significantly improve the efficiency, resulting in comparable efficiency between conditional delegation and the manual rule-based approach. The reason can be attributed to participants leveraging keywords in explanations as hints to create delegation rules. Future research is required to explore means to encourage people to examine the performance of these hinted rules more carefully, to improve both efficiency and efficacy. 

\subsection{Subjective Perception}

Finally, we report the subjective perception of participants (subjective workload, confidence, and perceived understanding of AI). All results are based on answers in exit survey, with a five-point Likert scale.

\begin{figure}
    \centering
    \begin{subfigure}[t]{0.23\textwidth}
        \centering
        \includegraphics[width=\textwidth]{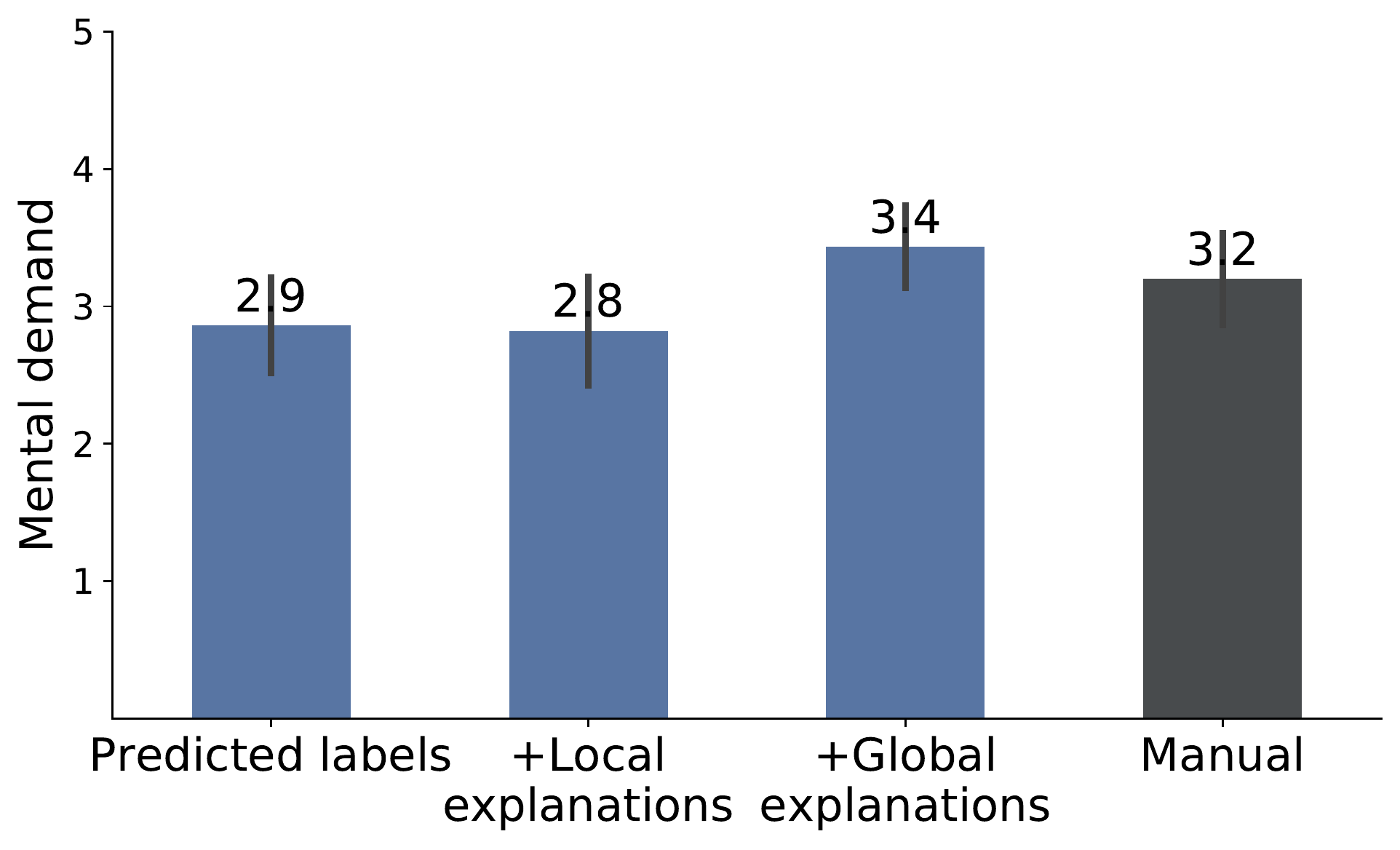}
        \caption{Mental demand (Wiki)}
        \label{fig:mental_demand_wiki}    
    \end{subfigure}
    \begin{subfigure}[t]{0.23\textwidth}
        \centering
        \includegraphics[width=\textwidth]{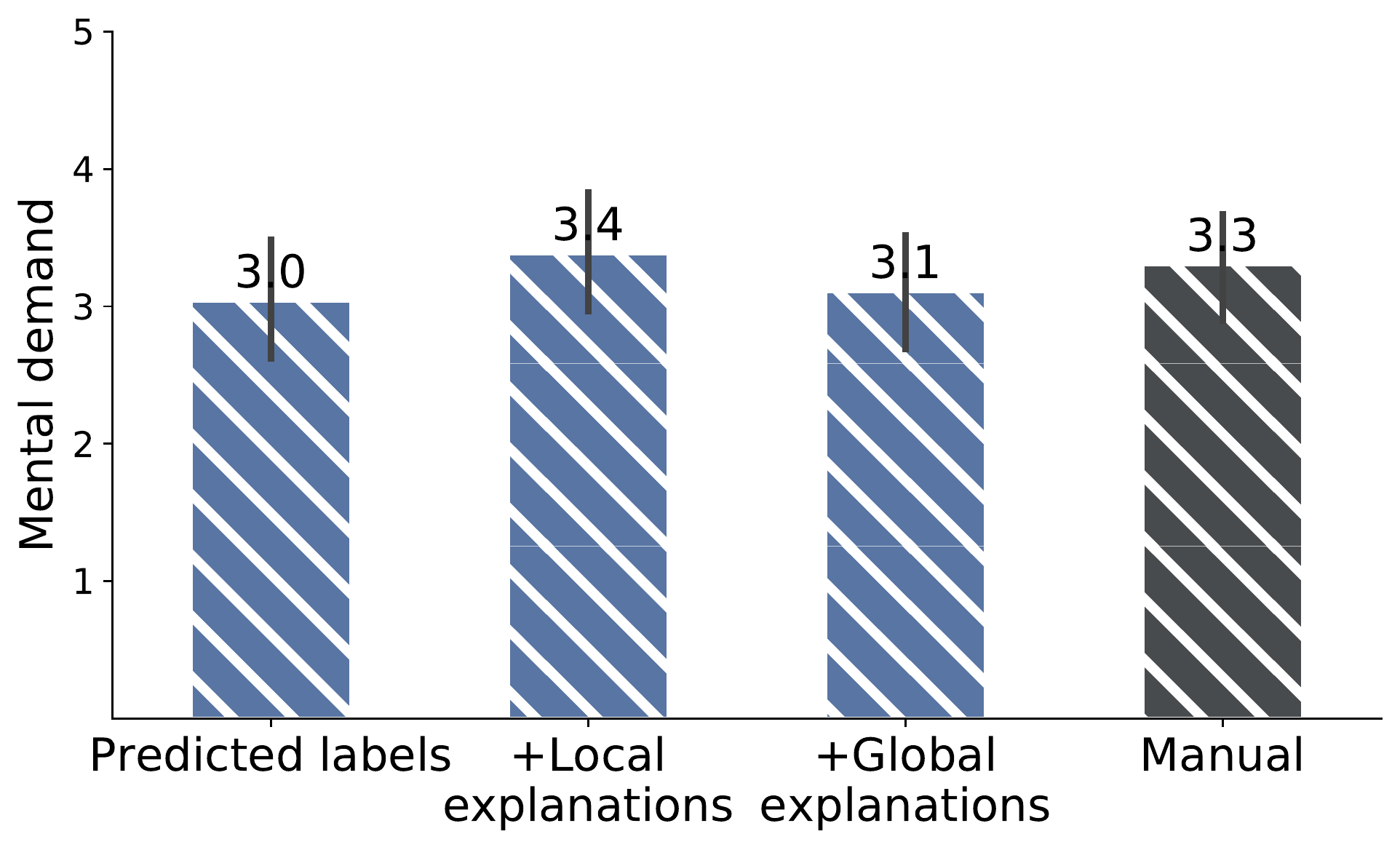}
        \caption{Mental demand (Reddit)}
        \label{fig:mental_demand_reddit}
    \end{subfigure}
    \begin{subfigure}[t]{0.23\textwidth}
        \centering
        \includegraphics[width=\textwidth]{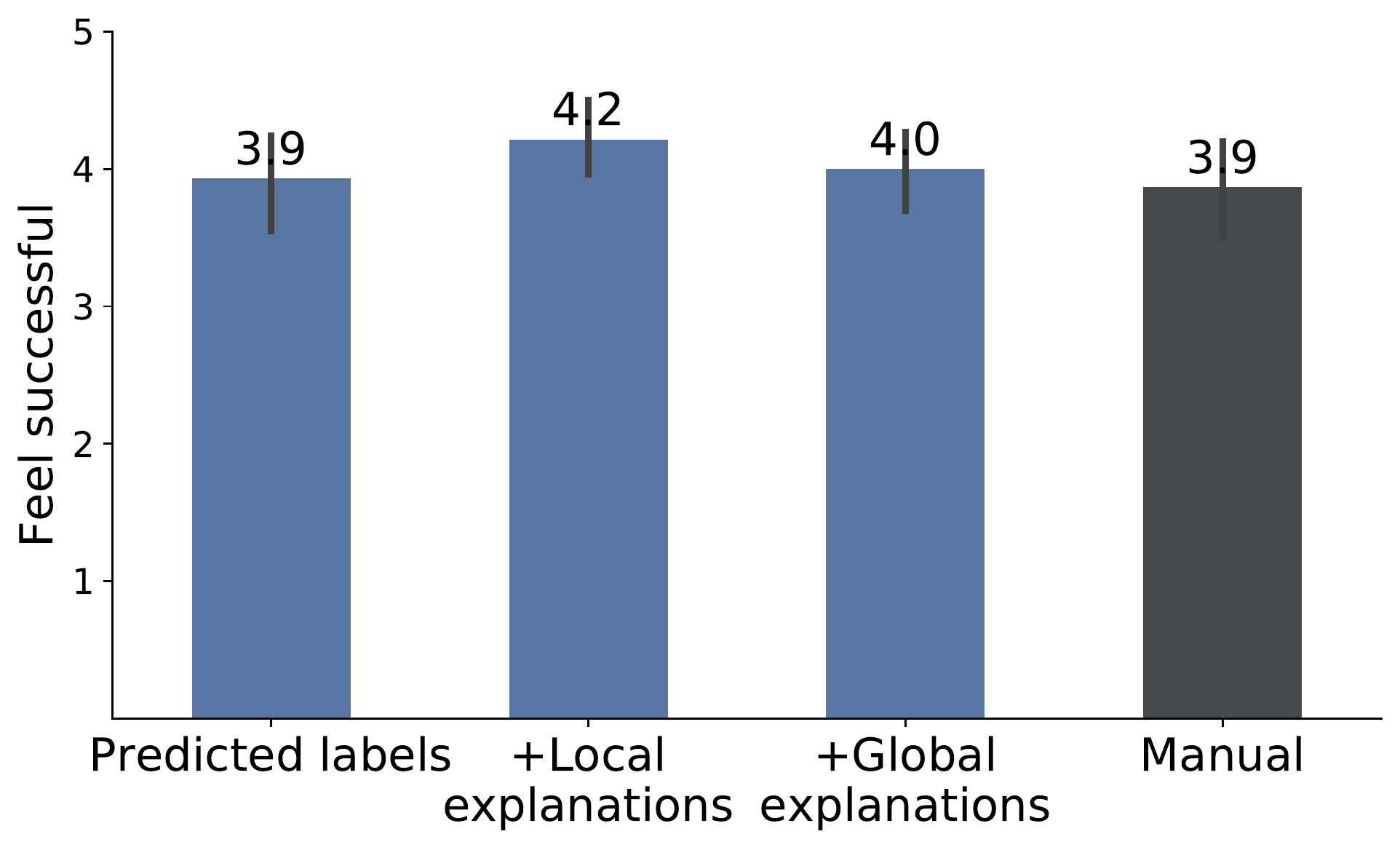}
        \caption{Feeling successful (Wiki)}
        \label{fig:feel_successful_wiki}
    \end{subfigure}
    \begin{subfigure}[t]{0.23\textwidth}
        \centering
        \includegraphics[width=\textwidth]{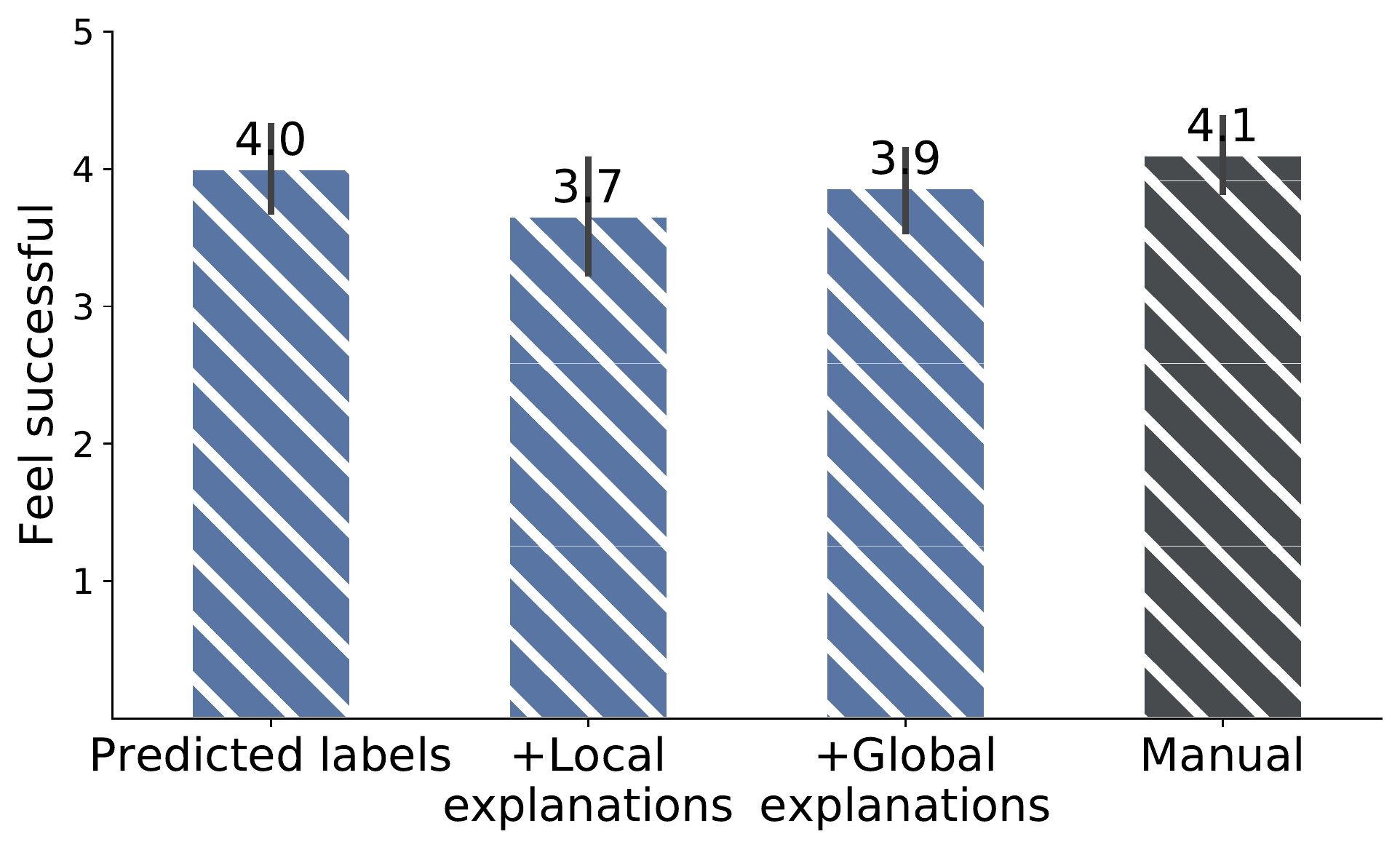}
        \caption{Feeling successful (Reddit)}
        \label{fig:feel_successful_reddit}
    \end{subfigure}
    \\
    \begin{subfigure}[t]{0.23\textwidth}
        \centering
        \includegraphics[width=\textwidth]{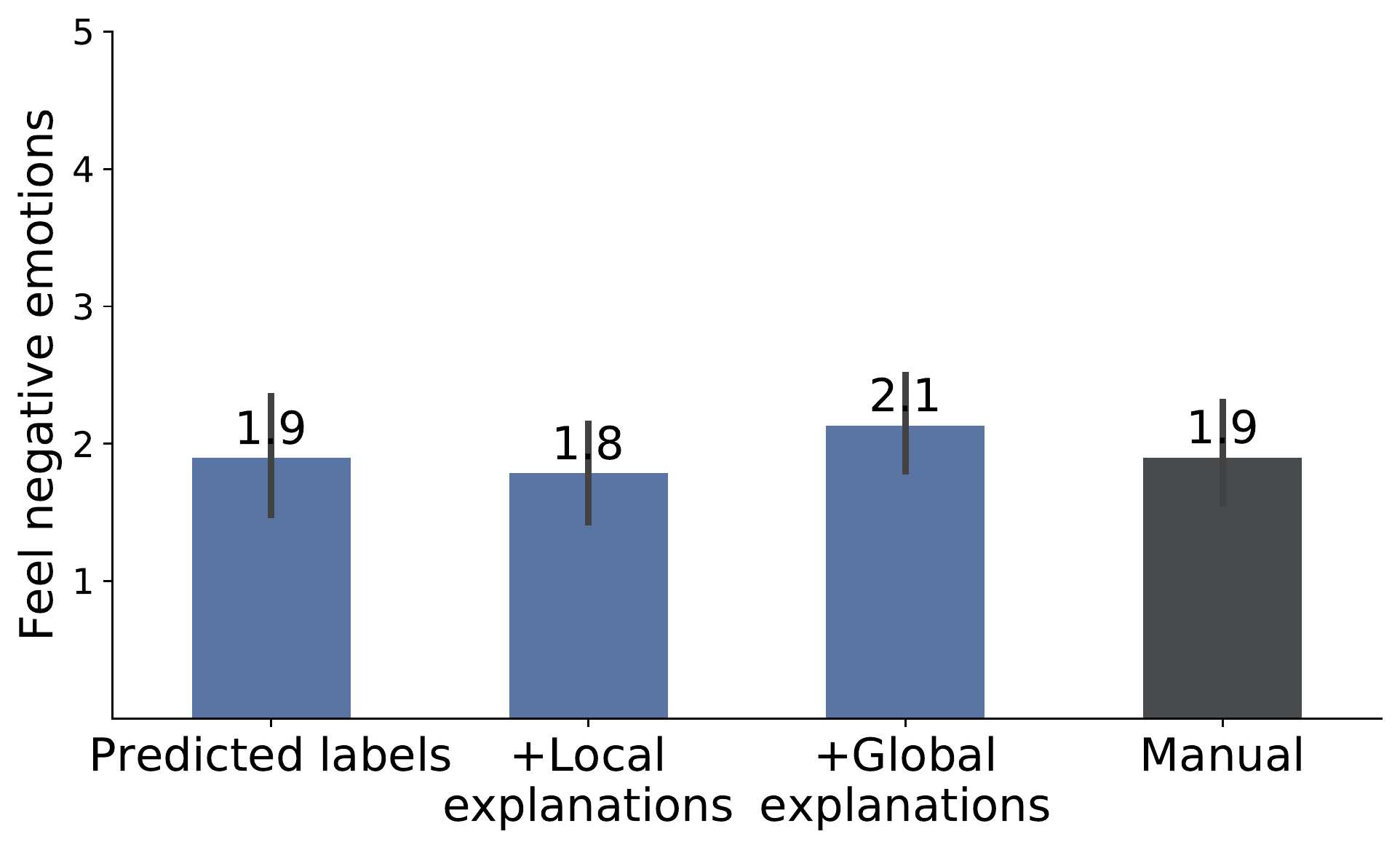}
        \caption{Negative emotions (Wiki)}
        \label{fig:negative_emotions_wiki}
    \end{subfigure}
    \begin{subfigure}[t]{0.23\textwidth}
        \centering
        \includegraphics[width=\textwidth]{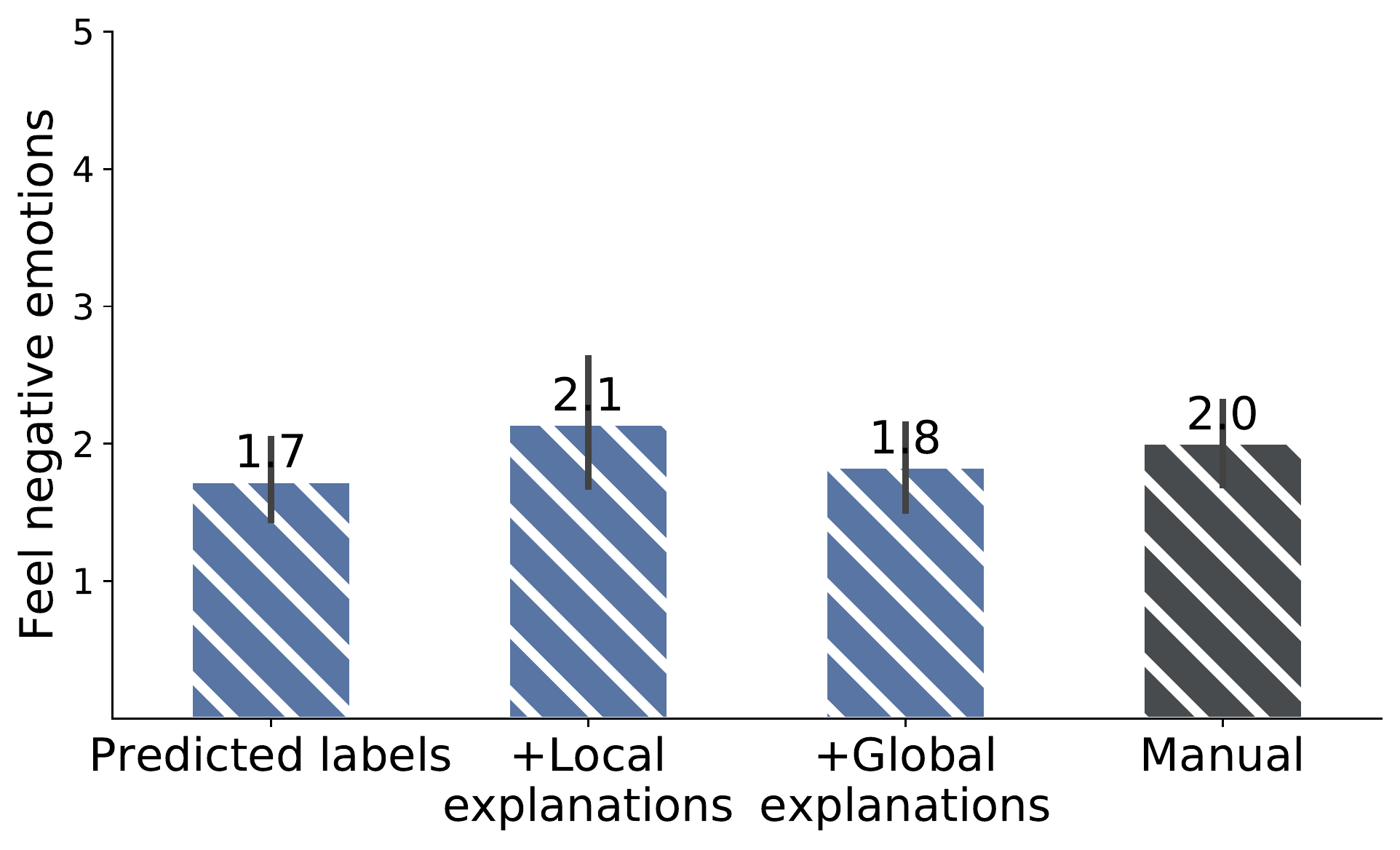}
        \caption{Negative emotions (Reddit)}
        \label{fig:negative_emotions_reddit}
    \end{subfigure}
    \caption{Subjective workload. Overall, participants were neutral about whether the task was mentally demanding (M=3.15, SD=1.08), agreed that they felt successful in accomplishing the task (M=3.95, SD=0.91), and disagreed that they felt negative emotions (M=1.93, SD=1.03).}
    \Description{The plots show the different subjective measures' ratings adopted from NASA-TLX by different experiment conditions.}
    \label{fig:task_satisfaction}
\end{figure}

\para{Subjective workload.} Overall, participants were neutral about whether the task was mentally demanding (M=3.15, SD=1.08), agreed that they felt relatively successful in accomplishing the task (M=3.95, SD=0.91), and disagreed that they felt negative emotions (M=1.93, SD=1.03).
Two-way ANOVA does not show any statistically significant effects of distribution types and experimental conditions. For WikiAttack, we observe a weak trend that local explanations lead to less subjective workload (lower mental demand, more feeling of success, and less negative emotion) while adding global explanation has the opposite effect. These patterns, however, do not hold for Reddit.

\begin{figure}
    \centering
    \begin{subfigure}[t]{0.23\textwidth}
        \centering
        \includegraphics[width=\textwidth]{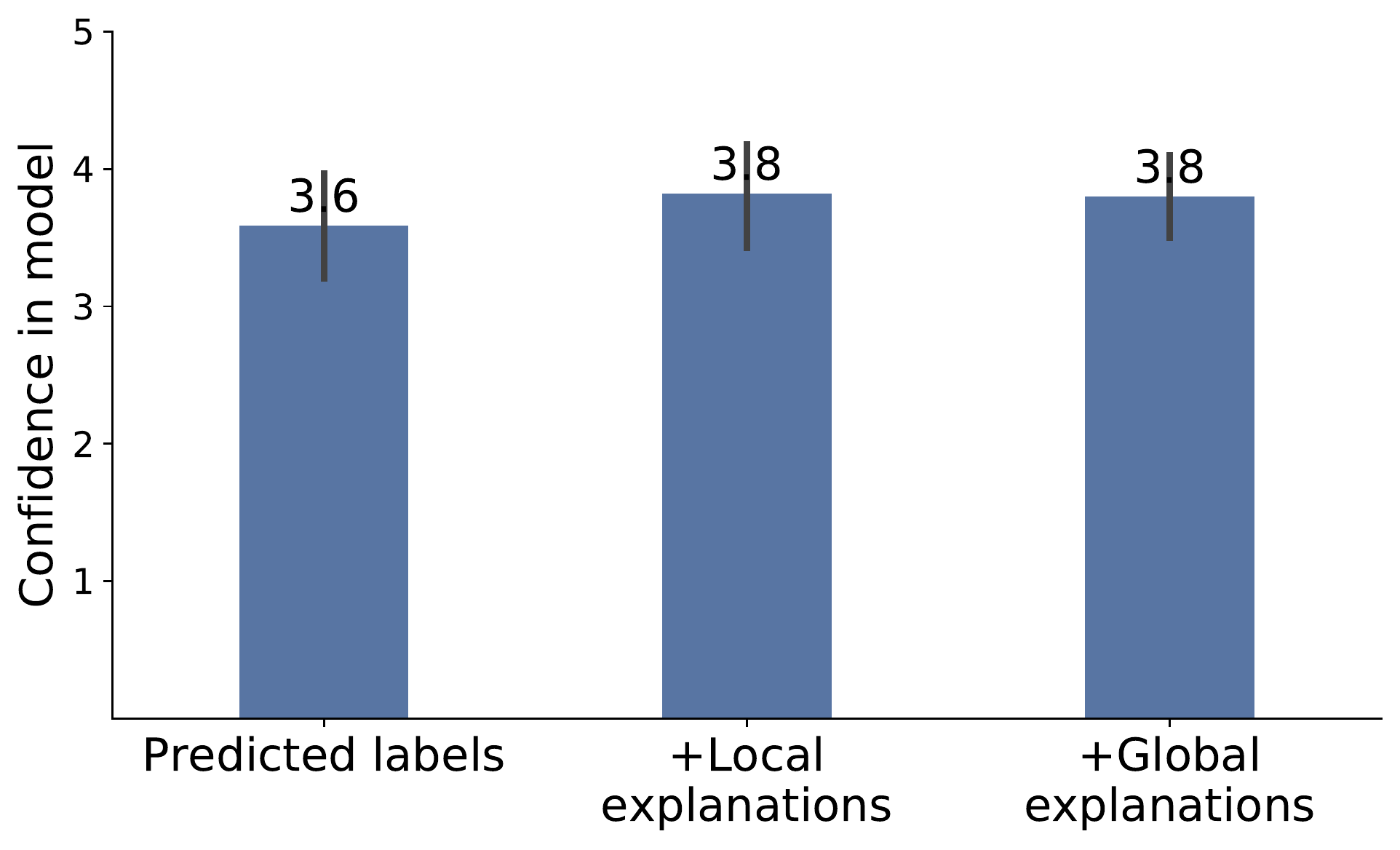}
        \caption{Confidence in model (Wiki)}
        \label{fig:confidence_model_wiki}    
    \end{subfigure}
    \begin{subfigure}[t]{0.23\textwidth}
        \centering
        \includegraphics[width=\textwidth]{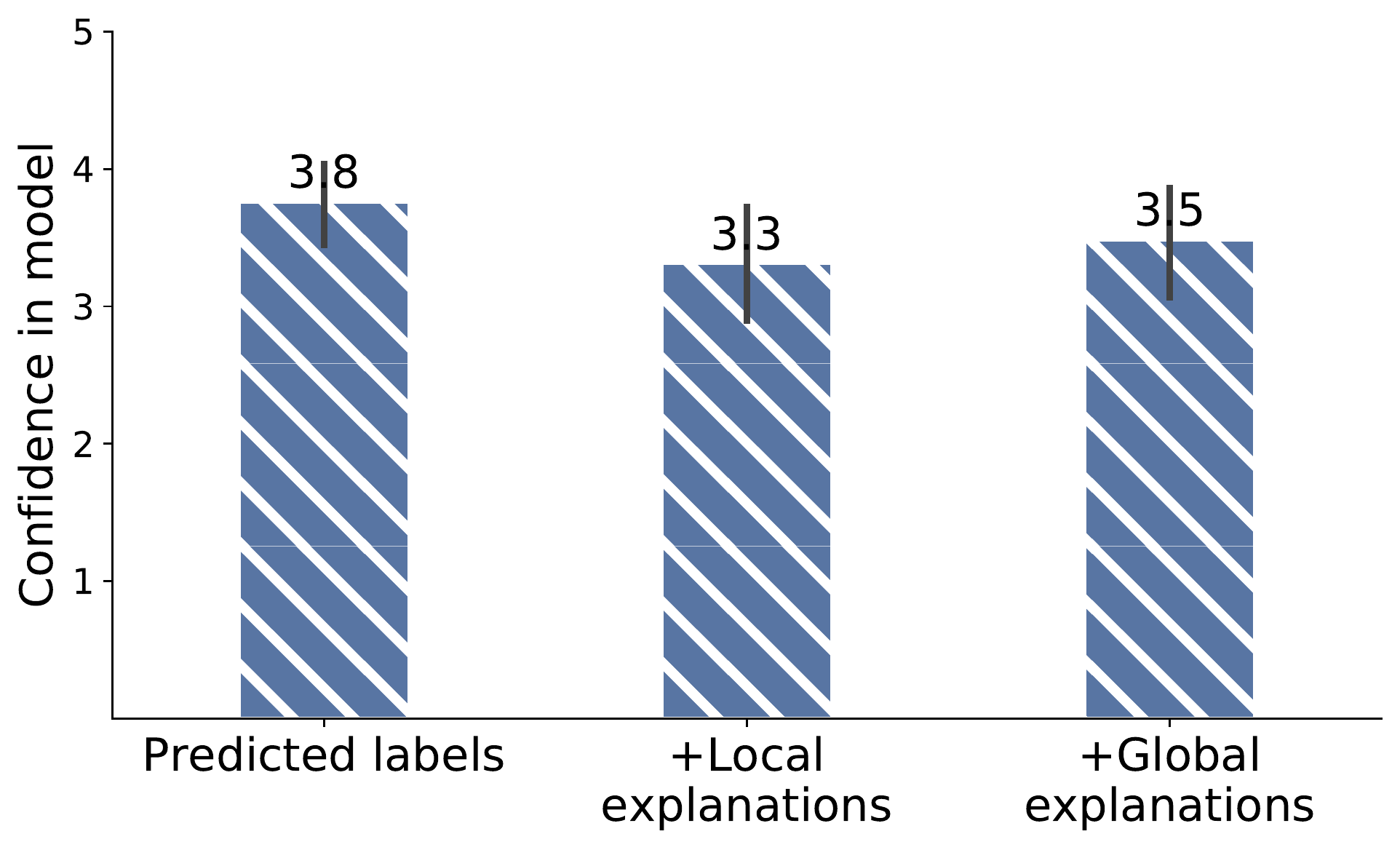}
        \caption{Confidence in model (Reddit)}
        \label{fig:confidence_model_reddit}
    \end{subfigure}
    \begin{subfigure}[t]{0.23\textwidth}
        \centering
        \includegraphics[width=\textwidth]{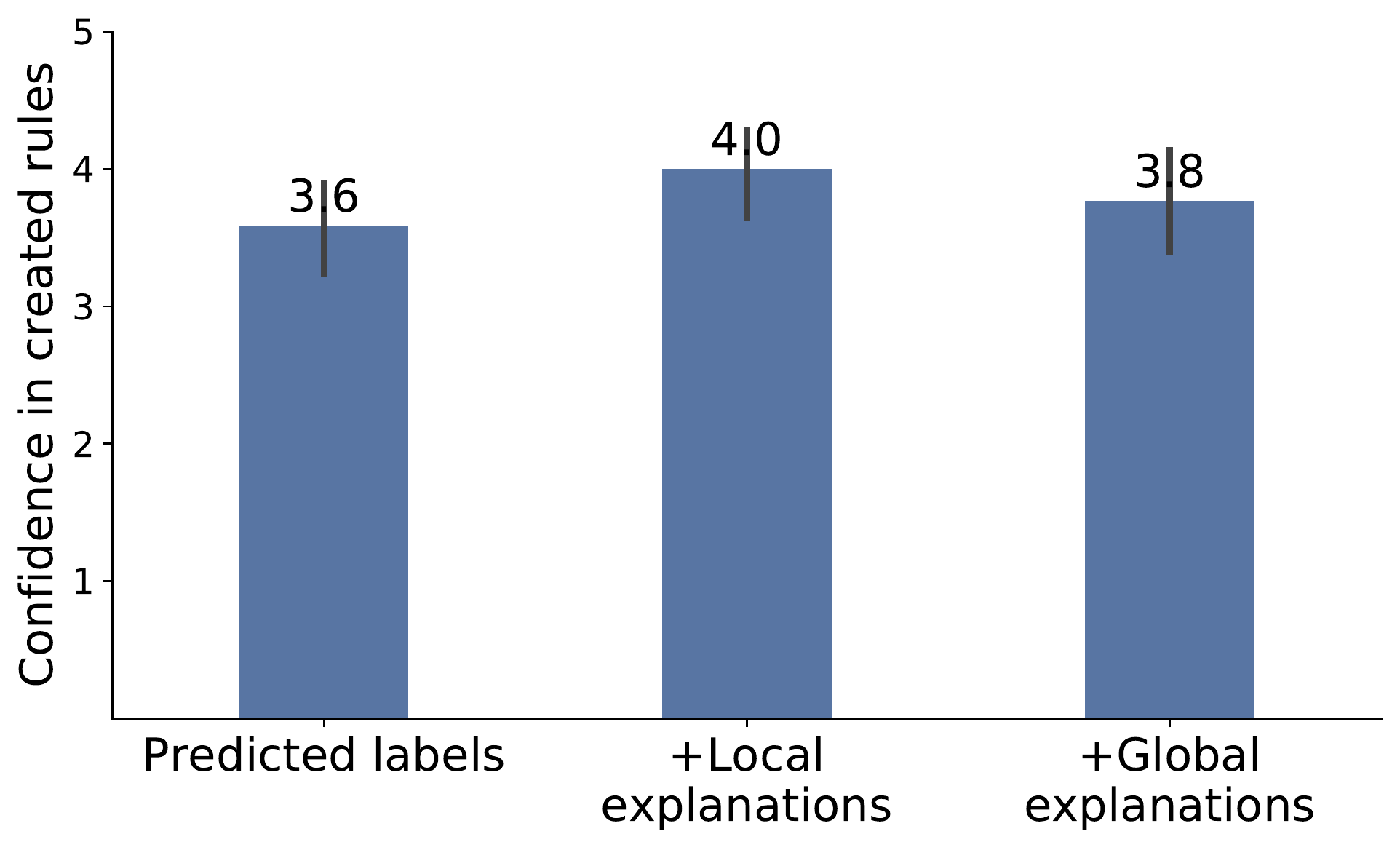}
        \caption{Confidence in created rules (Wiki)}
        \label{fig:confidence_improve_accuracy_wiki}
    \end{subfigure}
    \begin{subfigure}[t]{0.23\textwidth}
        \centering
        \includegraphics[width=\textwidth]{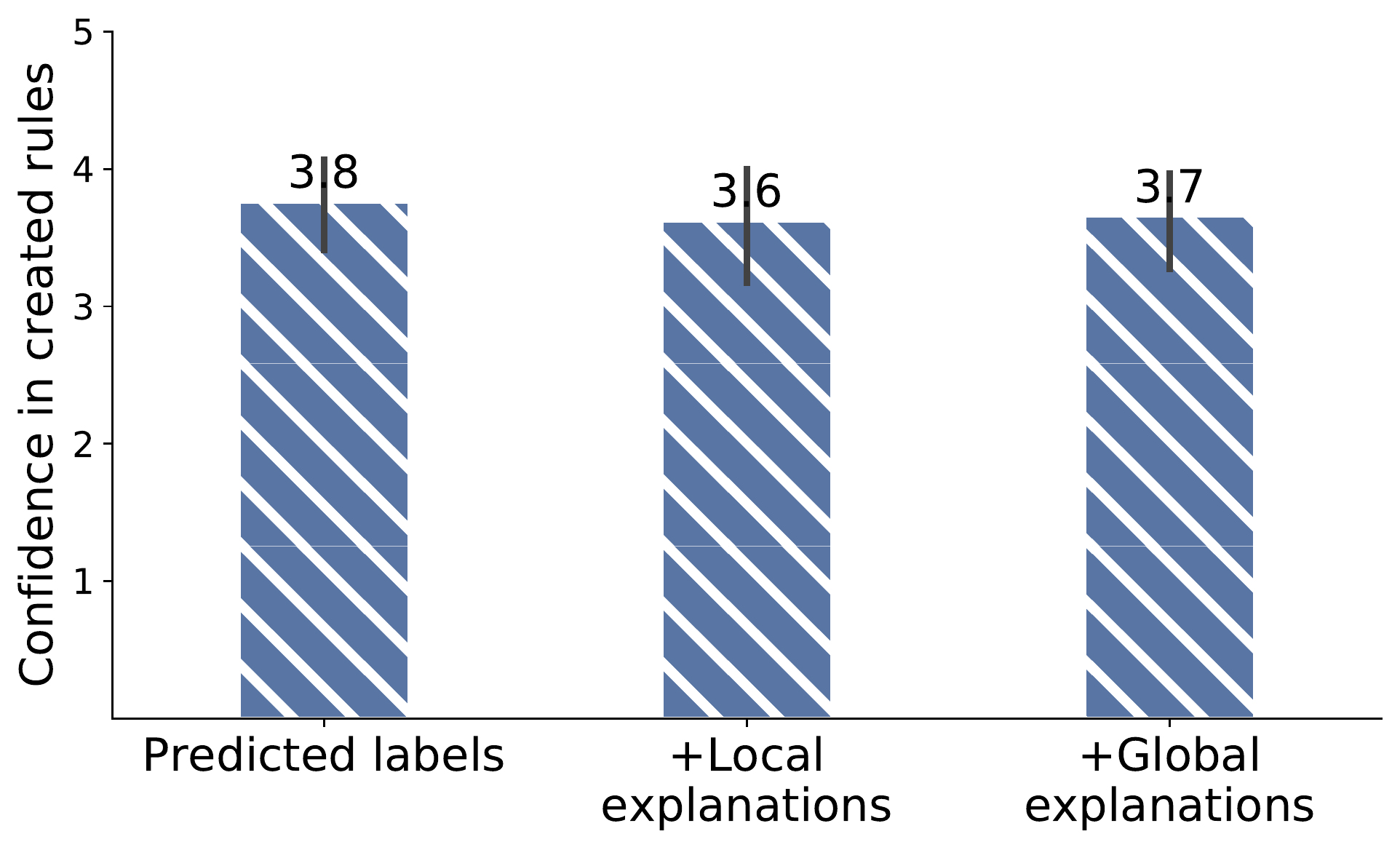}
        \caption{Confidence in created rules (Reddit)}
        \label{fig:confidence_improve_accuracy_reddit}
    \end{subfigure}
    \\
    \begin{subfigure}[t]{0.23\textwidth}
        \centering
        \includegraphics[width=\textwidth]{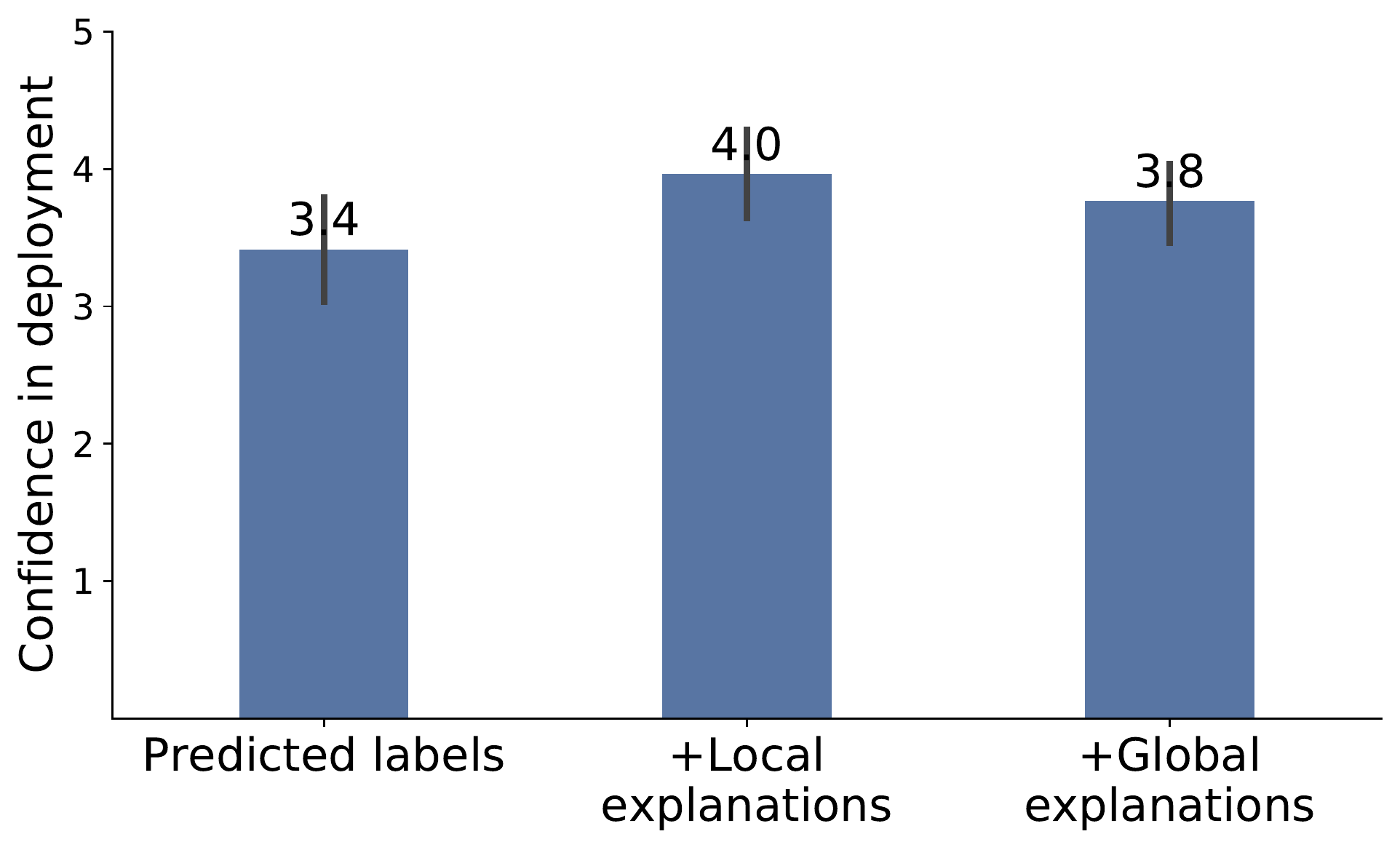}
        \caption{Confidence in deployment (Wiki)}
        \label{fig:confidence_use_my_rules_moderator_wiki}
    \end{subfigure}
    \begin{subfigure}[t]{0.23\textwidth}
        \centering
        \includegraphics[width=\textwidth]{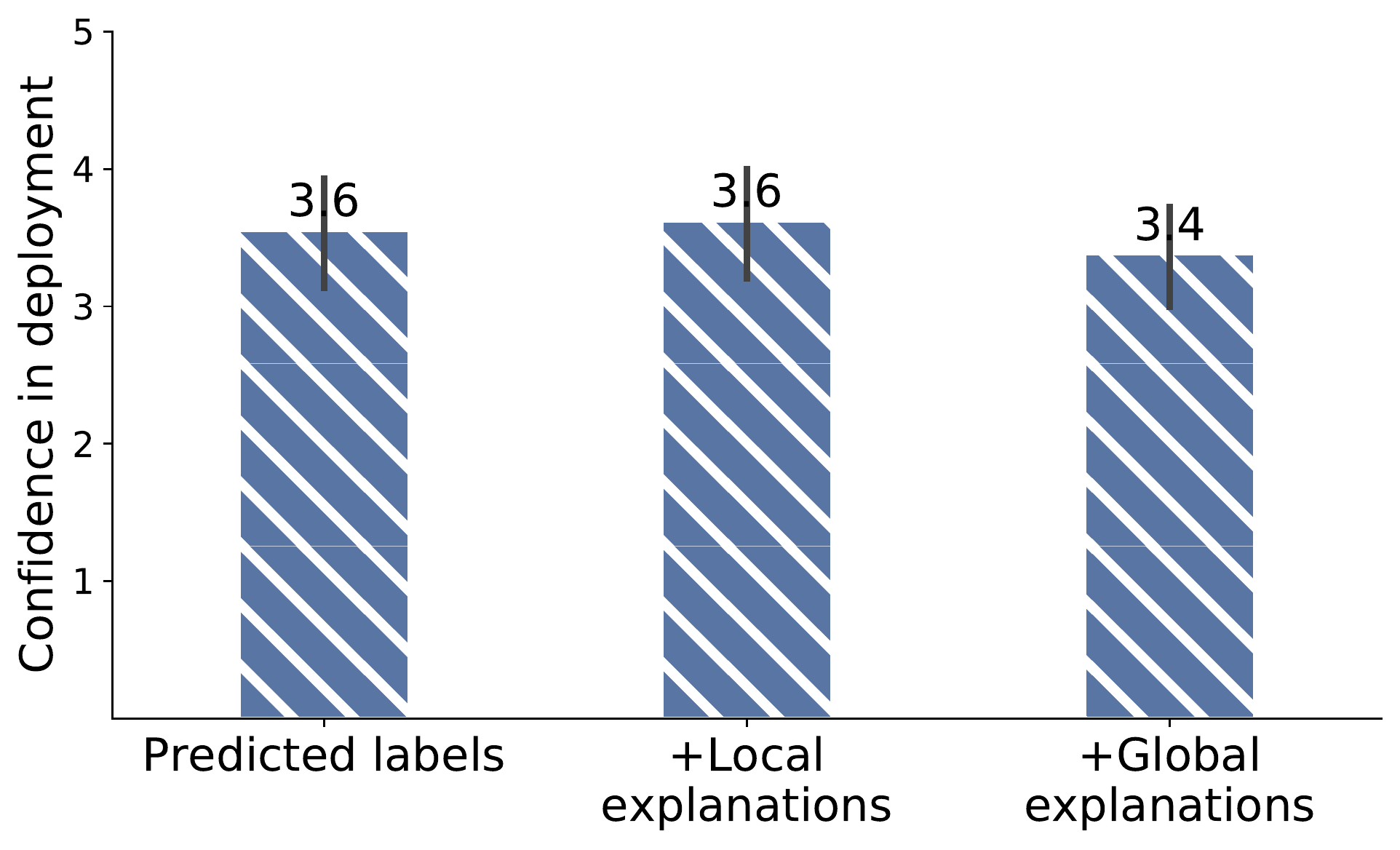}
        \caption{Confidence in deployment (Reddit)}
        \label{fig:confidence_use_my_rules_moderator_reddit}
    \end{subfigure}
    \caption{Confidence. Overall, participants show strong confidence in the model, their performance, and moderators' potential adoption.}
    \Description{The plots show confidence in model, confidence in created rules, and confidence in deployment ratings by different experiment conditions.}
    \label{fig:confidence}
\end{figure}

\para{Confidence.}
Overall, participants reported relatively strong confidence in all of our measures:
confidence in the model (M=3.63, SD=1.01),
confidence in the rules they created (M=3.73, SD=0.98),
confidence in the deployment of the system from human-AI collaboration  (M=3.61, SD=1.0), leaning towards agreeing with all these statements.
We do not find any statistically significant effect of distribution type and experimental condition with two-way ANOVA. There is a non-significant trend that conditions with explanation, especially local explanation, result in better confidence for WikiAttack, but not for Reddit, where the model performs relatively poorly. 

\begin{figure}
    \centering
    \begin{subfigure}[t]{0.23\textwidth}
        \centering
        \includegraphics[width=\textwidth]{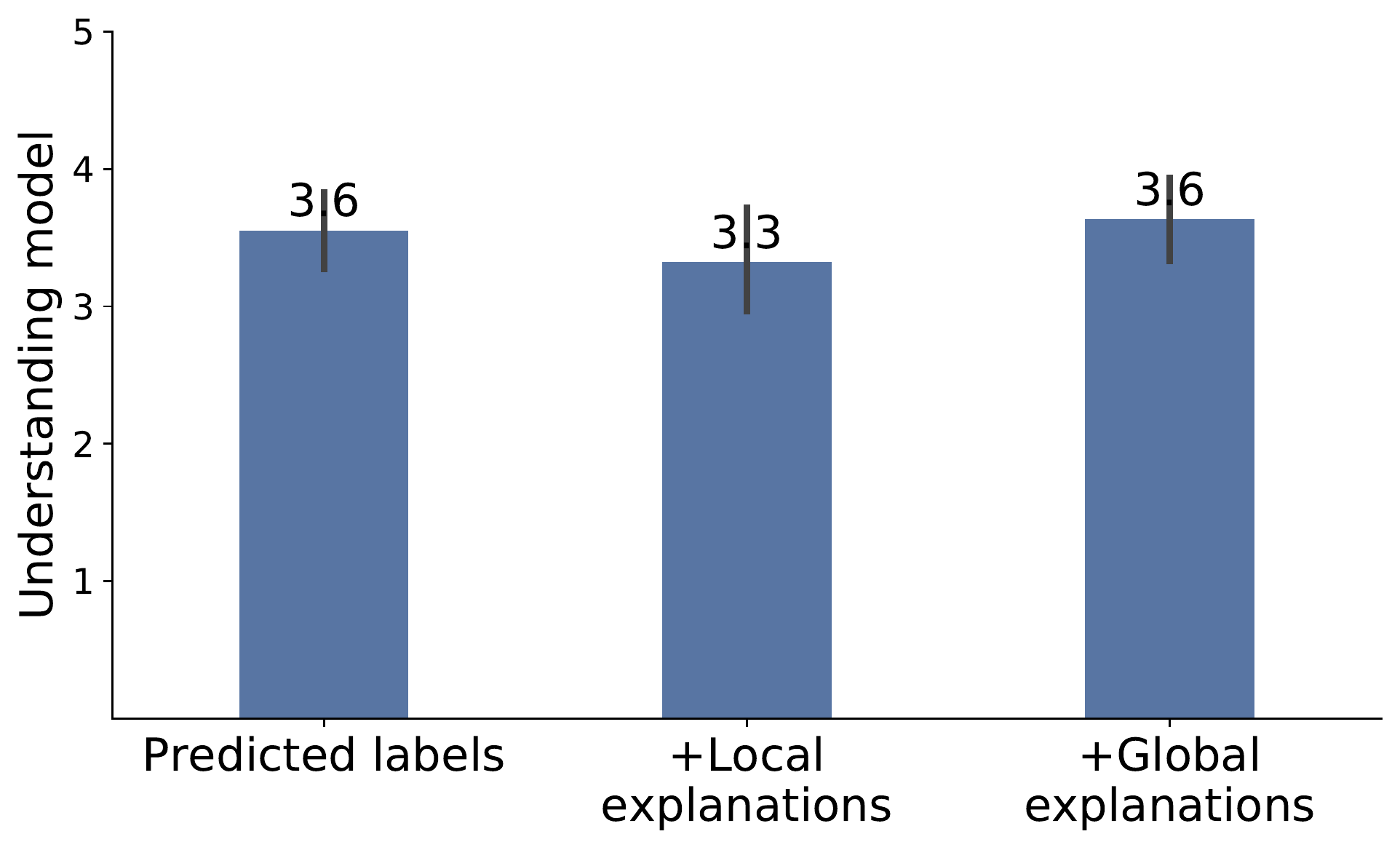}
        \caption{Understanding model (Wiki)}
        \label{fig:understanding_model_wiki}   
    \end{subfigure}
    \begin{subfigure}[t]{0.23\textwidth}
        \centering
        \includegraphics[width=\textwidth]{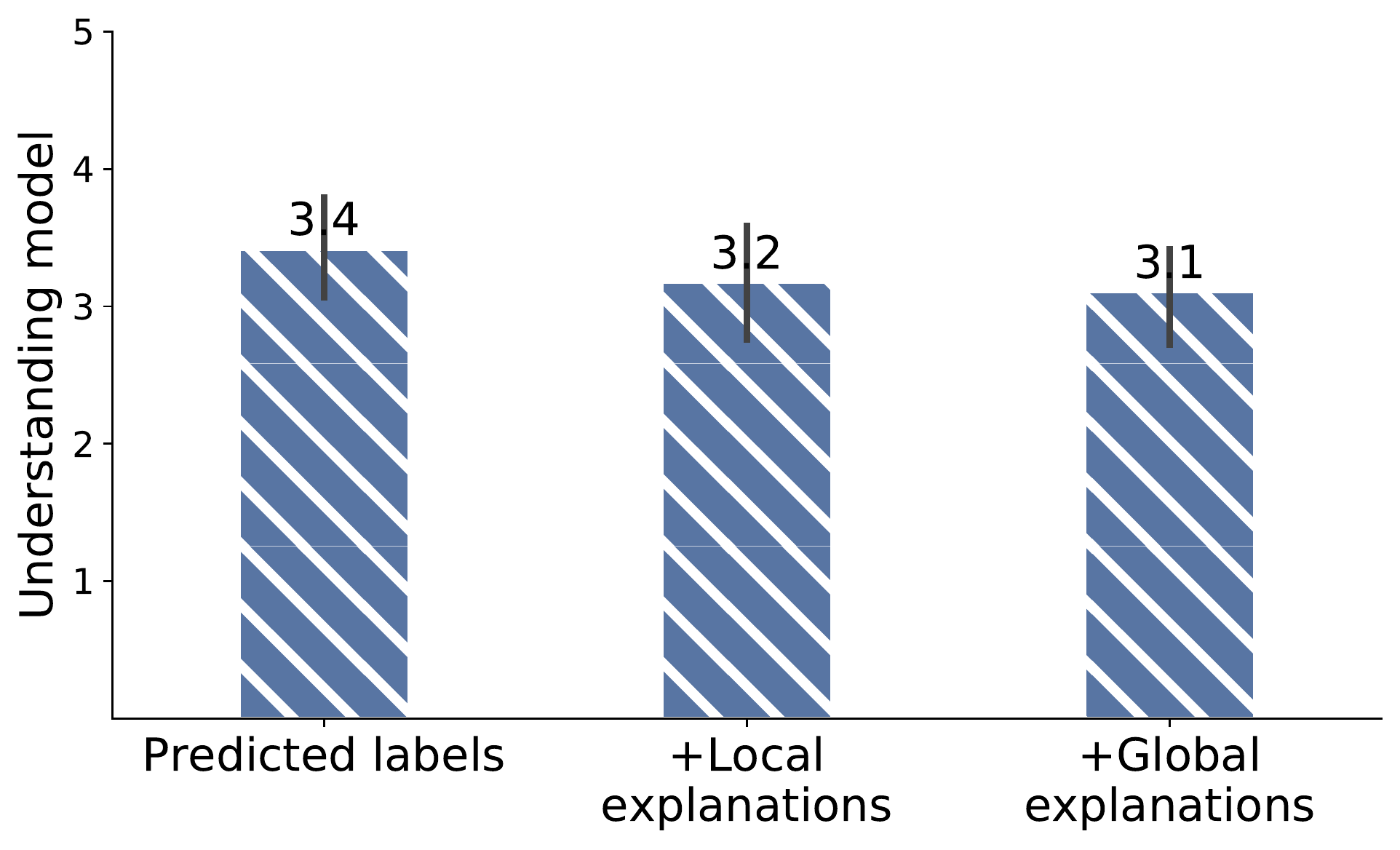}
        \caption{Understanding model (Reddit)}
        \label{fig:understanding_model_reddit} 
    \end{subfigure}
    \begin{subfigure}[t]{0.23\textwidth}
        \centering
        \includegraphics[width=\textwidth]{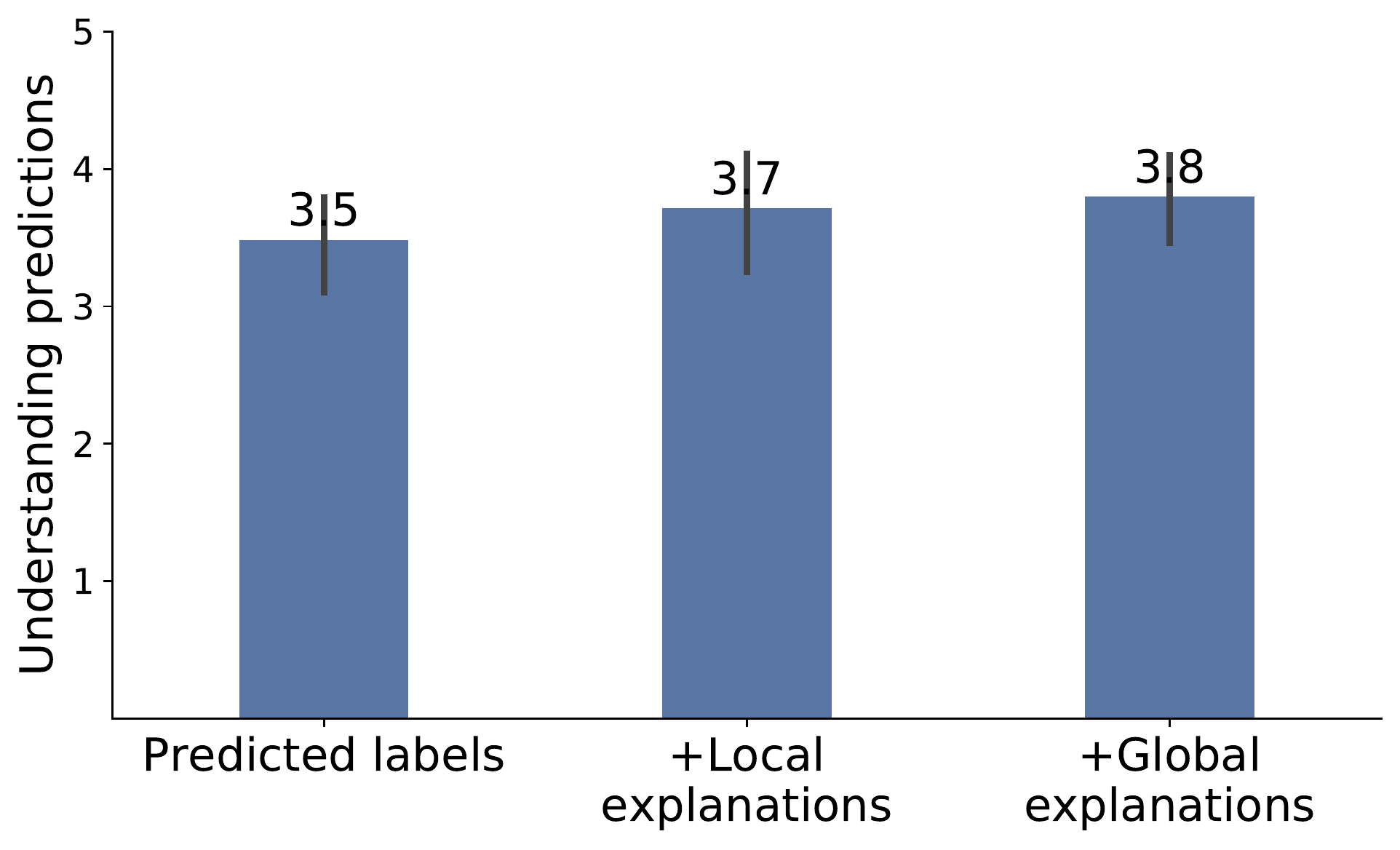}
        \caption{Understanding predictions (Wiki)}
        \label{fig:understanding_predictions_wiki}
    \end{subfigure}
    \begin{subfigure}[t]{0.23\textwidth}
        \centering
        \includegraphics[width=\textwidth]{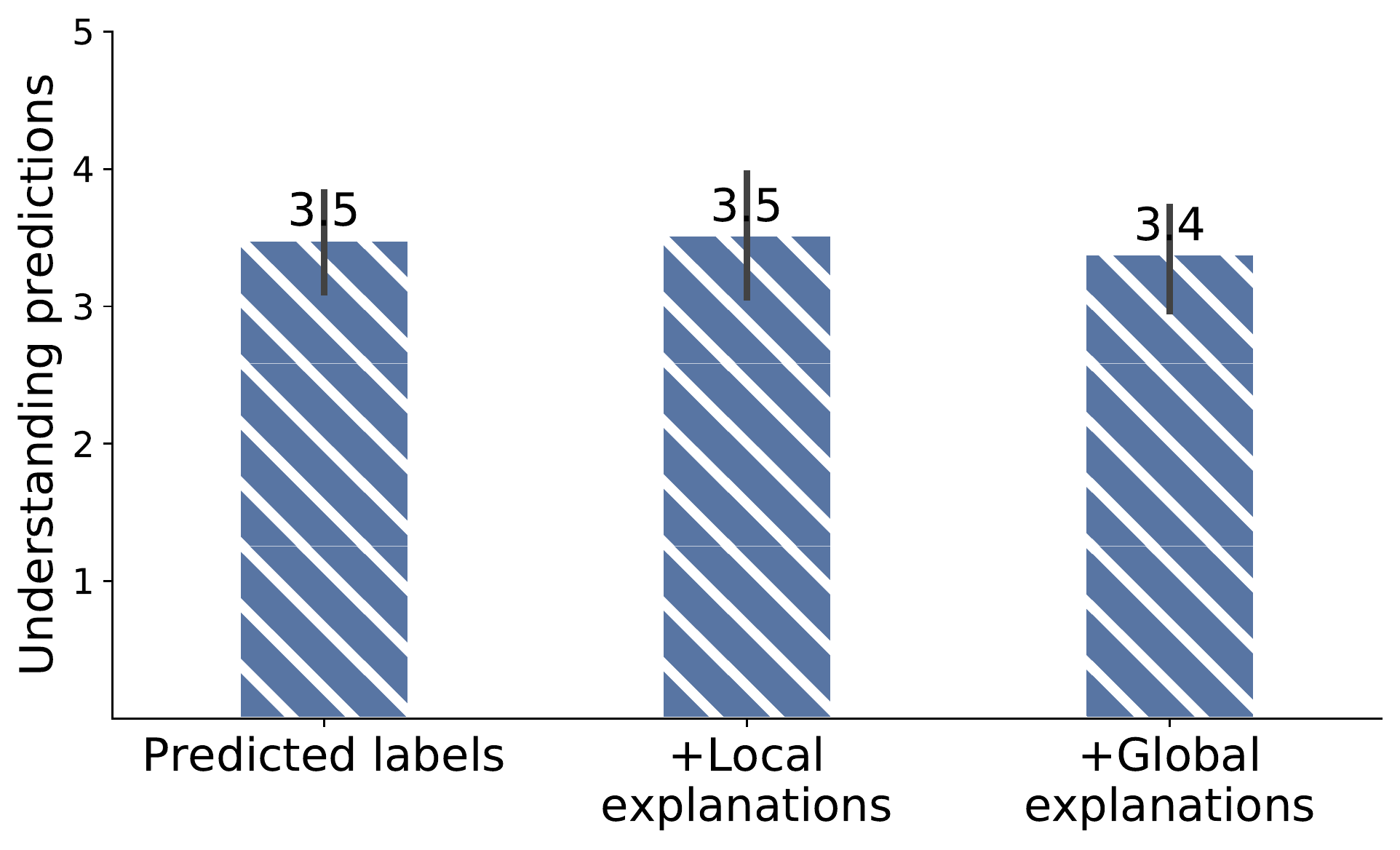}
        \caption{Understanding predictions (Reddit)}
        \label{fig:understanding_predictions_reddit}
    \end{subfigure}
    \caption{Understanding. Overall, participants report a better understanding of the model as a whole and individual predictions on WikiAttack than Reddit, although the differences are not statistically significant.}
    \Description{The plots show understanding of model and predictions ratings by different experiment conditions.}
    \label{fig:understanding}
\end{figure}

\para{Perceived understanding.} Overall, participants report a good global understanding of the model (M=3.37, SD=0.97) and local understanding of individual predictions (M=3.56, SD=1.05) on WikiAttack than Reddit, possibly related to the difference in model performance between distribution types.
Two-way ANOVA only shows a marginally significant effect of distribution type in global understanding of the model ($p=$0.063).
It is somewhat surprising that model performance leads to this difference in perceived understanding. Interestingly, there is a trend that local explanations lead to better perceived local understanding on predictions, but worse perceived global understanding on the model, but not when global explanations are added.

\para{Summary.} Overall, subjective measures show a relatively positive experience across board, but not strong difference between conditions. There is some evidence that when the model performs well (WikiAttack), local explanations provide the best experience: strong performance with relatively high efficiency, less subjective workload and more confidence in the outcomes.

\section{Discussion}
\label{sec:discussion}

Through investigating the three research questions introduced in the beginning, our study shows the promise of conditional delegation as a new paradigm for human-AI collaboration. Even with crowdworkers who are not experts of the content moderation task, conditional delegation can achieve better performance than the model working alone. However, whether the human-AI collaboration can outperform the manual rule-based approach varies for in-distribution and out-of-distribution AI. Out-of-distribution AI has significant performance disadvantage that cannot be adequately compensated by conditional delegation. We also found that, in general, providing predicted labels with our keyword search based interface is sufficiently effective in supporting people to create delegation rules. Providing explanations can improve efficiency by hinting on rules to consider, but can also mislead people to use high-frequency but not necessarily high-precision rules. We discuss implications of these results below.

\para{The promise of conditional delegation.}
Our study is a first step towards understanding and leveraging the promise of conditional delegation.
It is an intuitive approach that can be used in a wide variety of domains so that users can proactively decide when to use an AI model and in what ways based on the output of the AI.
For instance, judges can specify when to show the risk estimates for recidivism prediction and when to hide the model output. Doctors can identify subsets of patients for which they rely on AI to send alerts. 
Our study only explored one type of workflow and one type of action. Different applications may require a diverse set of workflows and actions, and have varying tradeoffs between false positives and false negatives.

Moreover, we only begin to define the design space for supporting users in conditional delegation.
An essential requirement is to help users make sense of model behaviors under different delegation conditions.
Keyword-based rules are a reasonable approach in content moderation given that rule-based methods are already used in AutoModerator.
\added{We used a rationale-style model to facilitate this kind of interaction, although we expect post-hoc explanation methods to play similar roles. It is important to empirically validate the effect of the underlying models and explanation methods.}
\added{We also focused exclusively on conditional delegation based on trustworthy regions in the input space. A promising direction is to investigate the joint effect of delegation based on inputs and outputs \citep{chandrasekharan2019crossmod,keswani2021towards}.}
\replaced{Another}{One} limitation of our study lies in that crowdworkers only came up with about 10 rules because of our minimum requirement. 
\added{Although our results are encouraging with non-expert users, 
additional experiments are required to validate the potential of conditional delegation with expert users.
Notably,}
future research can explore means to facilitate people to identify a greater number of rules, examine the combined effect of rules, and monitor the performance of rules after model deployment.
\added{In long-term deployment, it is especially valuable to investigate how to update the delegation conditions once the model is updated.}

\added{Additionally, conditional delegation can potentially alleviate AI bias as we give users the freedom to choose trustworthy regions based on their domain knowledge or notion of fairness. However, the flip side is that this process could introduce human bias, if for example one's notion of fairness is ill conceived.  We encourage future work to understand and develop ways to rail-guard the impact of human biases in conditional delegation.
}

\para{The effect of distribution shift.}
Our results highlight the importance of considering the effect of distribution shift in designing experiments on human-AI collaboration, to better understand the generalizability of results.
We are able to achieve complementary performance \ind on WikiAttack, but not \ood on Reddit.
In practice, it is rarely the case that an AI model faces exactly the same distribution in deployment.
Therefore, it is critical to understand the outcomes in \ood contexts to understand the generalizability or applicable scope of a given form of human-AI collaboration. 

It is useful to note that although our results are presented as \ind vs \ood, the differences are complicated between WikiAttack and Reddit.
First, there exists a clear difference in model performance, so our results can be seen as comparing a high-performance model with a low-performance model.
Second, the nature of comments on Wikipedia and Reddit differs substantially.
It is possible that crowdworkers are more used to comments on Reddit or that common swearing words such as ``retard'' and ``cunt'' happened to work well on the Reddit dataset that we used.
This complexity demonstrates that the contrast of \ind versus \ood contexts is not a monolithic dimension, which further adds to the challenge of experimental design to account for the effect of distribution shift.

\para{The priming effect of explanations.}
While explanations can improve efficiency, global explanations are found to slightly hurt performance when working with \ood AI, as participants may have chosen the keywords in explanations without carefully examining the model behaviors with them. These observations echo concerns of unintended consequences with the use of explanations in human-AI collaboration \citep{lai2019human,bansal2021does,green2019principles}. 

In other words, for our task of creating keywords rules, keywords-based explanations have a priming effect that leads to biased adoption of presented words. Note that priming, if used appropriately, can shape user behaviors in a positive way. The challenge is that with the technique we used to generate global explanations (i.e., most frequent tokens in rationale), the top tokens do not necessarily correlate with high precision (\figref{fig:frequent_word_precision}). Future work can explore techniques that can exploit some proxy of precision, such as considering the uncertainty or confidence of predictions. Another direction is to utilize de-biasing technique to mitigate the effect of priming, such as explicitly reminding people to attend to wrong predictions with the chosen keywords. 

It is worth noting that local explanations seem to have less of a priming effect than global explanations but still improves efficiency. It is possible that the many highlights in search results are too scattered to have a salient effect. Future work can explore other XAI techniques or provide additional support, such as to help users have an overview of the rationales in all search results.

\para{Implications on content moderation.}
It is impressive that crowdworkers can already create keyword-based rules that achieve greater precision than the model working alone.
However, we recognize that our experiment setup is only a first step towards using conditional delegation in content moderation.
First, crowdworkers are not representative of moderators, who have way more experience with their platform's data.
\added{As moderators are more familiar with the moderation process and more knowledgeable about important words, experts might find the interface more useful than crowdworkers. However, participatory design and future work can develop more serendipitous features.}
Second, in practice, moderators usually have historical data on which moderation decisions were made.
This historical data can be used in the process of creating keyword-based rules.
Third, prior work has shown that moderators often update the rules used by AutoModerator \citep{jhaver2019human,chandrasekharan2019crossmod} and our work does not take into account any future updates.
Neither do we leverage any existing rules that moderators have created.
\added{For future work, we hope to integrate a model that receives feedback from moderators and allow updates to the model to reflect the feedback. The ideal pipeline would require careful development in the model architecture and interface to refrain any unnecessary actions from interfering with moderators' tasks.}
Last but not least, content moderation involves a wide range of different rules beyond toxicity, and even the policies under the umbrella term of toxicity can vary, so the AI model that we uses represents a narrow component in content moderation.
In short, our work uses content moderation as a testbed to illustrate the promise of conditional delegation. Much future work is required to realize the impact of conditional delegation in content moderation.

\para{Limitations.}
First, our work represents one instantiation of conditional delegation.
We emphasize precision and coverage to increase the ability of moderators to deal with a large amount of comments (``true positives'') while minimizing unnecessary labor for moderators (``false positives'').
This tradeoff between true positives, true negatives, false positives, and false negatives can vary in practice depending on the application and the actions taken according to AI predictions.
Second, our participants are not representative of content moderators. 
It also follows that our evaluations are limited by the number of rules that participants created in about 10 minutes.
Our case study shows the promise of conditional delegation, but further study is required in each application domain of interest to develop the best design for human-AI collaboration in identifying delegation conditions.
Third, our choice of model, datasets, and explanations affect the experimental outcome.
It is important to further dissect the relevant dimensions and investigate the effect of alternative choices.

\section*{Acknowledgments}
We thank all anonymous reviewers for their insightful suggestions and comments. We thank all members of the Chicago Human+AI Lab for feedbacks on early versions of our website interface. 
All experiments were approved by the University of Colorado IRB (21-0385). 
This work was supported in part by NSF grants IIS-1837986, 2125116, and 2125113.

\bibliographystyle{ACM-Reference-Format}
\bibliography{refs}

\end{document}